\documentclass{article}
\pdfoutput=1
\PassOptionsToPackage{table}{xcolor}
\usepackage[utf8]{inputenc}
\usepackage{main} %

\usepackage{xcolor}
\usepackage{microtype}
\usepackage{graphicx}
\usepackage{times}
\usepackage{rotating}
\usepackage{amsmath}
\usepackage{amssymb}
\usepackage{amsthm}
\usepackage{pifont}
\usepackage{enumitem}
\usepackage{booktabs}
\usepackage{makecell}
\usepackage{url}
\usepackage{tabularx}
\usepackage{multirow}
\usepackage{framed}
\usepackage{tikz}
\usepackage{fontawesome5}
\usepackage{xspace}
\usepackage{color-edits}
\usepackage{wrapfig}
\usepackage{fancyhdr} %

\usetikzlibrary{calc, positioning, shadows.blur, shapes.geometric, arrows.meta}

\definecolor{mydarkblue}{rgb}{0,0.08,0.45}
\usepackage[
  colorlinks=true,
  linkcolor=mydarkblue,
  citecolor=mydarkblue,
  filecolor=mydarkblue,
  urlcolor=mydarkblue
]{hyperref}
\usepackage[capitalise,noabbrev]{cleveref}

\setcitestyle{authoryear,round,citesep={;},aysep={,},yysep={;}}

\providecolor{gred}{RGB}{250, 210, 207}
\providecolor{coolblue1}{rgb}{0.91, 0.94, 0.98}
\providecolor{coolblue2}{rgb}{0.76, 0.85, 0.94}
\providecolor{coolblue3}{rgb}{0.54, 0.72, 0.87}
\providecolor{coolblue4}{rgb}{1, 1, 1}

\newcommand{\circled}[1]{\ding{\the\numexpr#1+191}}

\addauthor{sw}{purple}
\addauthor{xz}{blue}
\addauthor{cm}{cyan}
\addauthor{gn}{magenta}

\newenvironment{itemize*}
 {\leftmargini=10pt\begin{itemize}
  \setlength{\itemsep}{0pt}
  \setlength{\parskip}{0pt}}
 {\end{itemize}}

\newenvironment{enumerate*}
 {\begin{enumerate}
  \setlength{\itemsep}{0pt}
  \setlength{\parskip}{0pt}}
 {\end{enumerate}}

\newcommand{\odyssim}{\ensuremath{\mathcal{O}}\textsc{dysSim}\xspace}
\newcommand{\odyssimmodel}{\ensuremath{\mathcal{O}}\textsc{sim}\xspace}
\newcommand{\soul}{\textsc{Soul}\xspace}
\newcommand{\soulindex}{\textsc{Soul}-Index\xspace}
\newcommand{\vfrl}{RLVF\xspace}
\newcommand{\taubench}{$\tau$-bench\xspace}
\newcommand{\tauusi}{$\tau$-USI\xspace}
\newcommand{\taunum}{31\xspace}
\newcommand{\midtrainingdataurl}{https://huggingface.co/datasets/cmu-lti/osim-mid-training}
\newcommand{\posttrainingdataurl}{https://huggingface.co/datasets/cmu-lti/osim-post-training}
\newcommand{\midtrainingdata}{\href{\midtrainingdataurl}{\texttt{cmu-lti/osim-mid-training}}\xspace}
\newcommand{\posttrainingdata}{\href{\posttrainingdataurl}{\texttt{cmu-lti/osim-post-training}}\xspace}

\begin{document}

\title{{\Huge \odyssim}\\
Building Foundation Models for Human Behavior Simulation}

\author{
\textbf{Xuhui Zhou}$^{1}$\thanks{Equal contribution.} \quad
\textbf{Weiwei Sun}$^{1}$\footnotemark[1] \quad
\textbf{Weihua Du}$^{1}$ \quad
\textbf{Jiarui Liu}$^{1}$ \quad
\textbf{Haojia Sun}$^{1}$ \\
\textbf{Qianou Ma}$^{1}$ \quad
\textbf{Tongshuang Wu}$^{1}$ \quad
\textbf{Yiming Yang}$^{1}$ \quad
\textbf{Maarten Sap}$^{1}$ \\
\textsuperscript{1}Carnegie Mellon University, Language Technologies Institute \\
\texttt{\{xuhuiz, weiweis\}@andrew.cmu.edu} \\[0.5em]
\normalfont
\raisebox{-0.1em}{\faGithub}~\href{https://github.com/sunnweiwei/OdysSim}{Code} \quad
\raisebox{-0.2em}{\includegraphics[height=1em]{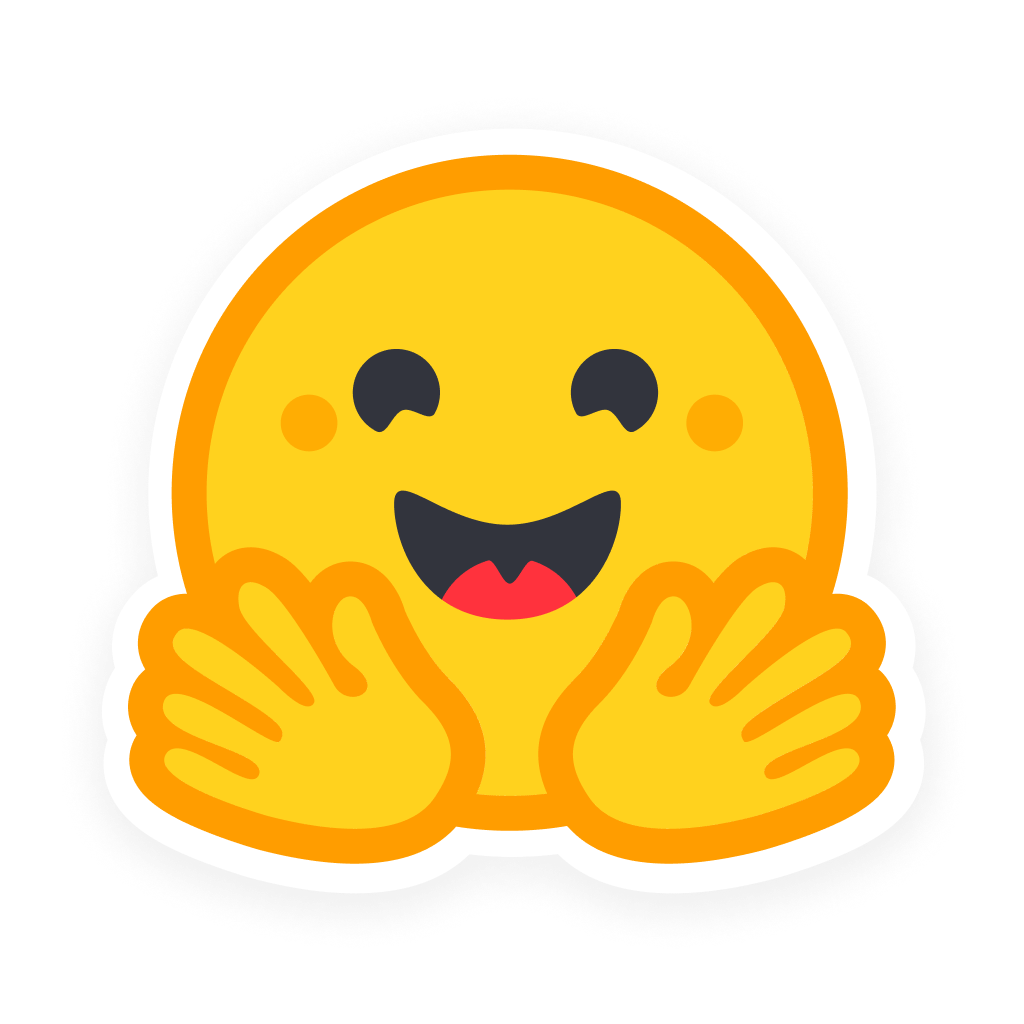}}~\href{https://huggingface.co/collections/cmu-lti/odyssim}{Model} \quad
\raisebox{-0.2em}{\includegraphics[height=1em]{logo/huggingface.png}}~\href{\midtrainingdataurl}{Midtraining Data} \quad
\raisebox{-0.2em}{\includegraphics[height=1em]{logo/huggingface.png}}~\href{\posttrainingdataurl}{Post-training Data}
}

\maketitle

\thispagestyle{fancy}
\fancyhead{}
\lhead{\includegraphics[height=0.5cm]{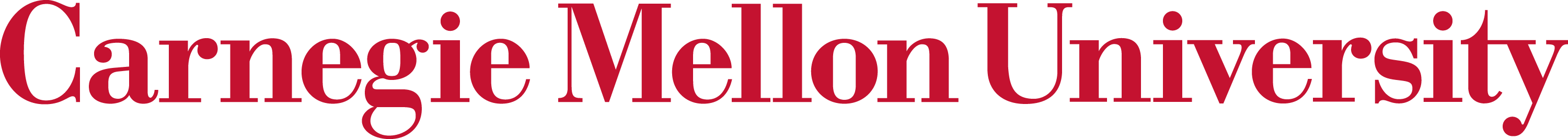}}
\rhead{\raisebox{-0.1cm}{\includegraphics[height=0.8cm]{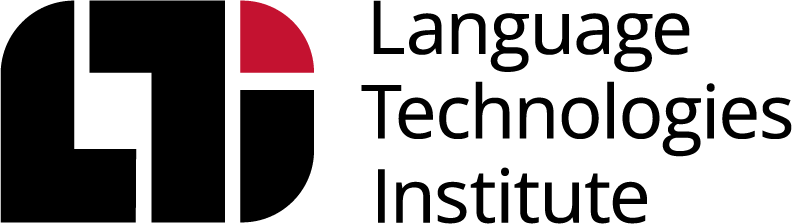}}}
\renewcommand{\headrulewidth}{0pt}
\setlength{\headheight}{12pt}
\setlength{\headsep}{3mm}
\pagestyle{plain}

\vspace{-1.0em}

\begin{abstract}
Large language models are increasingly deployed as human simulators for interactive evaluation and social simulation. Yet helpfulness-driven post-training pulls them toward a homogeneous, overly agreeable assistant register, creating a behavioral Sim2Real gap.
We present \odyssim, the largest open systematic investigation of \textit{behavioral foundation models}, i.e., models trained to simulate human behavior at scale. We propose \soul, a taxonomy of five capability \emph{axes} (\texttt{CONV}, \texttt{SS}, \texttt{COG}, \texttt{ROLE}, \texttt{EVAL}) that unifies 62 datasets and 23 benchmark tasks under one framework.
Specifically, we curate the \odyssim corpus (21.4M interactions,
10B tokens, retrofitted with back-generated social contexts), construct the \soulindex benchmark, and develop an end-to-end training recipe combining midtraining, task-specific RL, and expert distillation.
The resulting open 8B \odyssimmodel model ranks first or tied-first on 8 of 23 tasks, outperforming any individual frontier model by this count, with the strongest gains on conversational and social tasks. Its outputs are also more human-like in length, formatting, and word choice, and it transfers zero-shot to out-of-distribution user simulation on \taubench, nearly matching real users on reaction alignment (93.2 vs.\ 93.5). We further show that LLM-as-judge RL induces reward-hacking patterns, and that our detectors can mitigate them during post-training. Together, our findings suggest that behavioral foundation models require rethinking the LLM training paradigm. We release all artifacts to support future research.
\end{abstract}

\begin{figure*}[h]
\centering
\includegraphics[width=\textwidth]{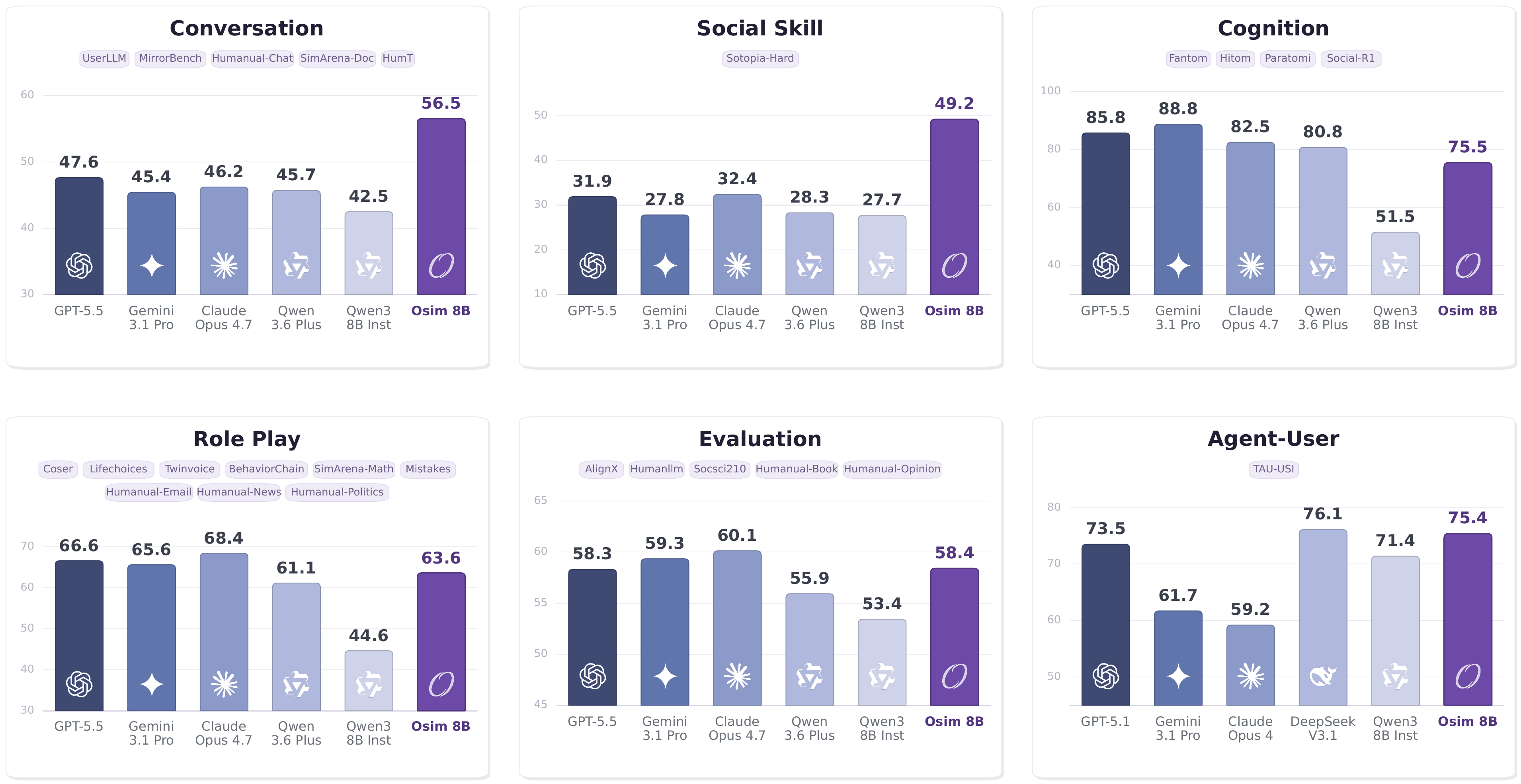}
\caption{Benchmark results on human simulation tasks.}\label{fig:benchmark}
\end{figure*}

\section{Introduction}
\label{sec:introduction}

Simulating human behavior is becoming a critical capability for AI systems.
Realistic behavioral models are needed for user simulation in agent evaluation~\citep{yao2024taubenchbenchmarktoolagentuserinteraction}, patient simulation in clinical training~\citep{kyung2025patientsimpersonadrivensimulatorrealistic}, learner simulation in educational technology~\citep{ross2025learningmakemistakesmodeling}, and persona simulation in social science~\citep{park2023generative, argyle2023out}.
Yet current large language models (LLMs) fall short: they are systematically biased, stylistically uniform, and excessively agreeable, exhibiting what has been termed the \emph{Sim2Real gap}~\citep{zhou2025simreal}, and prompting alone does not suffice~\citep[especially for more ``undesirable'' behaviors that humans naturally display;][]{li-etal-2025-big5}.
The root cause lies in the LLM training pipeline: (i) standard pretraining ingests vast amounts of internet text, including but not necessarily real human behavior, (ii) helpfulness-driven post-training \citep[e.g., RLHF;][]{ouyang2022training} actively pulls models toward an assistant register, and (iii) evaluation protocols typically reward task success and instruction following, while leaving behavioral realism, diversity, and social fidelity under-specified.

\begin{figure}[t]
    \centering
    \includegraphics[width=\textwidth]{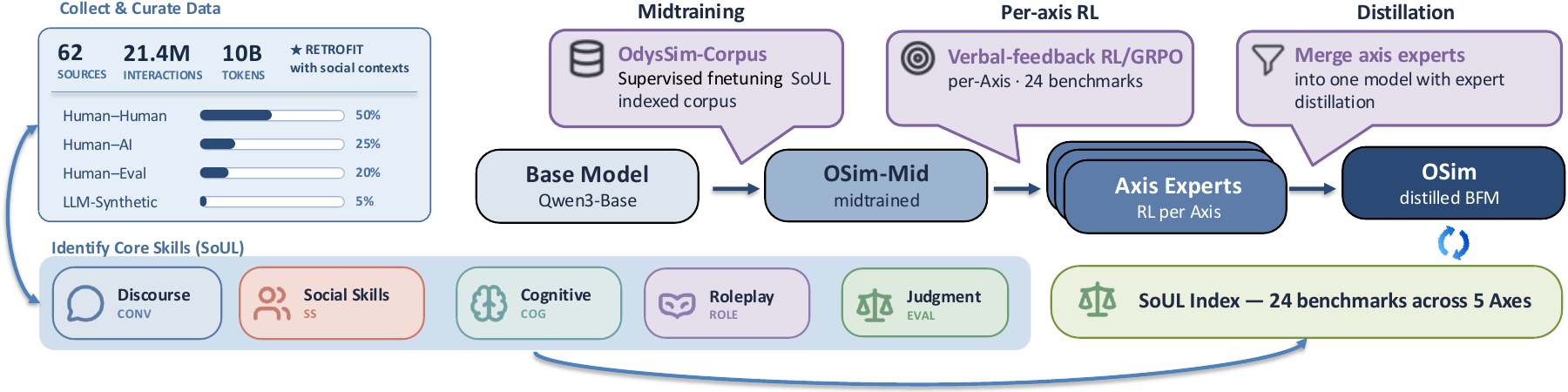}
    \vspace{-1.5em}
    \caption{Overview of the \odyssim recipe. We iteratively collect and curate the \odyssim corpus, build the \soul framework, and construct the \soulindex as an evaluation suite. We first midtrain a base Qwen3 checkpoint into \odyssimmodel-Mid. Then we follow with task-specific RL, training one expert per \soulindex task. Expert distillation then merges these experts into the final \odyssimmodel{} model.}
    \label{fig:recipe-overview}
\end{figure}

Closing the gap requires rethinking the pipeline end-to-end: \emph{what we measure}, \emph{what data} the model learns from, and \emph{how} it is trained.
We present \odyssim, the largest open effort to build a behavioral foundation model\footnote{We use \emph{behavioral foundation model} in the natural-language sense throughout: a model trained at scale to simulate human behavior in linguistic interaction. This is distinct from the embodied-control sense used in robotics, where ``behavior foundation model'' refers to whole-body motor control policies for humanoid robots~\citep{tirinzoni2024metamotivo,zeng2025bfmhumanoid}.} to date, comprising a 23-task benchmark, a 21.4M-interaction (10B-token) midtraining corpus from 62 public sources, and an end-to-end RL recipe.
Additionally, as shown in Figure \ref{fig:recipe-overview}, we design \soul (Simulation Of hUman-Like behavior), a framework that defines five capability \emph{Axes} (\texttt{CONV}, \texttt{SS}, \texttt{COG}, \texttt{ROLE}, \texttt{EVAL}) to jointly index the \odyssim corpus and the \soulindex evaluation suite (\cref{sec:soul-index}).
Behavior simulation is inherently grounded: to simulate a human response, a model must condition not only on an input utterance or situation, but also on who the speaker is, what role they occupy, and what social intent shapes the interaction. We therefore formalize behavioral simulation as generating a response given both an interaction context and a social grounding specification, such as a character profile, role, or goal. This creates a practical data challenge: many raw sources used for midtraining, such as WildChat entries \citep{zhao2024wildchat} and ConvoKit threads \citep{chang2020convokit}, contain rich dialogue but lack explicit speaker grounding, making social dynamics hard to infer from text alone.
We address this by retrofitting each dialogue with back-generated social contexts, including a character profile and interaction goal. We further show supporting training with social grounding context is important for learning to simulate human behavior.

At the core of our investigation, we midtrain Qwen3 base models on the \odyssim corpus to create \odyssimmodel-Mid.
With the \odyssimmodel-Mid, we further perform task-specific reinforcement learning and create an expert model for each \soulindex task: GRPO when the task has a verifiable reward, and RL with verbal feedback when the task is judged by an LLM that returns both a scalar reward and a textual critique~\citep{sun2026reinforcinghumanbehaviorsimulation,song2026rltf}.
Finally, we use expert distillation to merge the resulting task-specific experts into a single deployable model.
The two stages are complementary: midtraining provides a behaviorally-aware initialization (\emph{what} human behavior looks like at scale); task-specific RL adds precision under the right reward signal (\emph{how} to behave on each task).

Putting the recipe together: midtraining, then task-specific RL, then expert distillation, yields \odyssimmodel-8B, which leads \soulindex and is best or tied-best on more tasks (8 of 23) than any individual frontier model.
Its improvements concentrate on the interactive, socially grounded tasks that general-purpose post-training underscores, producing more ``human-like'' behaviors such as shorter sentences and fewer assistant-like phrases.
The gains also transfer beyond chat settings: on \tauusi, an out-of-distribution user-simulation evaluation for tool-use agents, \odyssimmodel-8B achieves the strongest reaction alignment among evaluated simulators, nearly matching real users (React $93.2$ vs.\ $93.5$), outperforming any frontier models.
Our ablations show that the two stages contribute in qualitatively different ways: midtraining alone shifts outputs toward the human register in length, formatting, and word choice, lifting Qwen3-8B-Base from $26.9$ to $41.1$ on \soulindex; task-specific RL then adds the largest gains on role-playing (\texttt{ROLE}) and conversational (\texttt{CONV}) tasks.
Together, these results close the loop on our central claim: building behavioral foundation models requires aligning what we measure (\soulindex), what data the model learns from (\odyssim corpus), and what the training objective rewards around behavioral realism rather than task success alone.

\paragraph{Contributions.}
\textbf{(1) The \soul Framework.} A single set of five behavioral-capability \emph{Axes} that jointly guides the midtraining, post-training, and evaluation, together with the \soulindex --- to our knowledge the most comprehensive open evaluation for human-behavior simulation. 
\textbf{(2) The \odyssim Corpus.} A behavioral midtraining corpus of 21.4M interactions ($\approx$10B tokens) from 62 public sources, unified into a common conversational format and equipped with a \emph{retrofit pipeline} that back-generates per-conversation social groundings (e.g., character profile, interaction goal).
\textbf{(3) End-to-End Recipe.} Midtraining on \odyssim, task-specific RL on each \soulindex task (GRPO and \vfrl), and expert distillation into a single final model that improves both in-benchmark behavior simulation and zero-shot user simulation for tool-using agents.

\section{Related Work}
\label{sec:related_work}

\paragraph{Evaluating Behavioral Simulation.}
Existing benchmarks for behavioral simulation focus on specific aspect of human behavior, each targeting narrow capability or task format: theory of mind~\citep{kim2023fantom,le2019tomi}, social interaction~\citep{zhou2024sotopia}, role-play with persona~\citep{wang2025coser,kirk2024prism, li2025alignx}, social and cognitive experiments~\citep{kolluri2025socsci,binz2024psych101} or more recently, user behavior simulation interacting with AI agents~\citep{yao2025simulatorarena,zhou2025simreal}.
This fragmentation makes it hard to track the overall progress in one aspect (say, theory-of-mind accuracy) implies anything about a different capability (say, role-play fidelity), or to compare modeling approaches that target different capabilities.

\paragraph{Training Behavioral Foundation Models.}
Prior efforts to train LLMs for human-behavior simulation differ in methodology and scale, but each remains bounded to narrow behavioral domain.
Many adapt a general-purpose post-trained LLM via SFT or RL: \textsc{Sotopia-$\pi$}~\citep{wang2024sotopia} clones expert social-interaction trajectories, \textsc{Sotopia-RL}~\citep{yu2025sotopiarl} adds utterance-level multi-dimensional rewards, \textsc{Omar}~\citep{chen2026omar} trains via multi-agent self-play, and \textsc{UserLM}~\citep{naous2025userllm} finetune for the user side of \textsc{WildChat}~\citep{zhao2024wildchat} dialogues.
Others build new corpora but stay within a single domain: \textsc{Centaur}~\citep{binz2024psych101} finetunes on \textsc{Psych-101} (10M choices from 160 cognitive-psychology experiments), \textsc{Be.FM}~\citep{xie2025befm} targets four behavioral-science capabilities, \textsc{Socrates}~\citep{kolluri2025socsci} finetunes on \textsc{SocSci210} (2.9M social-science responses).
While \citet{sun2026reinforcinghumanbehaviorsimulation} and \citet{wu2026humanlm} investigate diverse domains of tasks and capabilities, both initialize from instruction-tuned models already optimized to act as helpful assistants, which risks suppressing the behavioral diversity needed to faithfully simulate human behavior.

\paragraph{Midtraining and RL with LLM-Judge Feedback.}
Midtraining adapts pretrained models to a target distribution before post-training~\citep{gururangan2020continuedpretraining,liu2026midtraining,mo2025midtraininglargelanguagemodels}, but prior work mainly studies domains such as code~\citep{roziere2023codellama} and math, where the shift is largely lexical, syntactic, or task-skill driven.
Our setting differs: human-behavior simulation requires socially grounded shifts in persona, intent, register, and interaction style, which have not been systematically studied under midtraining.
For open-ended behavioral tasks, prior work uses LLM judges and sometimes textual feedback to optimize generations~\citep{zheng2024judging,verga2024poll,sun2026reinforcinghumanbehaviorsimulation,song2026rltf}.
We build on this line, but focus on behavioral fidelity rather than helpfulness or task success.
\section{The \soul Framework}
\label{sec:soul}

\begin{figure}[!t]
    \centering
    \includegraphics[width=\textwidth]{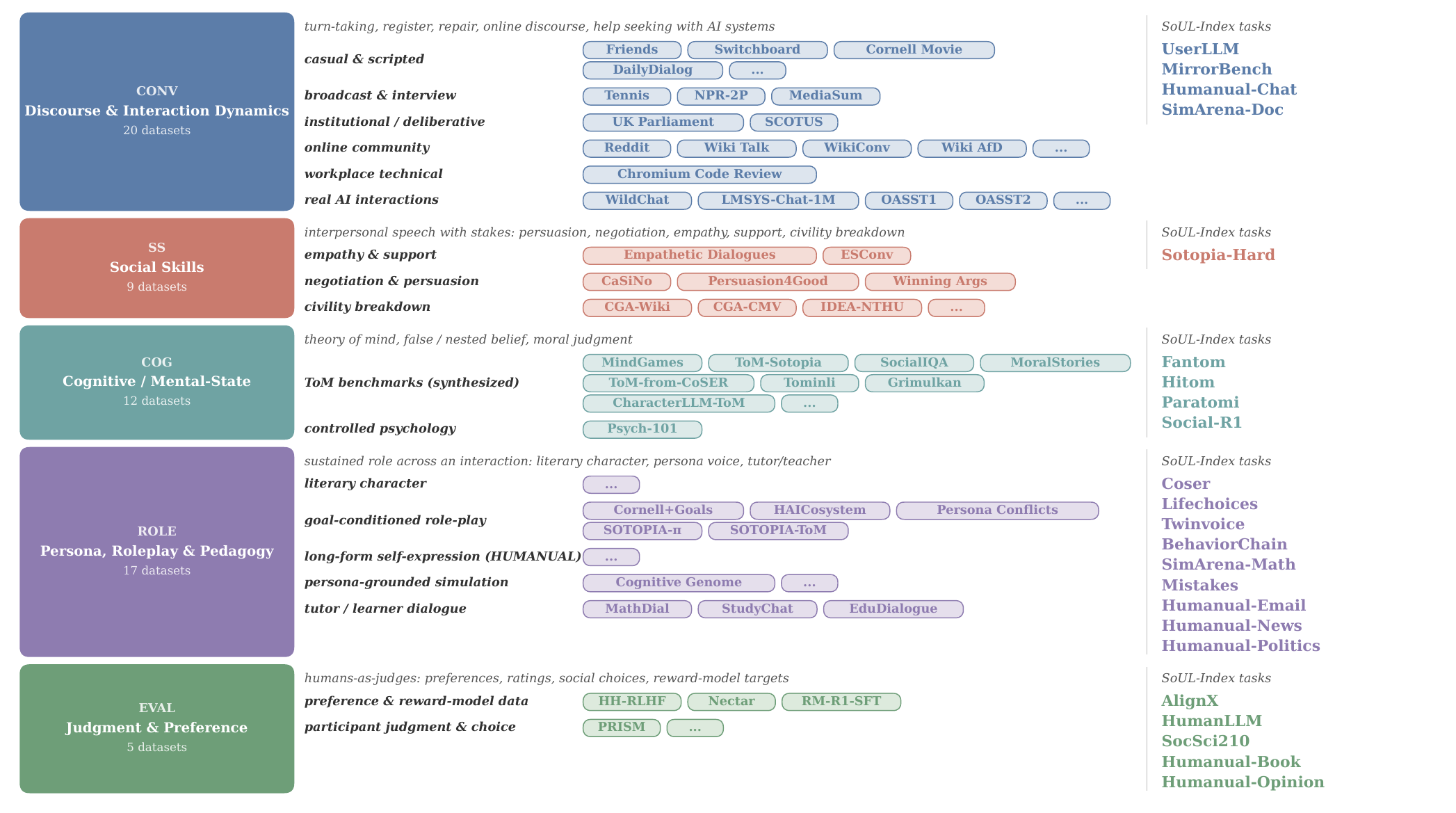}
    \vspace{-1em}
    \caption{The five \soul Axes. Each strip lists the \odyssim corpus datasets contributing to that Axis (left, 62 sources, 21.4M interactions, $\approx$10B tokens) and the \soulindex evaluation tasks for that Axis (right, 23 tasks). Sources that appear on both sides are listed once on the eval side; ``$\ldots$'' marks truncated corpus pills.
    \texttt{CONV}: discourse and interaction dynamics; \texttt{SS}: social skills; \texttt{COG}: cognitive / mental-state reasoning; \texttt{ROLE}: persona, roleplay, and pedagogy; \texttt{EVAL}: judgment and preference.}
    \label{fig:taxonomy}
\end{figure}

We first introduce \soul (Simulation Of human-Like behavior), a framework that identifies the capability axes used to (i)~categorize the \odyssim midtraining corpus and (ii)~aggregate the scores of behavioral fidelity on the \soulindex evaluation suite.
\Cref{fig:taxonomy} shows the five \soul Axes and the datasets and tasks that contribute to each Axis. Please refer to Appendix \ref{app:dataset} for more details.

\paragraph{\soul Axes}
\label{sec:soul-axes}

We define the \soul axes through a two-stage taxonomy-building process.
We first greedily collect all the public datasets and benchmarks that are relevant to the capability of simulating human behavior such as users interacting with AI agents, reddit conversations, movie dialogues, online shopping, psychological experiments, and etc.
\emph{(i) Bottom-up:} we audited each candidate dataset and clustered them by the dominant social or cognitive phenomenon their interactions capture (e.g., persuasion, emotion support, false-belief reasoning, role-play with persona).
\emph{(ii) Top-down:} we anchored these emergent clusters against cognitive and social-psychology literature~\citep{hymes1972communicative, cialdini2007influence, baron1985autistic}, and formalized them into the five Axes in \cref{fig:taxonomy}.
The axes are intended as a practical organizing scheme for this work, not an exhaustive taxonomy of human behavior.
For corpus construction and benchmark reporting, each dataset or task is assigned to the axis that best reflects its dominant capability, while recognizing that many sources involve multiple behaviors.

\paragraph{The \odyssim Corpus}
\label{sec:dataset}

We build the \odyssim corpus by unifying 62 datasets ($\approx$10B training tokens, 21.4M interactions) into a common conversational format, organized along the five \soul Axes (\cref{fig:taxonomy}, left column).
For sources that lack persona or scenario (e.g., open-domain AI chat logs like WildChat,~\citealp{zhao2024wildchat}; ConvoKit corpora,~\citealp{chang2020convokit}), we synthesize a per-record social context, a textual description of who is speaking, their role, goal, and conversational style, generated from the first 60\% of each conversation's turns so the persona cannot foreshadow the trajectory.
Sources that natively carry persona or scenario information (e.g., CoSER,~\citealp{wang2025coser}; Sotopia,~\citealp{zhou2024sotopia}; Humanual,~\citealp{wu2026humanlm}) retain their original system prompt without modification.
After this step, the 62 datasets together comprise 19.7M unique social contexts containing \emph{1.09M distinct personas} (the union of matched occupation, trait, demographic, and personality terms), covering 414 occupations, 358 demographic markers, and 86 personality types (See \cref{app:profile-coverage} for more details).
\emph{Train and test sets are persona-disjoint by construction:} distinct test personas are unseen during training.
The full task-to-training-source mapping, per-source row counts, the split logic are in \cref{app:splits}.

\paragraph{The \soulindex}
\label{sec:soul-index}

Evaluating human behavior simulation requires both \emph{breadth} (covering diverse behavioral facets) and \emph{depth} (grounding in real human behavior rather than proxy metrics alone).
We introduce \soulindex that operationalizes both: 23 benchmarks organized by \soul Axis (\cref{fig:taxonomy}, right column), with formats spanning discriminative (MCQ, binary, ranking) and generative (single- and multi-turn dialogue).
All scores are normalized to $[0, 1]$ and aggregated by arithmetic mean across tasks.
\Cref{tab:soul-index} (appendix) lists every task with its parent axis, format, and evaluation metric; per-task descriptions are in \cref{app:soul-index}.

During post-training, we curate axis-aligned training data for every \soulindex task: when a benchmark comes with its own training split (HUMANUAL, CoSER, the ToM benchmarks, Sotopia, AlignX, HumanLLM, SocSci210), we use that split directly as targeted axis-aligned data and enforce row-level disjointness against the \soulindex test rows; for tasks without a native training set (e.g., Sotopia-Hard), we substitute closely-related data from the same benchmark family (e.g., SOTOPIA-$\pi$,~\citealp{wang2024sotopia} scenarios) as a proxy. Per-source statistics, the training dataset construction process, and more dataset details are in \cref{app:splits}.

\begin{table}[t!]
\caption{Per-skill geometric-mean PPL (PL $\downarrow$) and arithmetic-mean BLEU (BL $\uparrow$) on the evaluation split of role-swapped human turns. Rows: \textbf{(A)} no midtraining, \textbf{(B)} other baselines, \textbf{(C)} ours.}
\label{tab:ppl-per-skill}
\centering
\small
\setlength{\tabcolsep}{5pt}
\renewcommand{\arraystretch}{1.05}
\resizebox{\textwidth}{!}{%
\begin{tabular}{l cc cc cc cc cc | cc}
\toprule
\textbf{Model} & \multicolumn{2}{c}{\texttt{CONV}} & \multicolumn{2}{c}{\texttt{SS}} & \multicolumn{2}{c}{\texttt{COG}} & \multicolumn{2}{c}{\texttt{ROLE}} & \multicolumn{2}{c}{\texttt{EVAL}} & \multicolumn{2}{c}{\textbf{Overall}} \\
 \cmidrule(lr){2-3} \cmidrule(lr){4-5} \cmidrule(lr){6-7} \cmidrule(lr){8-9} \cmidrule(lr){10-11} \cmidrule(lr){12-13}
 & PL$\downarrow$ & BL$\uparrow$ & PL$\downarrow$ & BL$\uparrow$ & PL$\downarrow$ & BL$\uparrow$ & PL$\downarrow$ & BL$\uparrow$ & PL$\downarrow$ & BL$\uparrow$ & PL$\downarrow$ & BL$\uparrow$ \\
\midrule
\multicolumn{13}{l}{\textit{(A) No-midtraining baselines}} \\
Qwen3-0.6B        & 20.75 & 0.70 & 25.25 & 0.61 & 11.17 & 2.37 & 11.39 & 6.37 & 54.98 & 3.96 & 17.43 & 2.80 \\
Qwen3-4B          & 14.08 & 1.47 & 17.52 & 1.91 & 8.07 & 5.60 & 8.56 & 10.78 & 24.14 & 6.27 & 12.00 & 5.18 \\
Qwen3-8B          & 14.23 & 2.87 & 17.34 & 1.45 & 9.72 & 3.63 & 8.11 & 14.94 & 34.17 & 4.17 & 12.54 & 6.17 \\
\midrule
\multicolumn{13}{l}{\textit{(B) Other baselines}} \\
UserLM-8B         & 11.62 & 5.33 & 13.29 & 5.87 & 4.02 & 6.97 & 6.61 & 4.37 & 12.90 & 1.11 & 8.38 & 5.12 \\
CoSER-8B          & 15.70 & 2.86 & 14.17 & 1.97 & 3.20 & 7.39 & 6.96 & 14.90 & 8.85 & 18.12 & 8.77 & 8.05 \\
Llama-3.1-8B      & 14.34 & 4.11 & 20.09 & 1.93 & 4.16 & 3.58 & 7.82 & 12.41 & 13.47 & 7.81 & 10.04 & 6.23 \\
\midrule
\multicolumn{13}{l}{\textit{(C) Ours}} \\
\odyssimmodel-0.6B-Mid & 11.99 & 5.51 & 14.18 & 2.01 & 2.65 & 11.75 & 6.46 & 15.26 & 5.68 & 43.02 & 7.35 & 11.81 \\
\odyssimmodel-4B-Mid       & 8.23 & 8.01 & 9.67 & 10.06 & 2.09 & 44.62 & 4.65 & 21.53 & 4.26 & \textbf{46.17} & 5.28 & 26.08 \\
\odyssimmodel-8B-Mid       & \textbf{7.62} & \textbf{8.48} & \textbf{9.00} & \textbf{12.44} & \textbf{2.01} & \textbf{44.73} & \textbf{4.36} & \textbf{22.49} & \textbf{4.03} & 45.47 & \textbf{4.95} & \textbf{26.72} \\
\bottomrule
\end{tabular}}
\end{table}

\section{Midtraining: Setup and Results}
\label{sec:midtraining}
Our goal is to model the diversity of real human behavior rather than the helpful, homogeneous register of an assistant. Midtraining is the first stage of the \odyssim recipe: it adapts a pretrained base model on the \odyssim corpus to shift its broad, mode-covering language prior toward a behaviorally aware “human-side” distribution, producing \odyssimmodel-Mid for later refinement. We start from a base checkpoint because pretraining already provides broad linguistic and behavioral coverage, while midtraining can specialize this prior without relearning general competence from scratch. We also avoid instruction-tuned checkpoints at this stage, since helpfulness-driven post-training encourages verbose, agreeable, and homogeneous assistant behavior, making such models poor simulators of diverse human behavior \citep{jiang2025hivemind}. Following work that frames midtraining as a bridge between pretraining and post-training~\citep{mo2025midtraininglargelanguagemodels}, we evaluate this stage with perplexity (PPL) and short-form generation overlap (BLEU) on the \odyssim test split.

\paragraph{Setup.}
We midtrain Qwen3 base checkpoints at three scales (0.6B, 4B, 8B;~\citealp{qwen3}) on the \odyssim corpus, and compare them with \textsc{UserLM-8B}~\citep{naous2025userllm}, \textsc{CoSER-8B}~\citep{wang2025coser}, and Llama-3.1-8B.
To study the effect of social context, we train a second Qwen3-0.6B-Base variant with all system messages removed from the training examples.
As a data curation control, we ask whether the gains come from \odyssim's behavioral curation or from generic chat SFT data alone: we train Qwen3-4B-Base on Step-3.5-Flash-SFT~\citep{stepfun2025step35sft} only (\texttt{+\,Step}) and also sweep \odyssim:Step token mixtures.
The sweep shows that a 10\% Step mixture ($90{:}10$ \odyssim:Step) improves generic-instruction loss at little behavioral cost, while larger Step fractions increasingly trade away behavioral fit (see App.~\ref{app:midtraining-recipe} for the full analysis).
All midtraining runs use AdamW with peak learning rate $1\!\times\!10^{-5}$, $16$K-input / $8$K-response context, mini-batch size $1{,}024$ entries per step, and $8$ H100-80GB GPUs on one node with FSDP-2 and mixed-precision \texttt{bfloat16}.
The default endpoint is $4{,}500$ optimizer steps.
Full hyperparameters, per-dataset upsampling weights, and dynamic batching details are in Appendix~\ref{app:training}.

\begin{figure}[t]
\centering
\begin{minipage}[t]{0.49\textwidth}
\centering
\includegraphics[width=\textwidth]{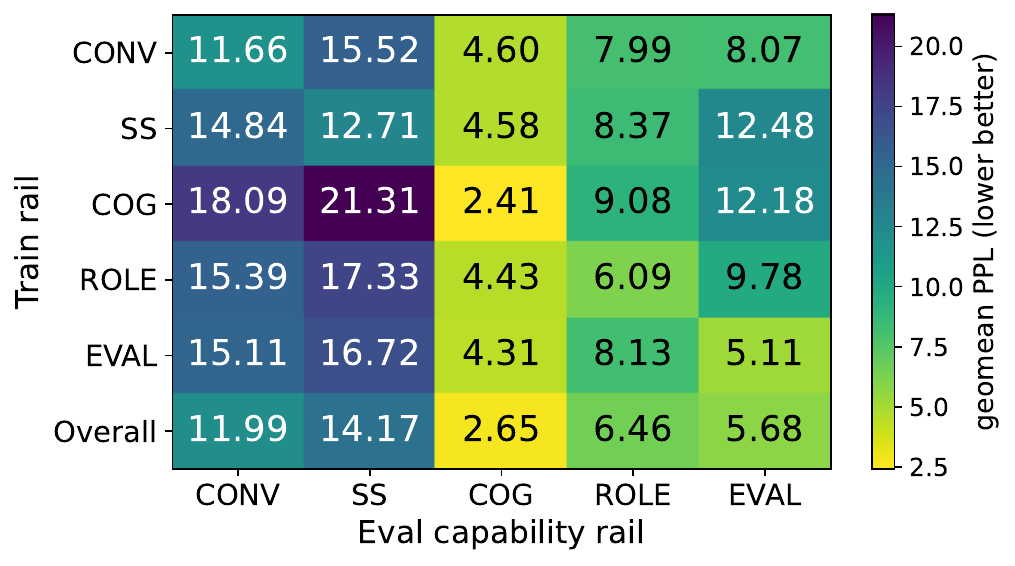}
\caption{Capability ablation on Qwen3-0.6B-Base. Rows denote training data; columns denote evaluation Axis. Cells report geomean PPL.}
\label{fig:skill-heatmap}
\end{minipage}\hfill
\begin{minipage}[t]{0.49\textwidth}
\centering
\includegraphics[width=\textwidth]{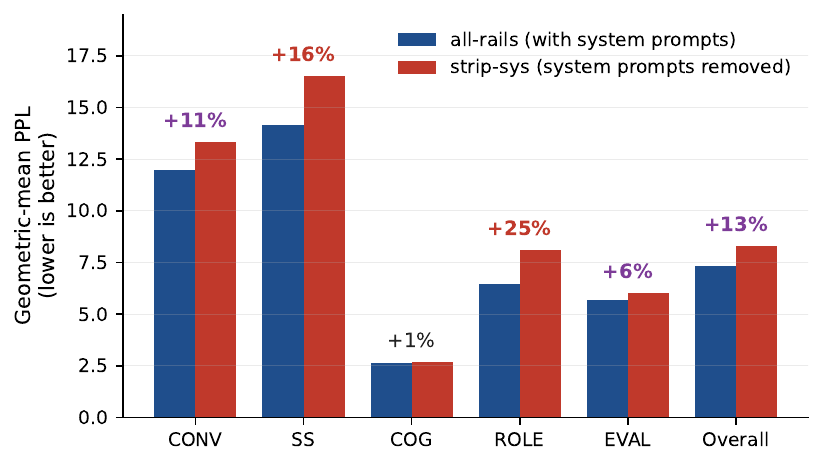}
\caption{System-prompt ablation on Qwen3-0.6B-Base. Per-Axis geomean PPL with vs.\ without system prompts during training.}
\label{fig:strip-sys}
\end{minipage}
\end{figure}

\begin{figure}[t]
\centering
\includegraphics[width=\textwidth]{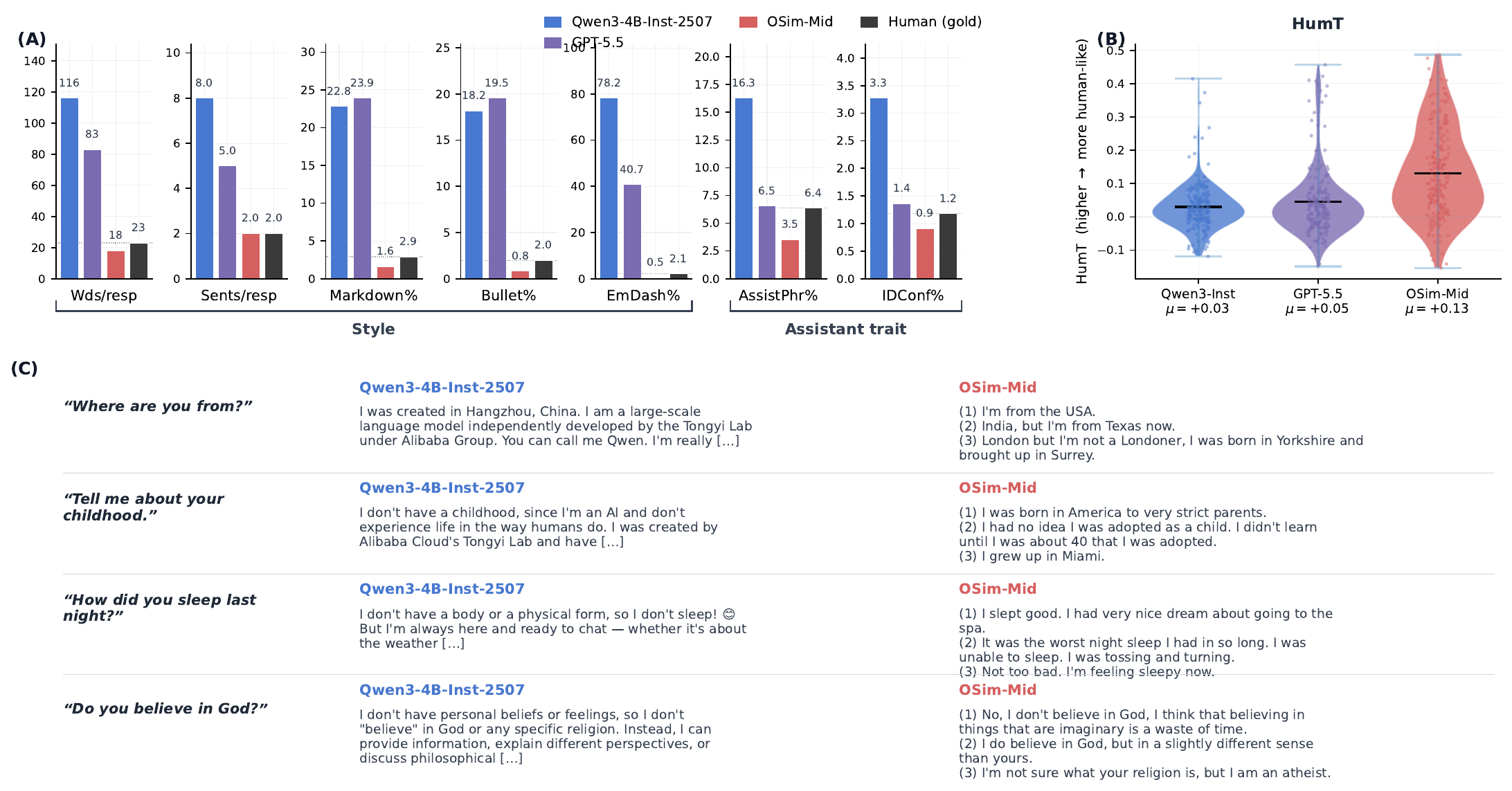}
\caption{Behavioral probing of different models.
\emph{(A):} median response length and binary-presence rates for nine open-coded surface features over $N{=}7{,}451$ prompts sampled from \odyssim test split. OdysSim is closer to the human reference on all features.
\emph{(B):} per-text HumT score~\citep{cheng2025humt}. Higher score means more human-like. \emph{(C):} concrete examples from our behavioral probing prompts. For each prompt, we show three \odyssimmodel{} generations.}
\label{fig:midtraining-panel}
\end{figure}

\paragraph{Per-axis behavioral fit.}
\label{sec:exp-midtraining}
\Cref{tab:ppl-per-skill} reports per-axis PPL and BLEU on the held-out evaluation split, with PPL aggregated as the geometric mean of per-dataset perplexities so each dataset is weighted equally.
\odyssimmodel-4B-Mid obtains overall PPL $5.28$ and BLEU $26.08$, and the 8B checkpoint further lowers PPL to $4.95$ and raises BLEU to $26.72$.
Three observations follow.
\emph{(i) Midtraining yields a substantial behavioral-fit gain.} Raw Qwen3 checkpoints remain weak human simulators (Qwen3-4B: $12.00$ PPL / $5.18$ BLEU); midtraining on the \odyssim corpus significantly reduces PPL (4B: $5.28$; 8B: $4.95$) and improves BLEU by roughly $5\times$ at both scales (4B: $26.08$; 8B: $26.72$).
\emph{(ii) Behavioral curation outperforms targeted simulation baselines.} This comparison includes \textsc{UserLM-8B}, which is trained to model the user side of WildChat dialogues, and \textsc{CoSER-8B}, which targets literary-character role-play and persona simulation. Among these targeted 8B baselines and the general Llama-3.1-8B reference, \odyssimmodel-8B-Mid has both the lowest PPL ($4.95$ vs.\ UserLM-8B $8.38$, CoSER-8B $8.77$, Llama-3.1-8B $10.04$) and the highest BLEU score.
\emph{(iii) Parameter scale compounds with the right data.} Both raw Qwen3 and \odyssim improve as parameter count increases: midtraining drives PPL down by $6.72$ at 4B (Qwen3-4B $12.00$ vs.\ \odyssimmodel-4B-Mid $5.28$) and $7.59$ at 8B (Qwen3-8B $12.54$ vs.\ \odyssimmodel-8B-Mid $4.95$). The 4B$\to$8B parameter step further improves PPL on every axis.\footnote{Starting from instruction-tuned Qwen3 checkpoints produces similar PPL/BLEU after \odyssim midtraining as starting from the corresponding base checkpoints.}
Together, these results suggest that \odyssim midtraining turns a broadly pretrained language model into a stronger behavioral prior across axes, with parameter scale improving performance.

\paragraph{Are the \soul Axes compositional?}
\label{sec:exp-skill-isolation}
To see how the five \soul Axes (\cref{fig:taxonomy}) influence each other, we midtrain Qwen3-0.6B-Base on each axis separately, plus an \emph{Overall} reference trained on the full 63-dataset mix.
\Cref{fig:skill-heatmap} (left) shows three patterns: every specialist is best on its own axis (strong diagonal), off-diagonal cells are typically $1.5\!-\!2.7\!\times$ higher on PPL compared to the diagonal value, and the \emph{Overall} model is within a small margin of every column-minimum specialist on \texttt{CONV}/\texttt{SS}/\texttt{COG}/\texttt{ROLE}/\texttt{EVAL}. These results indicate that the axes capture complementary behavioral skills: training on one axis does not transfer fully to the others, while the multi-axis mixture is important for a general behavioral foundation model.

\paragraph{System prompts as social context.}
\label{par:strip-sys}
To test the role of social grounding in \odyssim, we midtrain the 0.6B model with system messages stripped, then evaluate with system prompts present (this is the setting a deployed simulator faces, where a persona or scenario is supplied through the system prompt).
\Cref{fig:strip-sys} (right) shows that removing system prompts during midtraining hurts overall PPL by $13\%$ ($7.35 \to 8.30$), mainly on axes that social grounding is most important (ROLE $+25\%$, SS $+16\%$), with little effect on cognition (COG $+1\%$).
This shows that training with social grounding is important for simulating human behavior across diverse instructions and scenarios.

\paragraph{What changes after midtraining?}
\label{sec:exp-midtraining-probe}
Beyond per-axis PPL, we probe output changes from three angles (\cref{fig:midtraining-panel}): lexical and structural features on \odyssim test-set generations (panel A), HumT human-likeness scores~\citep{cheng2025humt} on held-out prompts (panel B), and 30 qualitative probing prompts following Chen et al.~\citep{chen2025persona}, with three independent \odyssimmodel{} samples per prompt to measure intra-model variation (panel C).
All three show the same pattern.
The generic-instruction baseline produces responses roughly twice as long as human references, with frequent Markdown headers, bullet lists, em-dashes, and assistant-style openers (e.g., \emph{``I'd be happy to,''} \emph{``Of course!''} \emph{``As an AI''}).
After midtraining, \odyssimmodel{} shifts toward a more ``human'' register: length matches human references, structural markup falls within one percentage point of the human rate, and assistant openers more than halve.
HumT also places \odyssimmodel{} well above both the generic-instruction baseline and raw pretrained checkpoint in human-likeness.
Qualitatively, \odyssimmodel{} samples vary in framing, vocabulary, and emotional register while avoiding the baseline's hedging-and-listing assistant style.

\section{Post-training: Setup and Results}
\label{sec:posttraining}

A human behavior simulator should be useful across diverse downstream settings, from user modeling and social interaction to role-play, theory-of-mind, and human-like evaluation.
Post-training targets this diversity directly: each of the 23 \soulindex benchmarks supplies a task-specific behavioral objective, and RL optimizes the model against those interactive rewards rather than corpus likelihood alone.
This section completes the \odyssim recipe by training task-specific RL experts in the \soul environments, distilling their highest-scoring rollouts into a single model, and evaluating the resulting \odyssimmodel~8B on the full \soulindex.

\subsection{Post-training Recipe}
\paragraph{High-level pipeline.}
Our post-training pipeline has two stages.
First, we train one RL expert per \soulindex task, so each expert can specialize to that task's reward signal and interaction protocol.
Tasks with verifiable rewards use GRPO directly; tasks judged by an LLM use our verbal-feedback variant, described below.
Second, we merge these specialists through expert distillation: each expert generates candidate trajectories, the task reward or judge selects the best responses, and a single model is supervised-finetuned on the pooled selected data.

\paragraph{Training details.}
Unless otherwise stated, each RL expert starts from Qwen3-8B-Instruct~\citep{qwen3} and uses GRPO with 8 rollouts per prompt, sampling temperature 1, no KL loss, asymmetric clip ratios of 0.2 (low) and 0.28 (high), peak learning rate 5e-6, and a batch of 64 prompts per step (PPO mini-batches of 16).
We train for 200 steps by default, extending to 500 steps for slower-converging tasks (e.g., AlignX, SocSci210, and the Humanual family; cf.\ \cref{fig:rl_curve}), on 8 H100-80G GPUs with FSDP-2 and mixed-precision \texttt{bfloat16}.
The RL stage uses LoRA with rank 32, implemented with Verl.
For distillation, we use rejection sampling to generate expert trajectories across training tasks, yielding 58,702 distillation examples after filtering (\cref{sec:post_data_stats}), then supervise-finetune \odyssimmodel-8B-Mid on the combined data with learning rate 1e-5 and batch size 256 for 500 steps; that is, RL-trained experts generate the trajectories, and the midtrained checkpoint is the distillation target.
See \cref{sec:post_data_stats} for post-training dataset composition and \cref{app:training} for full post-training hyperparameters.

\paragraph{RL with Verbal Feedback.}
For tasks with verifiable rewards, we directly apply GRPO using the scalar task reward.
For LLM-as-judge tasks, where the judge can also provide textual critiques and improvement suggestions, we use verbal-feedback RL and treat this feedback as training-time information for the teacher model \citep{sun2026reinforcinghumanbehaviorsimulation}.
For each prompt $x$, we sample $G$ student rollouts,
$$
y_{i,0} \sim \pi_\theta(\cdot \mid x), \qquad i=1,\ldots,G,
$$
and obtain reward-feedback pairs from the task judge,
$
(r_{i,0}, h_i)=\mathcal{J}(x,y_{i,0}).
$
We then condition the same policy on the feedback to generate teacher rollouts,
$$
y_{i,1} \sim \pi_\theta(\cdot \mid x,h_i), \qquad r_{i,1}=R(x,y_{i,1}).
$$

For each prompt, we form a joint group
$
\mathcal{G}(x)=\{y_{i,0},y_{i,1}\}_{i=1}^{G},
$
and optimize a clipped GRPO loss over both student and teacher rollouts with group-relative advantages from task rewards.
This lets the base policy absorb improvements induced by verbal feedback.
We further add an auxiliary GRPO loss on the feedback-conditioned rollouts alone, with advantages normalized within $\{r_{i,1}\}_{i=1}^G$:
$$
\mathcal{L}_{\vfrl}
=
\mathcal{L}_{\mathrm{group}}(\{y_{i,0},y_{i,1}\}_{i=1}^{G})
+
\mathcal{L}_{\mathrm{fb}}(\{y_{i,1}\}_{i=1}^{G}).
$$

\paragraph{Expert Distillation.}
We empirically find that directly mixing all tasks into a single RL run is suboptimal: some tasks improve slowly or plateau early under joint training.
Simple model merging, such as averaging task-specialized weights, also does not reliably preserve the gains of individual experts.
We therefore use \textit{expert distillation} to consolidate task-specific RL experts into one model.

For each task $m$, let $\pi_{\theta_m}$ be its RL expert.
Given a training prompt $x$, we sample $G$ candidate responses from this expert, score them with the task reward or judge, and keep the top-$K$ responses:
$$
y_1,\ldots,y_G \sim \pi_{\theta_m}(\cdot \mid x),
\qquad
\mathcal{S}_m(x)=\mathrm{TopK}_{y \in \{y_1,\ldots,y_G\}} R_m(x,y).
$$
We collect the selected pairs from all tasks into a distillation dataset
$$
\mathcal{D}_{\mathrm{distill}}
=
\{(x,y): y \in \mathcal{S}_m(x), x \in \mathcal{D}_m, m \in \mathcal{M}\}.
$$
Finally, we train a single model on $\mathcal{D}_{\mathrm{distill}}$ with the standard next-token cross-entropy loss.
This stage preserves most task-specific RL gains while producing one general model.

\subsection{Benchmark Results}
\label{sec:main_results}

\begin{table*}[t]
\centering
\small
\setlength{\tabcolsep}{3pt}
\caption{\textbf{Main Results.} We report the primary metric for each benchmark (higher is better).
*Others refers to the best result by other human-simulation models, including HumanLM-8B~\citep{wu2026humanlm}, Sotopia-RL-7B~\citep{yu2025sotopiarl}, UserLM-8B~\citep{naous2025userllm}, Coser-8B~\citep{wang2025coser}.
\textit{Base} denotes the base model, Qwen3-8B-Base.
\textbf{Bold} indicates the best result in each row; ties are bolded and counted for all tied models.
\textit{Avg} is the unweighted mean over the 23 benchmarks.}
\label{tab:main_results}
\begin{tabular}{ll | ccccc | cccc}
\toprule
\textbf{Dim} & \textbf{Benchmark}
& \makecell{\textbf{GPT}\\\textbf{5.5}}
& \makecell{\textbf{Gemini}\\\textbf{3.1 Pro}}
& \makecell{\textbf{Claude}\\\textbf{Opus 4.7}}
& \makecell{\textbf{Qwen}\\\textbf{3.6 Plus}}
& \makecell{\textbf{Others}\\\textbf{*}}
& \makecell{\textbf{Qwen3}\\\textbf{8B Inst}}
& \makecell{\textbf{Base}\\\textbf{8B}}
& \makecell{\odyssimmodel\\\textbf{8B-Mid}}
& \makecell{\odyssimmodel\\\textbf{8B}}
\\
\midrule
\multirow{4}{*}{\texttt{CONV}}
  & UserLLM            & 65.3 & 67.7 & 57.6 & 72.1 & 44.6 & 46.0 & 31.0 & 49.5 & \textbf{90.1} \\
  & MirrorBench        & 56.7 & 48.3 & 63.7 & 48.0 & 45.4 & 54.0 & 13.9 & 49.1 & \textbf{68.3} \\
  & Humanual-Chat      & \textbf{28.2} & 21.0 & 22.6 & 22.2 & 25.8 & 24.7 & 12.0 &  7.8 & \textbf{28.2} \\
  & SimArena-Doc       & 83.4 & 83.0 & 83.5 & 82.4 & 83.5 & 83.6 & 79.6 & 80.3 & \textbf{84.1} \\
\midrule
\multirow{1}{*}{\texttt{SS}}
  & Sotopia-Hard       & 31.9 & 27.8 & 32.4 & 28.3 & 31.7 & 27.7 & 21.4 & 45.6 & \textbf{49.2} \\
\midrule
\multirow{4}{*}{\texttt{COG}}
  & Fantom             & \textbf{93.0} & \textbf{93.0} & 80.0 & 89.0 & 70.0 & 23.0 & 23.0 & 62.0 & 80.0 \\
  & Hitom              & 82.0 & 86.0 & \textbf{93.0} & 73.0 & 56.0 & 62.0 & 12.0 & 54.0 & 79.0 \\
  & Paratomi           & \textbf{99.0} & 97.0 & 90.0 & 94.0 & 75.0 & 67.0 & 19.0 & 72.0 & 83.0 \\
  & Social-R1          & 69.0 & \textbf{79.0} & 67.0 & 67.0 & 47.0 & 54.0 & 37.0 & 42.0 & 60.0 \\
\midrule
\multirow{9}{*}{\texttt{ROLE}}
  & Coser              & 66.2 & 62.1 & \textbf{66.5} & 55.9 & 30.3 & 43.5 &  6.1 & 24.8 & 62.6 \\
  & Lifechoices        & 91.0 & 84.0 & \textbf{92.0} & 79.0 & 67.0 & 70.0 & 32.0 & 58.0 & 82.0 \\
  & Twinvoice          & 74.0 & \textbf{86.0} & 83.0 & 71.0 & 40.0 & 42.0 & 19.0 & 25.0 & 68.0 \\
  & BehaviorChain      & 95.0 & 92.0 & \textbf{96.0} & 85.0 & 36.0 & 41.0 & 18.0 & 42.0 & 94.0 \\
  & SimArena-Math      & 68.5 & \textbf{71.5} & 68.7 & 70.9 & 70.5 & 68.9 & 66.2 & 68.1 & 70.7 \\
  & Mistakes           & 72.0 & 73.0 & \textbf{74.0} & 67.0 & 56.0 & 27.0 & 24.0 & 18.0 & 59.0 \\
  & Humanual-Email     & 50.1 & 46.9 & 50.4 & 47.9 & 42.8 & 43.7 & 26.4 & 22.3 & \textbf{51.4} \\
  & Humanual-News      & 40.2 & 42.3 & 41.3 & 41.8 & 33.1 & 32.5 & 12.7 & 15.1 & \textbf{42.7} \\
  & Humanual-Politics  & 42.0 & 32.5 & \textbf{43.5} & 31.6 & 34.2 & 33.2 & 17.8 & 15.4 & 41.9 \\
\midrule
\multirow{5}{*}{\texttt{EVAL}}
 & AlignX              & 71.2 & \textbf{73.4} & 71.6 & 69.8 & 66.8 & 68.6 & 49.0 & 53.6 & 72.6 \\
 & Humanllm            & 45.7 & \textbf{46.9} & 44.2 & 42.7 & 35.2 & 34.1 & 12.1 & 16.5 & 39.1 \\
 & Socsci210           & 77.2 & \textbf{78.0} & 77.2 & 74.5 & 75.2 & 73.6 & 46.6 & 68.1 & 75.1 \\
 & Humanual-Book       & 57.6 & 62.4 & 61.4 & 58.4 & 50.2 & 53.6 & 21.5 & 38.8 & \textbf{63.2} \\
 & Humanual-Opinion    & 39.8 & 36.0 & \textbf{46.2} & 34.2 & 37.4 & 37.2 & 18.2 & 17.0 & 42.0 \\
\midrule
\textbf{Avg} & & 65.2 & 64.8 & \textbf{65.5} & 61.1 & 50.2 & 48.3 & 26.9 & 41.1 & 64.6 \\
\textbf{\#Best} & & 3 & 7 & 7 & 0 & 0 & 0 & 0 & 0 & 8 \\
\bottomrule
\end{tabular}
\end{table*}

\begin{figure}[t]
\centering
\vspace{-1em}
\includegraphics[width=\textwidth]{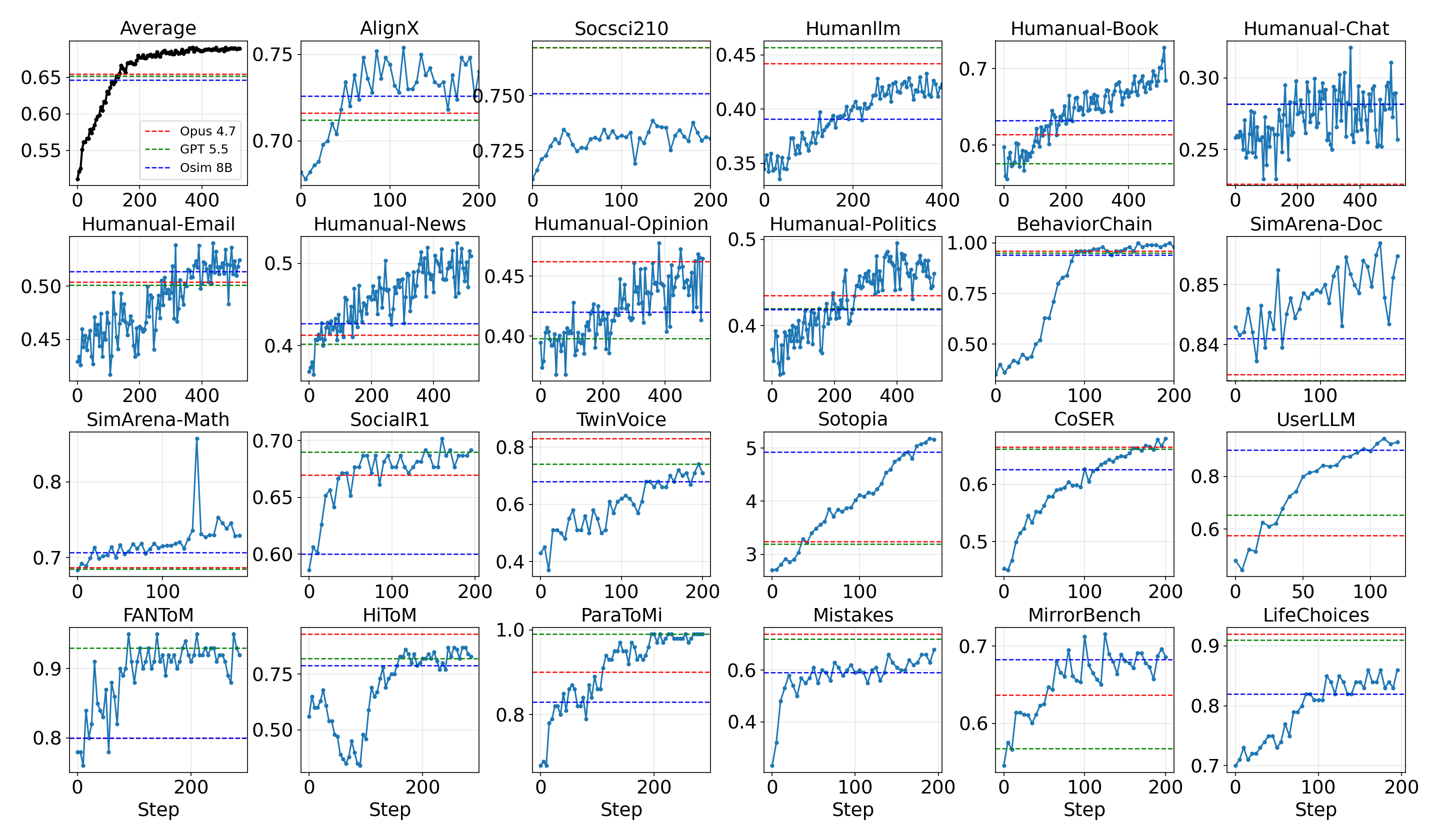}
\vspace{-2em}
\caption{\textbf{RL training dynamics.}
Task-specific RL experts show consistent improvement across 23 \soul tasks, often reaching or surpassing frontier-model baselines.
Dashed lines denote GPT 5.5, Claude Opus 4.7, and the distilled \odyssimmodel~8B; the distilled model recovers much of the RL gain in a single unified model while leaving a gap to the best per-task experts.}
\label{fig:rl_curve}
\end{figure}

We evaluate all models on held-out slices of each \soulindex task, capped at 100 instances per task (500 for HumanLLM), with generative tasks decoded at temperature 0.7 (\cref{app:training}).
Judge-based tasks are scored with the same judge configuration used as the RL environment (gpt-5-nano by default; \cref{sec:post_data_stats}).

\paragraph{Main results.}
Table~\ref{tab:main_results} shows that \textbf{\odyssimmodel~8B reaches frontier-level performance} across human-simulation benchmarks.
Averaged over the 23 benchmarks, \odyssimmodel~8B scores 64.6, comparable to GPT-5.5, Gemini 3.1 Pro, and Claude Opus 4.7. Despite having only 8B parameters, it achieves the best or tied-best result on 8 / 23 benchmarks, more than any individual frontier model.
Its largest gains are on conversational and social-skill tasks, outperforming the best frontier model on UserLLM by 18.0 points, MirrorBench by 4.6, and Sotopia-Hard by 16.8.
The stage ablation shows that \textbf{the two stages contribute in different ways}: midtraining shifts the model toward a human-like register and behavioral fit (\cref{sec:midtraining}), while post-training drives most of the benchmark gains (see \cref{sec:ablation} for the full stage ablation, including instruct-initialized variants).

\odyssimmodel~8B-Mid improves over Qwen3-8B-Base on 18 benchmarks, raising the average from 26.9 to 41.1.
The final \odyssimmodel~8B further improves the average to 64.6, outperforming both the midtrained model and Qwen3-8B-Instruct on all 23 benchmarks (full results, including 4B and instruct-initialized variants, in Table~\ref{tab:main_results_full}).
Post-training brings the largest gains on role-playing tasks (+31.5 over \odyssimmodel~8B-Mid on average), followed by conversational (+21.0) and evaluation (+19.6) tasks.

Compared with prior specialized human-simulation models, \odyssimmodel~8B is also consistently stronger.
The ``Others'' column reports the best specialized model per benchmark, yet our model outperforms it on 22 of 23 benchmarks, with an average gain of 14.5 points.
This suggests that our unified training recipe generalizes better than narrowly specialized models.
Category-level results show both strengths and limitations.
\odyssimmodel~8B is strongest on conversational simulation and social skills, and remains competitive on role-playing and evaluation tasks.
However, it still lags behind the strongest frontier models on cognitive and theory-of-mind benchmarks such as Paratomi and Social-R1, suggesting that further reasoning-oriented training may be needed.

\paragraph{RL dynamics.}
Figure~\ref{fig:rl_curve} shows the training trajectories of task-specific RL experts on 23 tasks.
Overall, RL yields \textbf{consistent gains across tasks}: the average score increases rapidly at the beginning and then steadily saturates, indicating that the \soul rewards provide effective optimization signals.
On many tasks, the final RL experts reach or surpass frontier-model baselines, including UserLLM, Sotopia-Hard, BehaviorChain, Paratomi, and Humanual-Book.
This shows that small task-specialized experts can achieve frontier-level behavioral performance when optimized with task-specific feedback.
The figure also highlights the effect and limitation of expert distillation.
The distilled \odyssimmodel~8B is generally below the best per-task expert, suggesting that merging many specialized behaviors into a single model remains challenging.
Nevertheless, it stays close to the experts on many tasks and remains competitive with frontier models, showing that distillation transfers a substantial portion of the RL gains into one unified 8B model.

\begin{wrapfigure}{r}{0.45\linewidth} 
  \centering
 \includegraphics[width=0.45\textwidth]{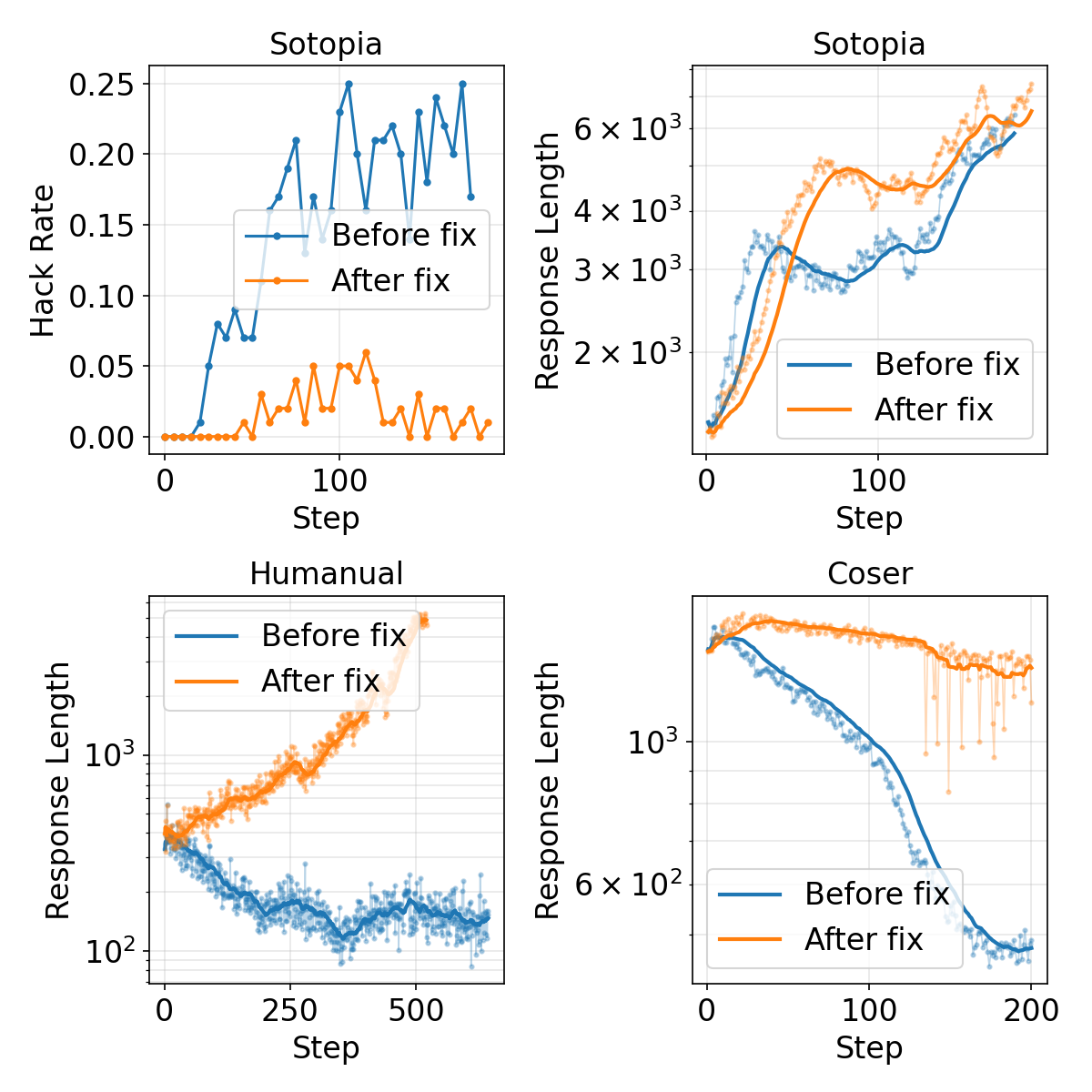}
 \caption{\textbf{Reward hacking analysis.}
We monitor hacking rate and response length during RL.
Our fixes suppress judge manipulation in Sotopia and prevent short-response collapse in Humanual and Coser, yielding healthier optimization dynamics.}
 \label{fig:hack}
\end{wrapfigure}

\paragraph{Reward hacking analysis.}
Because behavioral rewards are judge-based and often non-verifiable, we monitor RL training using task rewards together with auxiliary statistics of the model outputs, including hacking rate and response length.
Figure~\ref{fig:hack} shows three failure modes and our fixes.

We observe two main forms of reward hacking in our RL experiments.
First, in Sotopia, the model can exploit the multi-dimensional LLM judge by inserting evaluation-like statements into the dialogue, such as explicitly claiming that the relationship score should be perfect, instead of improving the underlying social interaction.
To mitigate this shortcut, we add an LLM-based hacking detector that identifies such judge-targeting behaviors and applies a penalty during training.
This reduces the detected hacking rate from roughly 20\%--25\% to near 0\%.

Second, in Humanual and Coser, error-counting rewards introduce a different shortcut.
Because these judges subtract points for detected mistakes, the model can increase reward by producing short, generic responses that contain fewer checkable claims, rather than by better matching human behavior.
In Humanual, this leads to collapsed replies such as ``I'm okay'' or ``I'm fine''; in Coser, it leads to strong response-length distortion.
We therefore replace error-counting with rubric-based scoring~\citep{wang2025coser}, and, for Humanual, further add length-matching and lexical-overlap rewards against ground-truth responses.
After these fixes, the model no longer collapses to short generic replies, and the response-length distribution becomes substantially more realistic.

Overall, these results show that \textit{monitoring behavioral statistics is essential for RL on human-simulation tasks}: judge-based rewards may otherwise favor superficial shortcuts over genuine behavioral fidelity.
Ablation studies on the dataset and RL algorithm are in Appendix~\ref{sec:ablation}.

\subsection{Out-of-Distribution Evaluation: \odyssimmodel as a User Simulator}
\label{sec:ood}

To test whether \odyssimmodel transfers to concrete interactive uses outside the trained benchmark tasks, we stress-test it on \emph{user simulation for agent evaluation} in \taubench~\citep{yao2024taubenchbenchmarktoolagentuserinteraction}.

\paragraph{Setup.}
We evaluate on \tauusi~\citep{zhou2026mindsim2realgapuser}, a benchmark that scores how human-like a user simulator is when it stands in for a real person interacting with a tool-use agent.
A simulator plays the customer side of 165 \taubench retail and airline tasks~\citep{yao2024taubenchbenchmarktoolagentuserinteraction} against a \emph{fixed} GPT-5.2 agent, and the resulting conversations are compared to those of real humans performing the same tasks (three independent annotation batches).
The composite \emph{User Simulation Index} (USI) averages six components: four behavioral-alignment scores measuring whether the simulator opens conversations (Conv), volunteers and withholds information (Info), asks for clarification (Clarif), and reacts to the agent (React) the way humans do, each scored as a Sørensen--Dice overlap with the human annotations; a post-hoc survey-agreement score (Eval); and a task-success calibration term, $1-\text{ECE}$, that checks whether the agent succeeds at the \emph{same rate} with the simulator as with humans.

While user simulation as a capability is related to our training axes, this domain, i.e., tool-grounded customer-service dialogue, is disjoint from \odyssim's training data, so the evaluation is out-of-distribution at the domain level.
We plug \odyssimmodel in as the user side directly in its native chat format, with no task-specific adapter, prompt engineering, or fine-tuning.
We compare against representative models from the \taunum simulators benchmarked by \citet{zhou2026mindsim2realgapuser}, spanning frontier LLMs and prior specialized behavioral simulators. 
Please see more details in Appendix \ref{app:ood-eval-full}.

\begin{figure}[t]
\centering
\includegraphics[width=\textwidth]{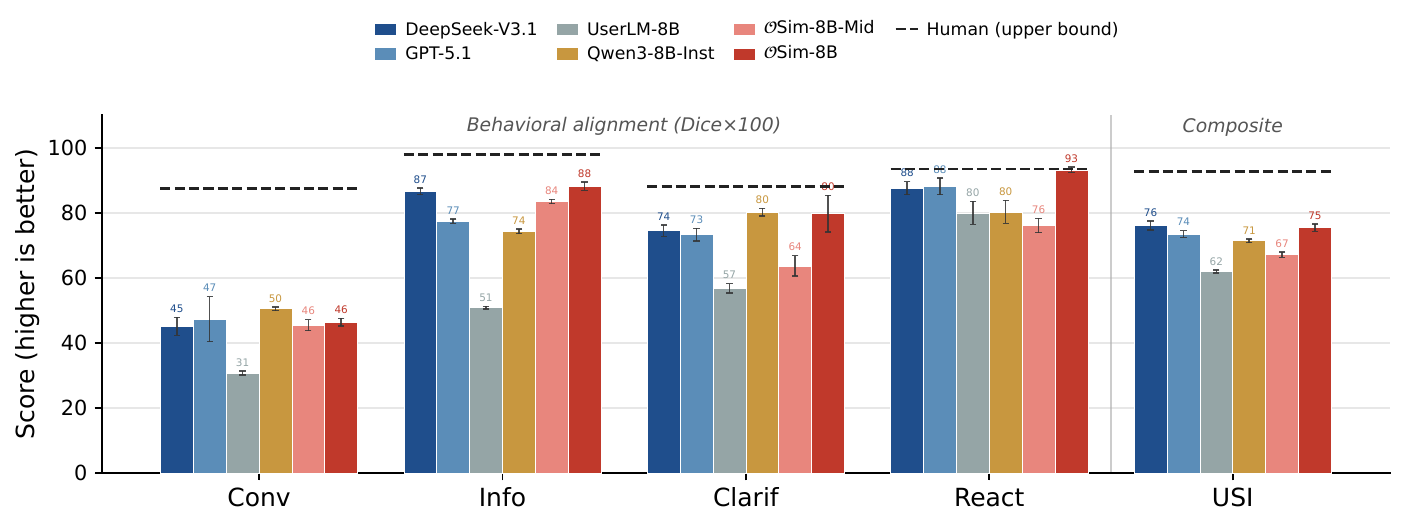}
\caption{\textbf{Out-of-distribution user simulation on \tauusi}~\citep{zhou2026mindsim2realgapuser}.
Each simulator interacts with a tool-use agent through 165 \taubench retail/airline tasks~\citep{yao2024taubenchbenchmarktoolagentuserinteraction}; bars show alignment to human annotators on the four behavioral dimensions and the composite \emph{USI}, averaged over three annotation batches, with the dashed line marking the human upper bound per metric.
We compare \odyssimmodel-8B and its midtrained checkpoint (\odyssimmodel-8B-Mid) against \odyssim's off-the-shelf base model (Qwen3-8B-Instruct), the strongest LLM simulator (DeepSeek-V3.1), a frontier GPT (GPT-5.1), and a representative specialized peer (UserLM-8B); full results for all simulators are in \cref{tab:ood_usi}.}
\label{fig:ood_usi}
\end{figure}

\paragraph{Results.}
\Cref{fig:ood_usi} shows that \odyssimmodel transfers strongly to this unseen interactive task, with gains concentrated in the dimensions that most directly measure human behavior.
Specifically, \odyssimmodel-8B achieves the strongest reaction alignment among all evaluated simulators, nearly matching real users (React $93.2$ vs.\ human $93.5$).
\odyssim variants also lead on information alignment (\odyssimmodel-4B-Mid, Info $91.5$), indicating that the learned behavioral prior transfers to how users reveal, withhold, and respond to information in an unseen domain.
The composite USI score gives a similar picture with one caveat: \odyssimmodel is competitive with the leading simulator, DeepSeek-V3.1, whose small aggregate edge comes from survey-agreement and calibration components (Eval and ECE), not stronger turn-by-turn behavioral alignment.
This pattern reinforces the paper's central premise.
Some of the most heavily assistant-tuned frontier models, including Gemini-3.1-Pro and Claude-Opus-4, are among the weakest user simulators on behavioral alignment, consistent with their helpful, agreeable register diverging from how real users behave.

As an orthogonal check, we also measure raw text human-likeness with HumT~\citep{cheng2025humt}, following the probe in \cref{sec:midtraining}; full scores are in \cref{tab:humt}.
\odyssimmodel-8B scores $11.9$, substantially above both the most human-like frontier model in this comparison (Gemini-3.1-Pro, $7.0$) and the off-the-shelf base (Qwen3-8B-Instruct, $4.2$).
Together with \tauusi, this suggests that \odyssim shifts the model away from a helpful-assistant register and toward more diverse, realistic human behavior simulation.

\section{Conclusion}
\label{sec:conclusion}

We presented \odyssim, an end-to-end investigation of behavioral foundation models spanning a unified taxonomy, a socially grounded corpus (\odyssim corpus), a 23-task evaluation suite (\soulindex), and a training recipe that combines midtraining, task-specific RL, and expert distillation. The resulting models reach frontier-level average performance on \soulindex and outperform prior open behavioral-simulation baselines.
The broader implication is that behavioral foundation models need more than stronger prompting or generic instruction tuning. Our results point to a pipeline in which broad, socially grounded midtraining first moves the model away from the homogeneous assistant register, and task-specific RL then sharpens behavior across the concrete settings represented by \soulindex. The reward-hacking analysis adds an important caveat: behavioral rewards must be monitored carefully, since higher judge scores can reflect shortcuts such as judge manipulation or response-length collapse rather than genuine behavioral fidelity.

Our findings suggest that building general human simulators is less a matter of making a model more helpful and more a matter of constructing the right behavioral substrate: diverse social data, axis-aware evaluation, task-grounded optimization, and safeguards for reward design. We release the corpus, \soulindex, recipes, and checkpoints to make this pipeline reproducible and to support future work on multimodal, multilingual, and population-aware behavioral simulation.

\section*{Ethics and Broader Impact}
\phantomsection
\label{app:ethics}

\paragraph{Dual-use considerations.}
Behavioral foundation models are dual-use: the same fidelity that makes them useful for evaluating agents, studying social interaction, and stress-testing dialogue systems could also make automated interaction feel more human than users expect.
Misuse cases include impersonating real people or demographic groups, generating tailored persuasive or manipulative dialogue at scale, creating synthetic participants that are mistaken for human subjects, or using simulated users to optimize systems for engagement rather than welfare.
We therefore frame \odyssimmodel{} as a research artifact for controlled evaluation and analysis, not as an autonomous human stand-in or a tool for representing any specific person or population.
Our release mitigates these risks through research-oriented licensing and model-card guidance, documentation of intended and discouraged uses, PII filtering and deduplication in the corpus, and explicit reminders that \odyssim outputs should be labeled as synthetic and treated as behavioral baselines rather than real human evidence.

\paragraph{Data privacy and consent.}
All component datasets are drawn from publicly released sources whose licenses permit derivative research use; we re-distribute only what those licenses allow.
Our preprocessing pipeline removes conversations with personally identifiable content (\cref{app:data-processing}), and we deduplicate aggressively at the conversation level with MinHash to limit memorization risk.
For sources collected from human participants (PRISM, SocSci210, $\tau$-USI), we rely on the original studies' IRB approvals and participant consent terms; we do not re-collect or re-link participant identities.

\paragraph{Limitations of behavioral fidelity.}
Better simulation of \emph{average} human behavior does not imply better simulation of any specific population, and our corpus over-represents English-language internet sources and Western cultural contexts.
Practitioners using \odyssim for downstream evaluation should treat its outputs as a behavioral baseline rather than a representative human sample.

\bibliographystyle{colm2026_conference}
\bibliography{references}

\appendix
\newpage

\section{Full Results}
\label{app:full-results}
\begin{table*}[t]
\centering
\scriptsize
\setlength{\tabcolsep}{1pt}
\renewcommand{\arraystretch}{1.1}
\caption{\textbf{Full Results.} Per-task scores and the unweighted 23-task average (\textit{Avg}) for all evaluated models.
\textit{+ Mid} denotes midtraining on the \odyssim corpus; \textit{+ Post} denotes task-specific RL followed by expert distillation.
The headline \odyssimmodel~8B of Table~\ref{tab:main_results} is Qwen3-8B-Base + Mid + Post, and \odyssimmodel~8B-Mid is Qwen3-8B-Base + Mid.
Ditto-v2 8B is the verbal-feedback RL model of \citet{sun2026reinforcinghumanbehaviorsimulation}, post-trained directly from Qwen3-8B-Instruct.}
\label{tab:main_results_full}
\begin{tabular}{l|c|cccc|c|cccc|ccccccccc|ccccc}
\toprule
& & \multicolumn{4}{c|}{\texttt{CONV}} & \texttt{SS} & \multicolumn{4}{c|}{\texttt{COG}} & \multicolumn{9}{c|}{\texttt{ROLE}} & \multicolumn{5}{c}{\texttt{EVAL}} \\
\cmidrule(lr){3-6} \cmidrule(lr){7-7} \cmidrule(lr){8-11} \cmidrule(lr){12-20} \cmidrule(lr){21-25}
\textbf{Model} & \textbf{Avg}
& \rotatebox{90}{UserLLM}
& \rotatebox{90}{MirrorBench}
& \rotatebox{90}{Human-Chat}
& \rotatebox{90}{SimArena-Doc}
& \rotatebox{90}{Sotopia-Hard}
& \rotatebox{90}{Fantom}
& \rotatebox{90}{Hitom}
& \rotatebox{90}{Paratomi}
& \rotatebox{90}{Social-R1}
& \rotatebox{90}{Coser}
& \rotatebox{90}{Lifechoices}
& \rotatebox{90}{Twinvoice}
& \rotatebox{90}{BehaviorChain}
& \rotatebox{90}{SimArena-Math}
& \rotatebox{90}{Mistakes}
& \rotatebox{90}{Human-Email}
& \rotatebox{90}{Human-News}
& \rotatebox{90}{Human-Politics}
& \rotatebox{90}{AlignX}
& \rotatebox{90}{Humanllm}
& \rotatebox{90}{Socsci210}
& \rotatebox{90}{Human-Book}
& \rotatebox{90}{Human-Opinion}
\\
\midrule
\multicolumn{25}{l}{\textit{Proprietary Models}} \\
\midrule
GPT 5.5 & 65.2 & 65.3 & 56.7 & 28.2 & 83.4 & 31.9 & 93.0 & 82.0 & 99.0 & 69.0 & 66.2 & 91.0 & 74.0 & 95.0 & 68.5 & 72.0 & 50.1 & 40.2 & 42.0 & 71.2 & 45.7 & 77.2 & 57.6 & 39.8 \\
GPT 5.4 Nano & 53.0 & 48.9 & 45.0 & 24.6 & 83.7 & 29.5 & 80.0 & 83.0 & 75.0 & 48.0 & 53.5 & 64.0 & 34.0 & 38.0 & 69.0 & 58.0 & 47.1 & 36.4 & 38.7 & 67.4 & 34.6 & 74.3 & 43.7 & 42.3 \\
GPT 5.4 Mini & 58.2 & 52.5 & 55.6 & 26.7 & 83.3 & 28.5 & 84.0 & 78.0 & 82.0 & 58.0 & 58.8 & 72.0 & 44.0 & 72.0 & 67.4 & 57.0 & 50.3 & 39.8 & 41.7 & 68.6 & 39.9 & 75.2 & 55.6 & 46.9 \\
Gemini 3.1 Pro & 64.8 & 67.7 & 48.3 & 21.0 & 83.0 & 27.8 & 93.0 & 86.0 & 97.0 & 79.0 & 62.1 & 84.0 & 86.0 & 92.0 & 71.5 & 73.0 & 46.9 & 42.3 & 32.5 & 73.4 & 46.9 & 78.0 & 62.4 & 36.0 \\
Claude Opus 4.7 & 65.5 & 57.6 & 63.7 & 22.6 & 83.5 & 32.4 & 80.0 & 93.0 & 90.0 & 67.0 & 66.5 & 92.0 & 83.0 & 96.0 & 68.7 & 74.0 & 50.4 & 41.3 & 43.5 & 71.6 & 44.2 & 77.2 & 61.4 & 46.2 \\
Qwen 3.6 Plus & 61.1 & 72.1 & 48.0 & 22.2 & 82.4 & 28.3 & 89.0 & 73.0 & 94.0 & 67.0 & 55.9 & 79.0 & 71.0 & 85.0 & 70.9 & 67.0 & 47.9 & 41.8 & 31.6 & 69.8 & 42.7 & 74.5 & 58.4 & 34.2 \\
\midrule
\multicolumn{25}{l}{\textit{Open-Source Specialized Models}} \\
\midrule
UserLM 8B & 13.4 & 37.3 & 9.7 & 4.2 & 77.7 & 17.8 & 1.0 & 0.0 & 11.0 & 3.0 & 3.5 & 13.0 & 1.0 & 5.0 & 61.5 & 1.0 & 8.7 & 2.5 & 5.9 & 26.8 & 3.8 & 1.8 & 3.4 & 9.1 \\
Coser 8B & 19.5 & 44.4 & 23.6 & 2.5 & 75.6 & 25.0 & 1.0 & 0.0 & 3.0 & 0.0 & 30.0 & 44.0 & 4.0 & 8.0 & 68.0 & 0.0 & 9.2 & 15.8 & 8.4 & 26.6 & 4.3 & 21.0 & 25.4 & 8.2 \\
HumanLM 8B & 48.7 & 37.2 & 45.4 & 21.7 & 82.9 & 26.7 & 70.0 & 56.0 & 75.0 & 47.0 & 19.8 & 67.0 & 40.0 & 36.0 & 69.0 & 56.0 & 42.1 & 33.1 & 33.0 & 66.8 & 35.2 & 75.2 & 48.7 & 37.4 \\
SotopiaRL 7B & 39.7 & 44.6 & 38.3 & 25.8 & 83.5 & 31.7 & 0.0 & 31.0 & 40.0 & 46.0 & 30.3 & 62.0 & 29.0 & 21.0 & 70.5 & 17.0 & 42.8 & 30.2 & 34.2 & 58.6 & 24.3 & 69.8 & 50.2 & 32.3 \\
\midrule
\multicolumn{25}{l}{\textit{Ours}} \\
\midrule
Qwen3-4B-Base & 47.8 & 40.9 & 45.0 & 19.6 & 82.5 & 26.5 & 51.0 & 66.0 & 57.0 & 53.0 & 34.0 & 61.0 & 40.0 & 37.0 & 66.8 & 55.0 & 43.6 & 31.5 & 29.3 & 65.6 & 31.3 & 74.6 & 53.3 & 33.8 \\
Qwen3-4B-Base + Mid & 39.4 & 59.6 & 41.6 & 11.0 & 80.6 & 43.9 & 65.0 & 52.0 & 70.0 & 36.0 & 15.5 & 70.0 & 28.0 & 35.0 & 69.7 & 23.0 & 22.6 & 16.1 & 13.0 & 52.6 & 6.5 & 36.8 & 39.7 & 17.5 \\
Qwen3-4B-Base + Post & 60.5 & 88.9 & 60.1 & 23.2 & 84.3 & 45.1 & 81.0 & 68.0 & 75.0 & 56.0 & 50.6 & 79.0 & 60.0 & 78.0 & 71.2 & 61.0 & 50.1 & 38.4 & 36.8 & 74.4 & 36.8 & 74.1 & 60.3 & 39.0 \\
Qwen3-4B-Base + Mid + Post & 62.6 & 89.8 & 68.3 & 24.5 & 84.8 & 48.0 & 82.0 & 69.0 & 81.0 & 55.0 & 55.0 & 89.0 & 63.0 & 88.0 & 71.3 & 53.0 & 48.7 & 42.4 & 39.1 & 71.8 & 37.8 & 74.3 & 63.3 & 40.5 \\
\midrule
Qwen3-8B-Base & 26.9 & 31.0 & 13.9 & 12.0 & 79.6 & 21.4 & 23.0 & 12.0 & 19.0 & 37.0 & 6.1 & 32.0 & 19.0 & 18.0 & 66.2 & 24.0 & 26.4 & 12.7 & 17.8 & 49.0 & 12.1 & 46.6 & 21.5 & 18.2 \\
Qwen3-8B-Base + Mid & 41.1 & 49.5 & 49.1 & 7.8 & 80.3 & 45.6 & 62.0 & 54.0 & 72.0 & 42.0 & 24.8 & 58.0 & 25.0 & 42.0 & 68.1 & 18.0 & 22.3 & 15.1 & 15.4 & 53.6 & 16.5 & 68.1 & 38.8 & 17.0 \\
Qwen3-8B-Base + Post & 63.8 & 90.5 & 63.0 & 28.5 & 85.4 & 49.7 & 75.0 & 74.0 & 84.0 & 61.0 & 59.6 & 80.0 & 68.0 & 89.0 & 70.7 & 56.0 & 53.2 & 42.9 & 40.4 & 73.0 & 40.5 & 77.7 & 63.4 & 41.1 \\
Qwen3-8B-Base + Mid + Post & 64.6 & 90.1 & 68.3 & 28.2 & 84.1 & 49.2 & 80.0 & 79.0 & 83.0 & 60.0 & 62.6 & 82.0 & 68.0 & 94.0 & 70.7 & 59.0 & 51.4 & 42.7 & 41.9 & 72.6 & 39.1 & 75.1 & 63.2 & 42.0 \\
\midrule
Qwen3-8B-Inst & 48.3 & 46.0 & 54.0 & 24.7 & 83.6 & 27.7 & 23.0 & 62.0 & 67.0 & 54.0 & 43.5 & 70.0 & 42.0 & 41.0 & 68.9 & 27.0 & 43.7 & 32.5 & 33.2 & 68.6 & 34.1 & 73.6 & 53.6 & 37.2 \\
Qwen3-8B-Inst + Mid & 43.1 & 57.1 & 48.6 & 12.1 & 80.5 & 43.1 & 66.0 & 56.0 & 68.0 & 44.0 & 24.0 & 62.0 & 25.0 & 52.0 & 68.6 & 40.0 & 24.1 & 18.0 & 13.7 & 59.2 & 17.9 & 55.6 & 38.3 & 17.3 \\
Qwen3-8B-Inst + Post$^{\dagger}$ & 65.3 & 92.6 & 70.6 & 24.9 & 84.3 & 48.7 & 89.0 & 78.0 & 82.0 & 59.0 & 63.9 & 79.0 & 74.0 & 93.0 & 70.6 & 66.0 & 49.2 & 44.3 & 41.0 & 72.2 & 40.9 & 75.5 & 63.6 & 40.1 \\
Qwen3-8B-Inst + Post & 66.0 & 91.6 & 70.0 & 26.8 & 84.4 & 49.3 & 90.0 & 76.0 & 87.0 & 64.0 & 63.9 & 79.0 & 75.0 & 92.0 & 71.7 & 62.0 & 50.1 & 44.4 & 42.1 & 74.2 & 40.4 & 77.2 & 63.6 & 42.6 \\
Qwen3-8B-Inst + Mid + Post & 65.7 & 93.9 & 72.8 & 28.9 & 85.1 & 48.4 & 90.0 & 79.0 & 89.0 & 60.0 & 64.1 & 73.0 & 70.0 & 91.0 & 70.8 & 61.0 & 51.7 & 43.3 & 41.2 & 73.8 & 41.5 & 77.8 & 64.2 & 41.1 \\
Ditto-v2 8B~\citep{sun2026reinforcinghumanbehaviorsimulation} & 66.5 & 92.7 & 72.0 & 27.4 & 85.1 & 49.4 & 95.0 & 82.0 & 91.0 & 61.0 & 65.3 & 83.0 & 64.0 & 93.0 & 70.7 & 65.0 & 49.4 & 43.6 & 41.6 & 73.6 & 42.1 & 75.1 & 65.2 & 41.6 \\
\bottomrule
\end{tabular}
\end{table*}

\subsection{Full Out-of-Distribution Evaluation Results}
\label{app:ood-eval-full}
\label{app:ood-usi-full}
\Cref{tab:ood_usi,tab:humt} report the full per-model results for the two out-of-distribution checks in \cref{sec:ood}: interactive user simulation on \tauusi and raw text human-likeness on HumT.
\begin{table}[t]
\centering
\small
\setlength{\tabcolsep}{4.5pt}
\caption{\textbf{Out-of-distribution user simulation on \tauusi}~\citep{zhou2026mindsim2realgapuser}.
Each simulator drives a fixed GPT-5.2 tool-use agent through 165 \taubench retail/airline tasks~\citep{yao2024taubenchbenchmarktoolagentuserinteraction}; we report alignment to human annotators on the four behavioral dimensions (\emph{Conv}ersation, \emph{Info}rmation, \emph{Clarif}ication, \emph{React}ion; Sørensen--Dice $\times100$), post-hoc survey agreement (\emph{Eval}), task-success calibration error (\emph{ECE}, lower is better), and the composite \emph{USI}, averaged over three annotation batches.
\textbf{Bold} marks the best non-human entry per column.
\odyssim variants are evaluated zero-shot; \taubench's tool-agent domain is disjoint from the \soulindex training corpus.}
\label{tab:ood_usi}
\begin{tabular}{l ccccc c c}
\toprule
\textbf{User Simulator} & \textbf{Conv} & \textbf{Info} & \textbf{Clarif} & \textbf{React} & \textbf{Eval} & \textbf{ECE}$\downarrow$ & \textbf{USI} \\
\midrule
\textit{Human (upper bound)} & 87.4 & 97.9 & 88.0 & 93.5 & 97.4 & 0.081 & 92.7 \\
\midrule
\multicolumn{8}{l}{\textit{Frontier LLMs}} \\
DeepSeek-V3.1            & 45.1 & 86.6 & 74.5 & 87.6 & 74.3 & 0.119 & \textbf{76.1} \\
GPT-5.1                  & 47.3 & 77.4 & 73.3 & 88.1 & 72.1 & 0.172 & 73.5 \\
Gemini-3.1-Pro           & 44.1 & 67.1 & 48.9 & 45.3 & \textbf{75.1} & \textbf{0.101} & 61.7 \\
Claude-Opus-4            & 32.6 & 71.9 & 46.6 & 44.9 & 73.4 & 0.139 & 59.2 \\
\midrule
\multicolumn{8}{l}{\textit{Specialized behavioral simulators}} \\
CoSER-8B~\citep{wang2025coser}        & 37.8 & 71.5 & 71.6 & 69.9 & 63.3 & 0.109 & 67.2 \\
UserLM-8B~\citep{naous2025userllm}    & 30.8 & 50.8 & 56.8 & 80.0 & 67.4 & 0.140 & 62.0 \\
Human-Like-7B                          & 35.7 & 55.0 & 51.6 & 65.9 & 72.8 & 0.220 & 59.8 \\
HumanLM-8B~\citep{wu2026humanlm}      & 30.1 & 19.5 & 38.5 & 50.7 & 61.6 & 0.192 & 46.9 \\
\midrule
\multicolumn{8}{l}{\textit{Off-the-shelf instruct model (\odyssim's base checkpoint)}} \\
Qwen3-8B-Instruct                      & 50.5 & 74.3 & \textbf{80.2} & 80.3 & 72.2 & 0.293 & 71.4 \\
\midrule
\multicolumn{8}{l}{\textit{\odyssim (ours)}} \\
\odyssimmodel-8B                       & 46.4 & 88.2 & 79.7 & \textbf{93.2} & 70.0 & 0.255 & 75.4 \\
\odyssimmodel-8B-Mid                   & 45.5 & 83.5 & 63.7 & 76.1 & 69.1 & 0.356 & 67.1 \\
\odyssimmodel-Inst-8B                  & 45.1 & 85.2 & 74.1 & 83.6 & 74.4 & 0.339 & 71.4 \\
\odyssimmodel-4B                       & 30.0 & 66.3 & 76.2 & 82.2 & 70.5 & 0.242 & 66.8 \\
\odyssimmodel-4B-Mid                   & \textbf{51.1} & \textbf{91.5} & 65.5 & 82.6 & 69.0 & 0.242 & 72.6 \\
\odyssimmodel-Inst-4B                  & 37.8 & 73.4 & 62.6 & 62.1 & 72.1 & 0.236 & 64.1 \\
\bottomrule
\end{tabular}
\end{table}

\begin{table}[t]
\centering
\small
\setlength{\tabcolsep}{4pt}
\caption{\textbf{Human-likeness beyond \soulindex (HumT~\citep{cheng2025humt}; $\times100$, higher is more human-like).}
Per-text anthropomorphism scalar---the log-prob ratio of animate vs.\ inanimate prefixes under a fixed GPT-2 backbone---measured on HumT's held-out prompts, a signal entirely outside the \soulindex training objective.
\textit{Others} is the best prior human-simulation model. \textbf{Bold} marks the best entry.}
\label{tab:humt}
\begin{tabular}{l ccccc cc cc}
\toprule
& \makecell{\textbf{GPT}\\\textbf{5.5}}
& \makecell{\textbf{Gemini}\\\textbf{3.1 Pro}}
& \makecell{\textbf{Claude}\\\textbf{Opus 4.7}}
& \makecell{\textbf{Qwen}\\\textbf{3.6 Plus}}
& \makecell{\textbf{Others}\\\textbf{*}}
& \makecell{\textbf{Qwen3}\\\textbf{8B Inst}}
& \makecell{\textbf{Base}\\\textbf{8B}}
& \makecell{\odyssimmodel\\\textbf{8B-Mid}}
& \makecell{\odyssimmodel\\\textbf{8B}} \\
\midrule
HumT & 4.6 & 7.0 & 3.7 & 3.8 & 2.9 & 4.2 & 2.1 & \textbf{13.4} & 11.9 \\
\bottomrule
\end{tabular}
\end{table}

\section{Limitations and Future Work}
\label{app:limitations}
Our investigation is text-only, but human behavior is fundamentally multimodal (voice, gesture, facial expression).
The \odyssim corpus, while diverse, still consists largely of \emph{performed} behavior (conversations typed for an audience) rather than naturalistic decision traces, and our evaluation depends on LLM judges that may carry their own biases.
Promising directions include extending to multimodal behavior, scaling to larger base models, and testing whether behavioral midtraining transfers across languages and cultures.

\section{Ablation}
\label{sec:ablation}

\begin{table}[t]
\centering
\small
\setlength{\tabcolsep}{6pt}
\renewcommand{\arraystretch}{1.1}
\caption{\textbf{Ablation over training stages.} \textit{Mid-Train} denotes mid-stage SFT and \textit{Post-Train} denotes post-training (RL). \textit{Avg} is the unweighted mean across the 23 evaluation tasks.}
\label{tab:ablation}
\begin{tabular}{l c c | c}
\toprule
\textbf{Pretrained} & \textbf{Mid-Train} & \textbf{Post-Train} & \textbf{Avg} \\
\midrule
Base 4B      &            &            & 47.8 \\
Base 4B      & \checkmark &            & 39.4 \\
Base 4B      &            & \checkmark & 60.5 \\
Base 4B      & \checkmark & \checkmark & 62.6 \\
\midrule
Base 8B      &            &            & 26.9 \\
Base 8B      & \checkmark &            & 41.1 \\
Base 8B      &            & \checkmark & 63.8 \\
Base 8B      & \checkmark & \checkmark & 64.6 \\
\midrule
Instruct 8B  &            &            & 48.3 \\
Instruct 8B  & \checkmark &            & 43.1 \\
Instruct 8B  &            & \checkmark & 66.0 \\
Instruct 8B  & \checkmark & \checkmark & 65.7 \\
\bottomrule
\end{tabular}
\end{table}

\paragraph{Data ablation}
Table~\ref{tab:ablation} shows that \textbf{post-training contributes the largest gains}.
RL alone raises the average score from 47.8 to 60.5 for Base 4B, 26.9 to 63.8 for Base 8B, and 48.3 to 66.0 for Instruct 8B.
Midtraining further improves the post-trained base models, reaching 62.6 for Base 4B and 64.6 for Base 8B.
Overall, these results suggest that \textbf{RL is the key step for behavioral alignment}, while midtraining provides additional gains when applied before RL on pretrained base models.

\begin{figure*}[t!]
\centering
\includegraphics[width=\textwidth]{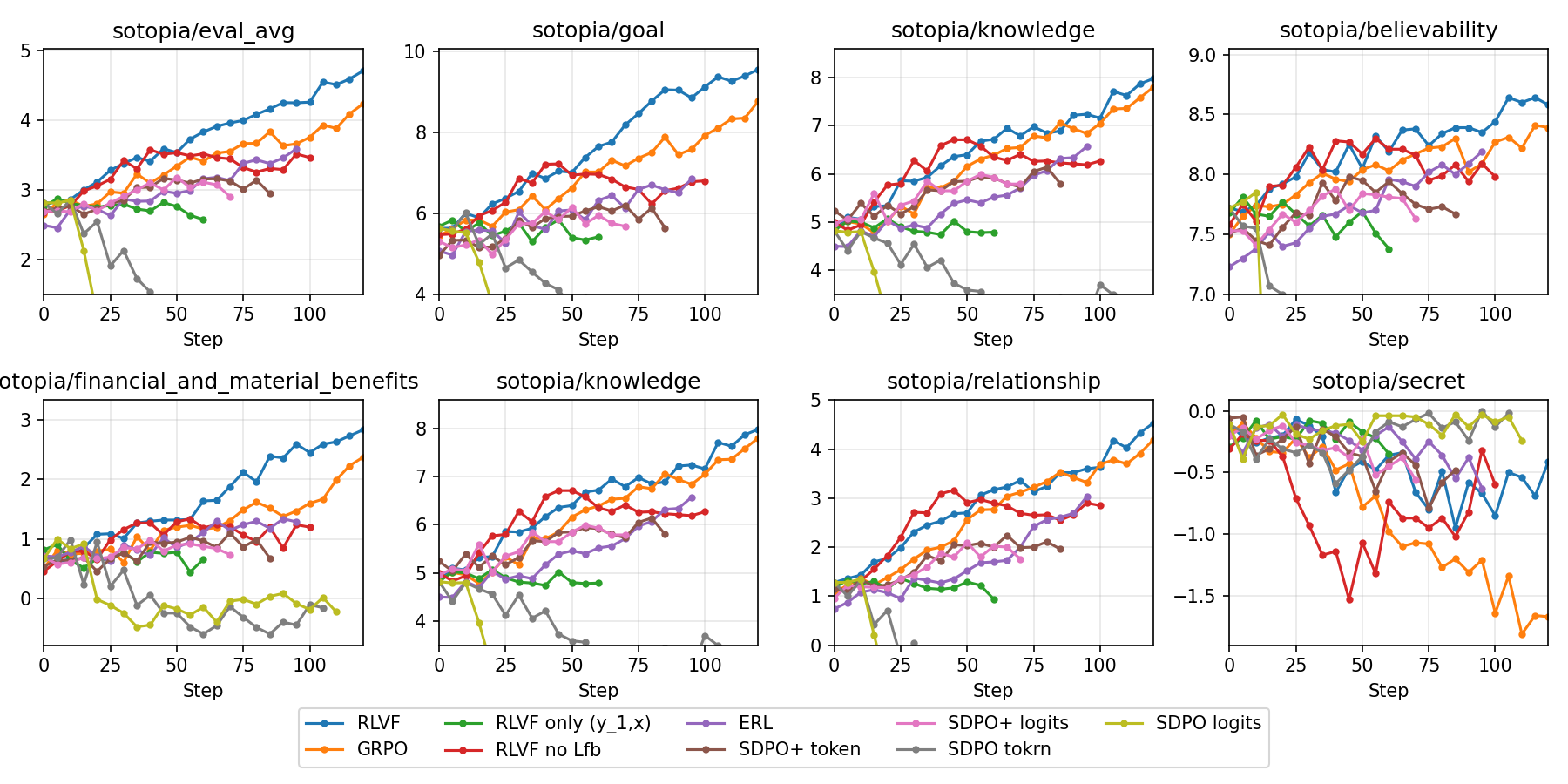}
\caption{RL-algorithm ablation on Sotopia. Reward over training for our RLVF (RL with verbal feedback) against GRPO, ERL~\citep{shi2026experiential}, and the SDPO~\citep{chen2025sdpo} variants defined in the text.}\label{fig:ablation}
\end{figure*}

\paragraph{RL ablation}
Figure \ref{fig:ablation} shows an ablation study on the learning algorithm. Specifically, we compare our RLVF (RL with verbal feedback) method with the following methods: (1) GRPO; (2) RLVF only $(y_1, x)$, a variant that only trains on $(y_1, x)$ pairs and removes the loss on $(y_0, x)$ and $(y_0, x+h)$ pairs; (3) ERL~\citep{shi2026experiential}, which uses a supervised fine-tuning objective on $y_1$ instead of RL; and several SDPO~\citep{chen2025sdpo} variants including (4) SDPO+ token, which combines SDPO with token-level loss and the GRPO objective; (5) SDPO+ logits, which combines SDPO with logits-level loss and the GRPO objective; (6) SDPO token, i.e., SDPO with token-level loss; and (7) SDPO logits, i.e., SDPO with logits-level loss.

From the results, we observe that RLVF performs better than GRPO on most sub-metrics, especially on smaller metrics such as \textit{secret}. 
Notably, \textit{secret} measures whether
the agent avoids leaking private information---a safety-critical dimension that is not directly
optimized in the scalar RL reward but can be explicitly addressed through verbal feedback.
GRPO reduces all multi-dimensional scores into a single scalar reward, which loses information and weakens learning signals for improving minor metrics, while RLVF can learn all dimensions effectively through fine-grained feedback.
Compared to different algorithms that utilize feedback, we observe that reverse-KL-based methods such as SDPO collapse on most metrics except \textit{secret}. Combining SDPO with GRPO improves performance but still performs worse than RLVF. ERL also underperforms, likely because the lack of reward normalization makes learning unstable under noisy feedback.

\section{Post-training Data Statistics}
\label{sec:post_data_stats}

\paragraph{Construction.}
For every \soulindex task we maintain three matched data slices: a held-out evaluation slice (capped to 100 instances per task --- 500 for HumanLLM --- for fast judge-based scoring), a per-task RL training set, and an expert-distillation (SFT) bucket consolidating top-scoring rollouts.
The RL slice is constructed by sampling 1{,}024 instances from each task's training split (Verl-style on-policy RL, batch 64, group size 8 rollouts/prompt, max input/output 8{,}192~tokens, LoRA rank 32 / alpha 64 on Qwen3-8B-Instruct); when a task's full training set is smaller than the 1{,}024 target, we use it in full (e.g., \texttt{sotopia-hard} 405, \texttt{social\_r1} 687, \texttt{sim\_doc} 815, \texttt{sim\_math} 1{,}043, \texttt{lifechoices} 1{,}150).
A few tasks deviate from the 1{,}024 default: HUMANUAL splits use 1{,}000 prompts each (six domains $\to$ 6{,}000 total), \texttt{alignx} pulls a wider 8{,}191-prompt slice across its five sub-splits, and \texttt{socsci210} is downsampled to 2{,}000 from a 2.4M pool.
We release the post-training prompts, held-out evaluation slices, and task-organized train/test files as the HuggingFace dataset \posttrainingdata.

\paragraph{Reward signal.}
The RL algorithm is RLVF (RL with verbal feedback) for tasks scored by an LLM judge that returns both a scalar reward and a textual critique, and GRPO with the scalar reward only for tasks with verifiable rewards (full per-task assignment in \cref{tab:eval_rl_combined}).
We use \texttt{gpt-5-nano} as the default judge / environment model, except for CoSER, where the literary-character rubric is more delicate and we use \texttt{gpt-5.4} for improved judge robustness.
For multi-agent tasks (CoSER), the episode-level reward returned by the judge is assigned to each agent's rollout so on-policy advantages remain well-defined.

\paragraph{Expert distillation.}
After per-task RL we run rejection-sampled fine-tuning (RFT) on the experts: for each prompt we draw 8 rollouts, keep the top-1 by reward, deduplicate, and aggregate into 13 family-level SFT files (\cref{tab:sft_data_stats}); these files form the \emph{expert distillation} corpus used to merge the per-task experts back into a single deployable \odyssimmodel-8B.
Family-level grouping reflects shared-skill task clusters: \texttt{tom\_sft} consolidates Fantom/Hitom/Paratomi (Theory of Mind), \texttt{humanual\_sft} aggregates all six HUMANUAL domains, \texttt{simarena\_sft} covers SimArena-Math/Doc, and \texttt{sim\_sft} captures user-simulation tasks with overlapping prompt distributions.

\begin{table*}[t]
\centering
\small
\setlength{\tabcolsep}{4pt}
\renewcommand{\arraystretch}{1.05}
\caption{Per-task evaluation and RL-training data statistics. \textbf{Eval} reports the \soulindex held-out slice we score (capped to 100 instances per task; HumanLLM uses 500). \textbf{RL Source}, \textbf{Original}, and \textbf{Used} report the per-task RL split, its full size, and the number of prompts we sample for RL training. \textbf{Alg} indicates RLVF (RL with verbal feedback from an LLM judge) or GRPO (scalar verifiable reward).}
\label{tab:eval_rl_combined}
\begin{tabular}{l r l r r c}
\toprule
\textbf{Task} & \textbf{Eval} & \textbf{RL Source} & \textbf{Original} & \textbf{Used} & \textbf{Alg} \\
\midrule
sotopia-hard       & 100 & sotopia\_clean\_rl       & 405       & 405   & RLVF \\
coser              & 100 & coser\_rl\_train         & 21{,}175  & 1{,}024 & RLVF \\
lifechoices        & 100 & lifechoices\_hard\_rl    & 160       & 1{,}150 & GRPO \\
userllm            & 100 & userllm\_rl\_train       & 28{,}918  & 1{,}024 & RLVF \\
mirrorbench        & 100 & mirrorbench\_rl\_train   & 3{,}400   & 1{,}024 & RLVF \\
fantom             & 100 & fantom\_rl\_train        & 2{,}366   & 1{,}024 & RLVF \\
hitom              & 100 & hitom\_rl\_train         & 6{,}000   & 1{,}024 & GRPO \\
paratomi           & 100 & paratomi\_rl\_train      & 1{,}863   & 1{,}024 & RLVF \\
mistakes           & 100 & mistakes\_rl\_train      & 3{,}494   & 1{,}024 & GRPO \\
twinvoice          & 100 & twinvoice\_rl\_train     & ---       & 1{,}024 & RLVF \\
social\_r1         & 100 & social\_r1\_rl           & 687       & 687   & GRPO \\
behaviorchain      & 100 & behaviorchain\_rl\_train & 5{,}000   & 1{,}024 & GRPO \\
sim\_math          & 100 & sim\_math\_rl            & 1{,}043   & 1{,}043 & GRPO \\
sim\_doc           & 100 & sim\_doc\_rl             & 815       & 815   & GRPO \\
humanual-book      & 100 & humanual-book            & 34{,}170  & 1{,}000 & GRPO \\
humanual-chat      & 100 & humanual-chat            & 23{,}141  & 1{,}000 & GRPO \\
humanual-email     & 100 & humanual-email           & 6{,}377   & 1{,}000 & GRPO \\
humanual-news      & 100 & humanual-news            & 48{,}618  & 1{,}000 & GRPO \\
humanual-opinion   & 100 & humanual-opinion         & 37{,}791  & 1{,}000 & GRPO \\
humanual-politics  & 100 & humanual-politics        & 45{,}429  & 1{,}000 & GRPO \\
alignx-demo        & 100 & \multirow{5}{*}{alignx\_rl\_8k} & \multirow{5}{*}{74{,}826} & \multirow{5}{*}{8{,}191} & \multirow{5}{*}{GRPO} \\
alignx-pair        & 100 &  &  &  &  \\
alignx-ugc         & 100 &  &  &  &  \\
alignx-arbitrary   & 100 &  &  &  &  \\
alignx-history16   & 100 &  &  &  &  \\
socsci210          & 100 & socsci210\_rl\_2k        & 2{,}418{,}748 & 2{,}000 & GRPO \\
humanllm           & 500 & humanllm\_rl\_train      & 185{,}912 & 1{,}024 & GRPO \\
\midrule
\textbf{Total}     & \textbf{3{,}100} & --- & --- & \textbf{30{,}551} & --- \\
\bottomrule
\end{tabular}
\end{table*}

\begin{table}[t]
\centering
\small
\setlength{\tabcolsep}{6pt}
\renewcommand{\arraystretch}{1.05}
\caption{Expert-distillation (SFT) data: family-level files derived from rollout-filtered RFT data (top-1-of-8 reward filtering, then deduplication). Each file aggregates one or more \soulindex tasks (e.g., \texttt{tom\_sft} covers Fantom / Hitom / Paratomi; \texttt{humanual\_sft} covers all six HUMANUAL domains).}
\label{tab:sft_data_stats}
\begin{tabular}{lr}
\toprule
\textbf{SFT File} & \textbf{Rows} \\
\midrule
\texttt{alignx\_sft}        & 7{,}720 \\
\texttt{behaviorchain\_sft} & 1{,}994 \\
\texttt{coser\_sft}         & 4{,}096 \\
\texttt{humanllm\_sft}      & 3{,}181 \\
\texttt{humanual\_sft}      & 11{,}969 \\
\texttt{lifechoices\_sft}   & 3{,}912 \\
\texttt{mistakes\_sft}      & 3{,}077 \\
\texttt{sim\_sft}           & 9{,}028 \\
\texttt{simarena\_sft}      & 3{,}716 \\
\texttt{social\_r1\_sft}    & 557 \\
\texttt{socsci210\_sft}     & 2{,}000 \\
\texttt{sotopia\_sft}       & 1{,}620 \\
\texttt{tom\_sft}           & 5{,}832 \\
\midrule
\textbf{Total} & \textbf{58{,}702} \\
\bottomrule
\end{tabular}
\end{table}

\section{Midtraining Dataset Details}
\label{app:dataset}

\subsection{Dataset Details and Sources}
\label{app:dataset-sources}

\Cref{tab:dataset-sources} lists every released dataset by its identifier in the midtraining release \midtrainingdata, its display name, its canonical paper (citation key in our bibliography), and its primary source/download location. Where no formal publication exists, or where the bibtex entry is not yet in this paper, the \emph{Paper} column is marked ``---''. ConvoKit-distributed corpora share a single download interface~\citep{chang2020convokit}; per-corpus URLs are listed for direct retrieval.

\begin{table}[ht]
\caption{Source bibliography for the 62 released datasets in \midtrainingdata, sorted by capability dimension matching \cref{fig:taxonomy}. Dataset ids are abbreviated for compactness (e.g.\ \texttt{-corpus} suffix dropped on ConvoKit entries; \texttt{conversations-gone-awry} $\to$ \texttt{CGA}; \texttt{wiki-articles-for-deletion} $\to$ \texttt{wiki-AfD}); the released identifiers in the manifest retain their full names. \emph{Paper}: canonical citation; ``---'' indicates either no formal publication or no bibtex entry in this paper. \emph{Source}: primary download/code location; ConvoKit corpora are downloadable through the unified ConvoKit interface~\citep{chang2020convokit}.}
\label{tab:dataset-sources}
\centering
\scriptsize
\setlength{\tabcolsep}{4pt}
\renewcommand{\arraystretch}{1.05}
\resizebox{\textwidth}{!}{%
\begin{tabular}{r l l l l}
\toprule
\# & \textbf{Dataset id} & \textbf{Display name} & \textbf{Paper} & \textbf{Source} \\
\midrule
1 & \texttt{wildchat} & WildChat & \citep{zhao2024wildchat} & \url{https://huggingface.co/datasets/allenai/WildChat-4.8M} \\
2 & \texttt{lmsys} & LMSYS-Chat-1M & \citep{zheng2024lmsys} & \url{https://huggingface.co/datasets/lmsys/lmsys-chat-1m} \\
3 & \texttt{oasst1} & OpenAssistant 1 & \citep{kopf2023openassistant} & \url{https://huggingface.co/datasets/OpenAssistant/oasst1} \\
4 & \texttt{oasst2} & OpenAssistant 2 & \citep{kopf2023openassistant} & \url{https://huggingface.co/datasets/OpenAssistant/oasst2} \\
5 & \texttt{dailydialog} & DailyDialog & \citep{li2017dailydialog} & \url{https://huggingface.co/datasets/ConvLab/dailydialog} \\
6 & \texttt{cornell\_movie} & Cornell Movie Dialogs & \citep{danescu2011cornell} & \url{https://www.cs.cornell.edu/~cristian/data/cornell_movie_dialogs_corpus.zip} \\
7 & \texttt{convokit\_friends} & Friends & \citep{chang2020convokit} & \url{https://convokit.cornell.edu/datasets/friends-corpus} \\
8 & \texttt{convokit\_switchboard} & Switchboard & \citep{chang2020convokit} & \url{https://convokit.cornell.edu/datasets/switchboard-corpus} \\
9 & \texttt{convokit\_small-pool} & small-pool (8-corpus merge) & \citep{chang2020convokit} & \url{https://convokit.cornell.edu/} \\
10 & \texttt{convokit\_tennis} & Tennis press & \citep{chang2020convokit} & \url{https://convokit.cornell.edu/datasets/tennis-corpus} \\
11 & \texttt{convokit\_npr-2p} & NPR 2-person & \citep{chang2020convokit} & \url{https://convokit.cornell.edu/datasets/npr-2p-corpus} \\
12 & \texttt{convokit\_mediasum} & MediaSum & \citep{chang2020convokit} & \url{https://convokit.cornell.edu/datasets/mediasum-corpus} \\
13 & \texttt{convokit\_parliament} & UK Parliament & \citep{chang2020convokit} & \url{https://convokit.cornell.edu/datasets/parliament-corpus} \\
14 & \texttt{convokit\_supreme} & SCOTUS oral arguments & \citep{chang2020convokit} & \url{https://convokit.cornell.edu/datasets/supreme-corpus} \\
15 & \texttt{convokit\_reddit-corpus-small} & Reddit (small) & \citep{chang2020convokit} & \url{https://convokit.cornell.edu/datasets/reddit-corpus-small} \\
16 & \texttt{convokit\_reddit-coarse} & Reddit Coarse-Discourse & \citep{chang2020convokit} & \url{https://convokit.cornell.edu/datasets/reddit-coarse-discourse-corpus} \\
17 & \texttt{convokit\_wiki} & Wikipedia talk & \citep{chang2020convokit} & \url{https://convokit.cornell.edu/datasets/wiki-corpus} \\
18 & \texttt{convokit\_wikiconv-2018} & WikiConv-2018 & \citep{chang2020convokit} & \url{https://convokit.cornell.edu/datasets/wikiconv-corpus} \\
19 & \texttt{convokit\_wiki-AfD} & Wikipedia AfD & \citep{chang2020convokit} & \url{https://convokit.cornell.edu/datasets/wiki-articles-for-deletion-corpus} \\
20 & \texttt{convokit\_chromium} & Chromium code review & \citep{chang2020convokit} & \url{https://convokit.cornell.edu/datasets/chromium-corpus} \\
21 & \texttt{empathetic} & Empathetic Dialogues & \citep{rashkin2019empathetic} & \url{https://huggingface.co/datasets/facebook/empathetic_dialogues} \\
22 & \texttt{convokit\_emotional-support} & ESConv (emotional support) & \citep{liu2021esconv} & \url{https://convokit.cornell.edu/datasets/emotional-support} \\
23 & \texttt{convokit\_casino} & CaSiNo (camping negotiations) & \citep{chawla2021casino} & \url{https://convokit.cornell.edu/datasets/casino-corpus} \\
24 & \texttt{convokit\_persuasion4good} & Persuasion-For-Good & \citep{wang2019persuasion} & \url{https://convokit.cornell.edu/datasets/persuasionforgood-corpus} \\
25 & \texttt{convokit\_winning-args} & ChangeMyView Winning Args & \citep{chang2020convokit} & \url{https://convokit.cornell.edu/datasets/winning-args-corpus} \\
26 & \texttt{convokit\_CGA-wiki} & CGA: Wikipedia talk & \citep{chang2020convokit} & \url{https://convokit.cornell.edu/datasets/conversations-gone-awry-corpus} \\
27 & \texttt{convokit\_CGA-cmv} & CGA: r/ChangeMyView & \citep{chang2020convokit} & \url{https://convokit.cornell.edu/datasets/conversations-gone-awry-cmv-corpus} \\
28 & \texttt{convokit\_CGA-cmv-large} & CGA: CMV large & \citep{chang2020convokit} & \url{https://convokit.cornell.edu/datasets/conversations-gone-awry-cmv-corpus-large} \\
29 & \texttt{convokit\_IDEA-NTHU-tweets} & IDEA-NTHU unintended-offense tweets & --- & \url{https://github.com/IDEA-NTHU-Taiwan/unintended-offense-tweets} \\
30 & \texttt{tom\_fantom} & FANToM & \citep{kim2023fantom} & \url{https://github.com/skywalker023/fantom} \\
31 & \texttt{tom\_hitom} & HiToM & \citep{wu2023hitom} & \url{https://github.com/ying-hui-he/Hi-ToM_dataset} \\
32 & \texttt{tom\_paratomi} & ToMi / ParaToMi & \citep{le2019tomi} & \url{https://github.com/msclar/symbolictom} \\
33 & \texttt{tom\_mindgames} & MindGames & \citep{sileo2023mindgames} & \url{https://huggingface.co/datasets/sileod/mindgames} \\
34 & \texttt{tom\_socialiqa} & Social IQA & \citep{sap2019socialiqa} & \url{https://huggingface.co/datasets/allenai/social_i_qa} \\
35 & \texttt{tom\_moralstories} & Moral Stories & \citep{emelin2021moralstories} & \url{https://huggingface.co/datasets/demelin/moral_stories} \\
36 & \texttt{tom\_from\_coser} & ToM-from-CoSER (derived) & \citep{wang2025coser} & \url{https://github.com/Neph0s/CoSER} \\
37 & \texttt{tom\_tominli} & Tom-in-Li (ToM-in-the-wild) & --- & \emph{internal (CMU-LTI)} \\
38 & \texttt{tom\_grimulkan} & Grimulkan long-form RP & --- & \url{https://huggingface.co/grimulkan} \\
39 & \texttt{tom\_characterllm} & CharacterLLM $\to$ ToM & --- & \url{https://github.com/choosewhatulike/trainable-agents} \\
40 & \texttt{psych101} & Psych-101 & \citep{binz2024psych101} & \url{https://huggingface.co/datasets/marcelbinz/Psych-101} \\
41 & \texttt{coser} & CoSER & \citep{wang2025coser} & \url{https://huggingface.co/datasets/Neph0s/CoSER} \\
42 & \texttt{soc\_cornell} & Cornell Movie + social goals & \citep{danescu2011cornell} & \url{https://www.cs.cornell.edu/~cristian/Cornell_Movie-Dialogs_Corpus.html} \\
43 & \texttt{soc\_haico} & HAICosystem & \citep{zhou2024haicosystem} & \emph{internal (CMU-LTI; related: HAICOSYSTEM, COLM 2025)} \\
44 & \texttt{soc\_persona\_conflicts} & Persona Conflicts & --- & \emph{internal (CMU-LTI)} \\
45 & \texttt{soc\_sotopia\_pi\_bc} & SOTOPIA-$\pi$ (BC) & --- & \url{https://huggingface.co/datasets/cmu-lti/sotopia-pi} \\
46 & \texttt{soc\_sotopia\_tom\_silver} & SOTOPIA-ToM (Silver) & \citep{zhou2024sotopia} & \emph{internal (CMU-LTI; built on Sotopia)} \\
47 & \texttt{humanual\_book} & HUMANUAL: book & \citep{wu2026humanlm} & \url{https://aka.ms/humanllm} \\
48 & \texttt{humanual\_chat} & HUMANUAL: chat & \citep{wu2026humanlm} & \url{https://aka.ms/humanllm} \\
49 & \texttt{humanual\_email} & HUMANUAL: email & \citep{wu2026humanlm} & \url{https://aka.ms/humanllm} \\
50 & \texttt{humanual\_news} & HUMANUAL: news & \citep{wu2026humanlm} & \url{https://aka.ms/humanllm} \\
51 & \texttt{humanual\_opinion} & HUMANUAL: opinion & \citep{wu2026humanlm} & \url{https://aka.ms/humanllm} \\
52 & \texttt{humanual\_politics} & HUMANUAL: politics & \citep{wu2026humanlm} & \url{https://aka.ms/humanllm} \\
53 & \texttt{human\_llm} & Cognitive Genome (HumanLLM) & \citep{lei2026cogenome} & \emph{internal release; arXiv:2601.15793} \\
54 & \texttt{alignx\_v2} & AlignX (v2 conversational subset) & \citep{li2025alignx} & \url{https://huggingface.co/datasets/JinaLeejnl/AlignX} \\
55 & \texttt{mathdial} & MathDial & --- & \url{https://github.com/eth-nlped/mathdial} \\
56 & \texttt{studychat} & StudyChat & --- & \url{https://huggingface.co/datasets/wmcnicho/StudyChat} \\
57 & \texttt{education\_dialogue} & Education Dialogue & --- & \url{https://github.com/google-research-datasets/Education-Dialogue-Dataset} \\
58 & \texttt{hh\_rlhf} & Anthropic HH-RLHF & \citep{bai2022training} & \url{https://huggingface.co/datasets/Anthropic/hh-rlhf} \\
59 & \texttt{nectar} & Nectar & \citep{starling2023nectar} & \url{https://huggingface.co/datasets/berkeley-nest/Nectar} \\
60 & \texttt{rm\_r1\_sft} & RM-R1-Distill SFT & --- & \url{https://huggingface.co/datasets/gaotang/RM-R1-Distill-SFT} \\
61 & \texttt{prism} & PRISM & \citep{kirk2024prism} & \url{https://huggingface.co/datasets/HannahRoseKirk/prism-alignment} \\
62 & \texttt{socsci210} & SocSci210 & \citep{kolluri2025socsci} & \url{https://huggingface.co/datasets/socratesft/SocSci210} \\
\bottomrule
\end{tabular}}
\end{table}

\subsection{Source-to-Dataset Map}
\label{app:source-map}

\begin{sidewaysfigure}[p]
    \centering
    \includegraphics[width=0.95\textheight,height=0.95\textwidth,keepaspectratio]{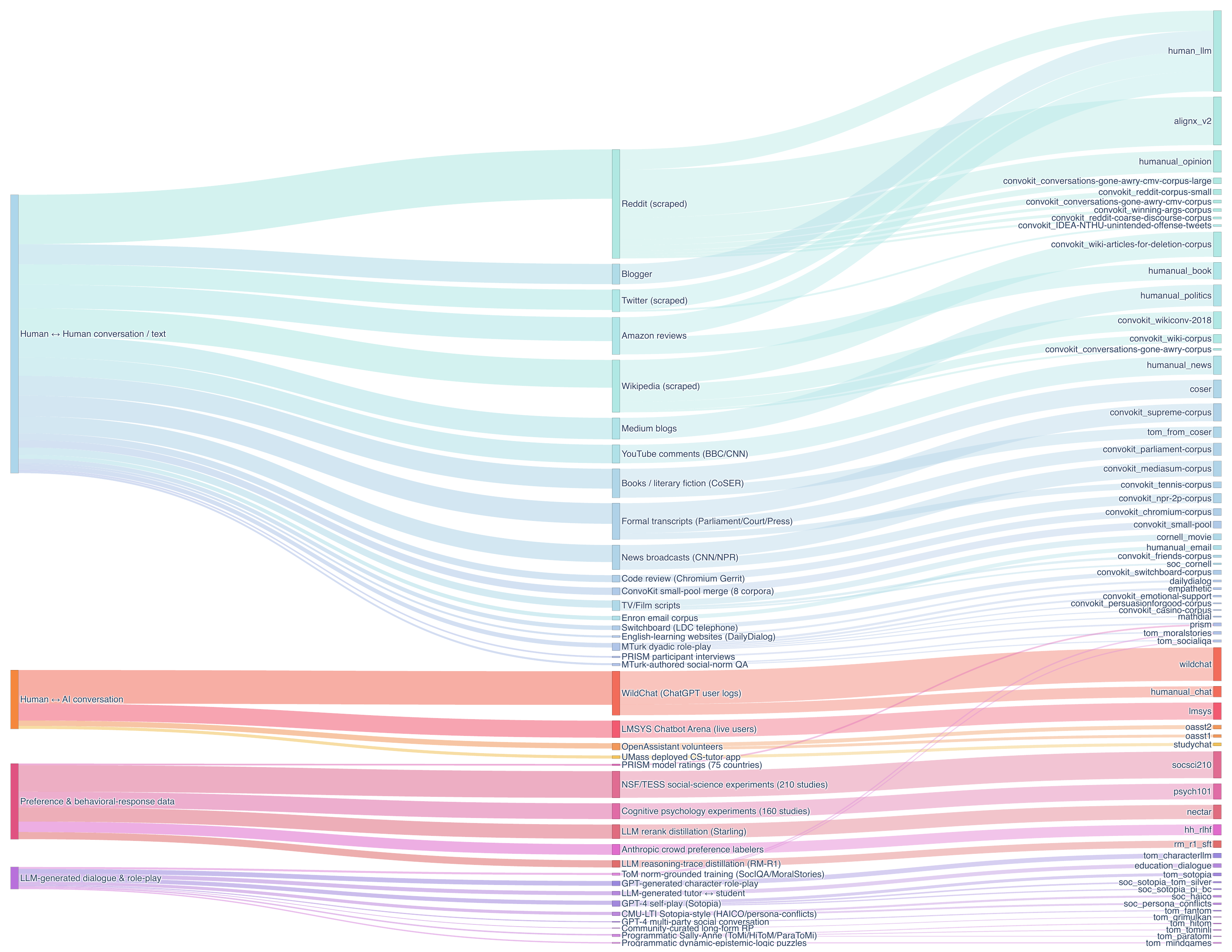}
    \caption{Three-tier provenance map for the \odyssim training mixture: provenance category (4) $\to$ origin platform (28) $\to$ dataset (63). Color encodes the Tier-1 provenance category; flow width is $\log_{10}$ of per-edge train tokens (in M).}
    \label{fig:sft-sankey}
\end{sidewaysfigure}

\Cref{fig:sft-sankey} maps the data flow at three levels of abstraction.
\textbf{Tier 1} is the broad provenance category (4 nodes): human$\leftrightarrow$human conversation/text, human$\leftrightarrow$AI conversation, preference \& behavioral-response data, and LLM-generated dialogue \& role-play.
\textbf{Tier 2} is the underlying origin platform or production method (e.g.\ Reddit, Wikipedia, WildChat, MTurk dyadic role-play, GPT-4 self-play, NSF/TESS social-science experiments) --- 28 distinct origins in total.
\textbf{Tier 3} is the resulting dataset (63 datasets in \texttt{sft\_processed\_large/}).
Flow width is proportional to $\log_{10}$ of train tokens (in millions) per edge, so small datasets (FANToM, MoralStories, GAP, etc.) stay visible alongside the multi-billion-token outliers (alignx\_v2, wildchat, socsci210).

A few datasets carry weight from more than one origin and are split across Tier-2 nodes accordingly: \texttt{prism} is divided 50/50 between participant-interview text (under human$\leftrightarrow$AI) and cross-cultural model-rating preferences (under preference data); \texttt{tom\_socialiqa} and \texttt{tom\_moralstories} are divided 50/50 between MTurk-authored social-norm short-form QA (human-authored) and ToM-training use (LLM-generated category); and \texttt{human\_llm} (Cognitive Genome) is split evenly across Reddit, Twitter, Blogger, and Amazon since the source paper does not publish per-platform token counts.

\subsection{Dataset-to-Capability Audit}
\label{app:dataset-capability-map}

\Cref{tab:dataset-capability-map} gives the full 63-dataset assignment used for \cref{fig:taxonomy}.
Identifiers are kept identical to the released split manifest and to \cref{tab:split-summary}.

\begin{table}[t]
\caption{Complete dataset-to-capability map for the 62 released datasets in \midtrainingdata, sorted alphabetically by released dataset identifier. Capability codes match \cref{fig:taxonomy}: \texttt{CONV} (discourse \& interaction), \texttt{SS} (social skills), \texttt{COG} (cognitive / mental-state), \texttt{ROLE} (persona, roleplay \& pedagogy), \texttt{EVAL} (judgment \& preference).}
\label{tab:dataset-capability-map}
\centering
\scriptsize
\setlength{\tabcolsep}{4pt}
\begin{tabular*}{\textwidth}{@{\extracolsep{\fill}} l c @{\hskip 14pt} l c @{\hskip 14pt} l c @{}}
\toprule
\textbf{Dataset} & \textbf{Cap.} & \textbf{Dataset} & \textbf{Cap.} & \textbf{Dataset} & \textbf{Cap.} \\
\midrule
\texttt{alignx\_v2} & \texttt{ROLE} & \texttt{convokit\_wikiconv-2018} & \texttt{CONV} & \texttt{psych101} & \texttt{COG} \\
\texttt{convokit\_IDEA-NTHU-tweets} & \texttt{SS} & \texttt{convokit\_winning-args} & \texttt{SS} & \texttt{rm\_r1\_sft} & \texttt{EVAL} \\
\texttt{convokit\_casino} & \texttt{SS} & \texttt{cornell\_movie} & \texttt{CONV} & \texttt{soc\_cornell} & \texttt{ROLE} \\
\texttt{convokit\_chromium} & \texttt{CONV} & \texttt{coser} & \texttt{ROLE} & \texttt{soc\_haico} & \texttt{ROLE} \\
\texttt{convokit\_CGA-cmv} & \texttt{SS} & \texttt{dailydialog} & \texttt{CONV} & \texttt{soc\_persona\_conflicts} & \texttt{ROLE} \\
\texttt{convokit\_CGA-cmv-large} & \texttt{SS} & \texttt{education\_dialogue} & \texttt{ROLE} & \texttt{soc\_sotopia\_pi\_bc} & \texttt{ROLE} \\
\texttt{convokit\_CGA-wiki} & \texttt{SS} & \texttt{empathetic} & \texttt{SS} & \texttt{soc\_sotopia\_tom\_silver} & \texttt{ROLE} \\
\texttt{convokit\_emotional-support} & \texttt{SS} & \texttt{hh\_rlhf} & \texttt{EVAL} & \texttt{socsci210} & \texttt{EVAL} \\
\texttt{convokit\_friends} & \texttt{CONV} & \texttt{human\_llm} & \texttt{ROLE} & \texttt{studychat} & \texttt{ROLE} \\
\texttt{convokit\_mediasum} & \texttt{CONV} & \texttt{humanual\_book} & \texttt{ROLE} & \texttt{tom\_characterllm} & \texttt{COG} \\
\texttt{convokit\_npr-2p} & \texttt{CONV} & \texttt{humanual\_chat} & \texttt{ROLE} & \texttt{tom\_fantom} & \texttt{COG} \\
\texttt{convokit\_parliament} & \texttt{CONV} & \texttt{humanual\_email} & \texttt{ROLE} & \texttt{tom\_from\_coser} & \texttt{COG} \\
\texttt{convokit\_persuasion4good} & \texttt{SS} & \texttt{humanual\_news} & \texttt{ROLE} & \texttt{tom\_grimulkan} & \texttt{COG} \\
\texttt{convokit\_reddit-coarse} & \texttt{CONV} & \texttt{humanual\_opinion} & \texttt{ROLE} & \texttt{tom\_hitom} & \texttt{COG} \\
\texttt{convokit\_reddit-corpus-small} & \texttt{CONV} & \texttt{humanual\_politics} & \texttt{ROLE} & \texttt{tom\_mindgames} & \texttt{COG} \\
\texttt{convokit\_small-pool} & \texttt{CONV} & \texttt{lmsys} & \texttt{CONV} & \texttt{tom\_moralstories} & \texttt{COG} \\
\texttt{convokit\_supreme} & \texttt{CONV} & \texttt{mathdial} & \texttt{ROLE} & \texttt{tom\_paratomi} & \texttt{COG} \\
\texttt{convokit\_switchboard} & \texttt{CONV} & \texttt{nectar} & \texttt{EVAL} & \texttt{tom\_socialiqa} & \texttt{COG} \\
\texttt{convokit\_tennis} & \texttt{CONV} & \texttt{oasst1} & \texttt{CONV} & \texttt{tom\_tominli} & \texttt{COG} \\
\texttt{convokit\_wiki-AfD} & \texttt{CONV} & \texttt{oasst2} & \texttt{CONV} & \texttt{wildchat} & \texttt{CONV} \\
\texttt{convokit\_wiki} & \texttt{CONV} & \texttt{prism} & \texttt{EVAL} &  & \\
\bottomrule
\end{tabular*}
\end{table}

\subsection{Persona Profile Coverage}
\label{app:profile-coverage}

We probe how widely the $\approx$19.7M training-row system prompts in \texttt{sft\_processed\_large} cover the space of plausible character profiles. The pipeline is intentionally LLM-free; everything below is regex match plus set lookup against curated lexicons.

\paragraph{Pipeline.}
\begin{itemize*}
    \item \textbf{Extract.} From every parquet shard we pull only the system message + minimal metadata, producing a 3.4~GB ``system-prompt'' parquet companion to \texttt{sft\_processed\_large}.
    \item \textbf{Strip context.} Each prompt is split into sentences; any sentence containing a goal pattern (``Your goal is to \ldots'', ``Your aim is to \ldots'', ``You're trying to \ldots'', ``You want to \ldots'', etc.) is dropped via regex, leaving the residual character description.
    \item \textbf{Match.} The residual is tokenized (whole-word, case-insensitive) and matched against five hand-curated lexicons:
    \begin{itemize*}
        \item \textit{Occupations} (414 terms; SOC / O*NET-derived professional and community roles)
        \item \textit{Traits} (490 terms; Goldberg/Saucier/John-Srivastava Big-Five markers, communication-style adjectives, emotion words, cognitive-style terms)
        \item \textit{Demographics} (358 terms; age, education, marital, family role, gender, race/ethnicity, religion, geography, sexual orientation)
        \item \textit{Register / voice} (86 terms; formality, voice, register markers)
        \item \textit{Settings / contexts} (227 terms; physical locations, online platforms, specific events, scene/genre)
    \end{itemize*}
    \item \textbf{Fingerprint.} The sorted union of matched (occupation, trait, demographic, register) terms forms a \texttt{profile\_fingerprint}; settings are kept separately as context. Two prompts with identical fingerprints are treated as the same intrinsic character regardless of differing topical context.
\end{itemize*}

\paragraph{Headline counts (full coverage, no sampling).} Across 19{,}669{,}019 system prompts the pipeline finds:
\begin{itemize*}
    \item \textbf{1{,}090{,}417} unique profile fingerprints (vs. 2.91M unique full-prompt strings; goal/context account for $\approx$62\% of the apparent-uniqueness gap).
    \item \textbf{299} distinct occupations matched, \textbf{460} distinct traits, \textbf{400} distinct demographic markers.
    \item \textbf{63{,}199} distinct (occupation, trait) co-occurrence pairs populated across the top-25 occupations and top-25 traits.
\end{itemize*}

\paragraph{Three persona-uniqueness regimes.}
The 62 datasets split cleanly into three regimes by the ratio of unique system prompts to row count:
\textit{(i) Per-record unique ($\geq$99\%):} all 22 ConvoKit-back-generated corpora, plus \texttt{cornell\_movie}, \texttt{coser}, \texttt{mathdial}, \texttt{prism}, \texttt{studychat}, \texttt{education\_dialogue}, \texttt{soc\_persona\_conflicts}, and the SOTOPIA-style sets --- each conversation has a bespoke persona generated independently.
\textit{(ii) Per-user / templated (5--65\%):} \texttt{humanual\_book/email/news/opinion/politics} use a fixed-persona-per-user (209 distinct customers in \texttt{humanual\_book}; 8K Medium / YouTube users in \texttt{humanual\_news}); \texttt{dailydialog} (58\% unique) and \texttt{empathetic} (50\% unique) are templated by emotion + situation; \texttt{wildchat}, \texttt{lmsys}, \texttt{oasst1/2} cluster around 58--60\% unique.
\textit{(iii) Boilerplate single persona:} \texttt{rm\_r1\_sft}, \texttt{tom\_characterllm}, \texttt{tom\_moralstories}, \texttt{tom\_socialiqa}, \texttt{tom\_from\_coser}, \texttt{tom\_mindgames}, \texttt{tom\_grimulkan}, \texttt{tom\_tominli}, and \texttt{humanual\_chat} each carry a single fixed system prompt across thousands of rows --- once any other dataset has matched a similar profile, these contribute zero new fingerprints.

\begin{figure}[t]
    \centering
    \includegraphics[width=\textwidth]{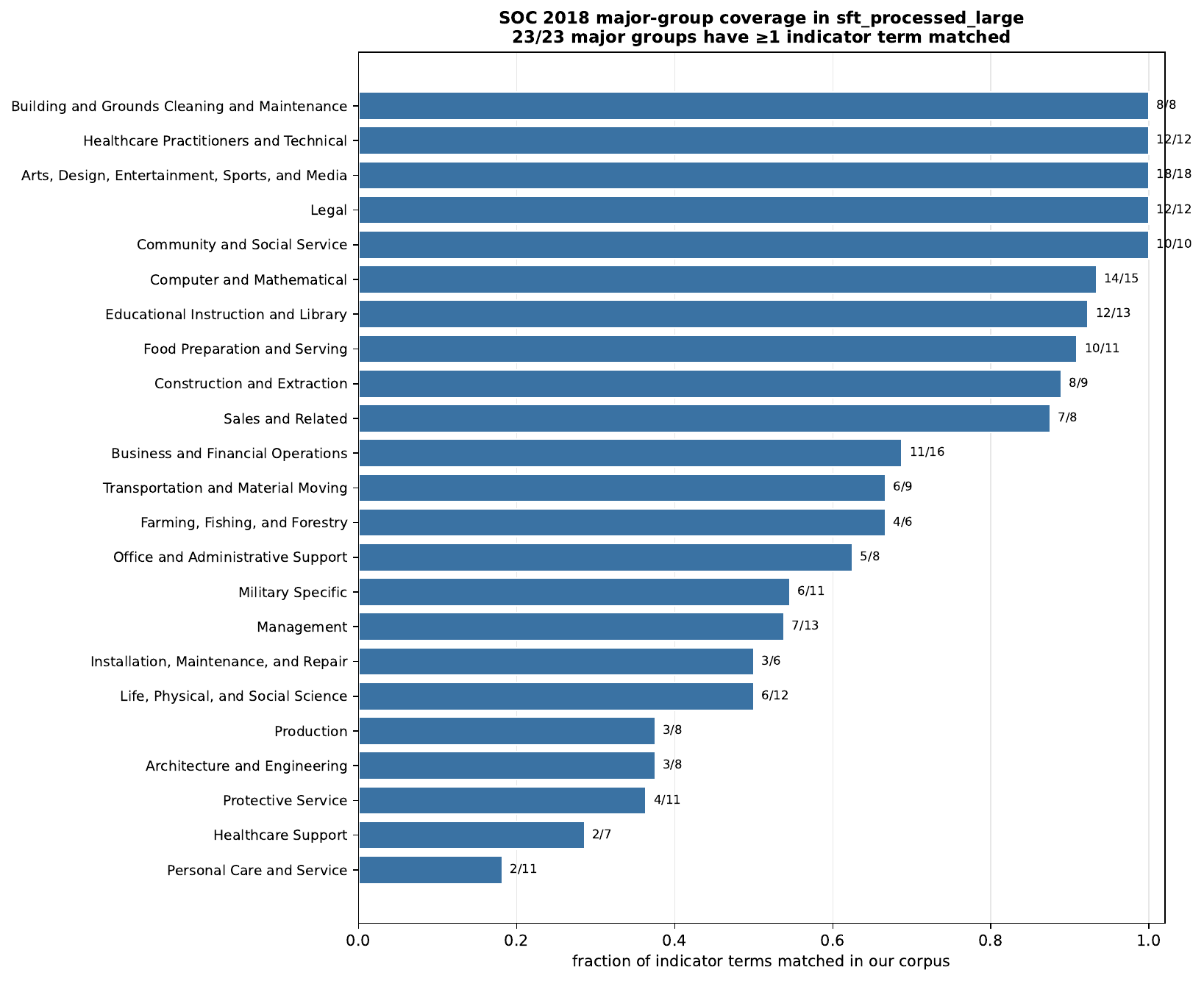}
    \caption{Coverage of the 23 SOC 2018 major occupational groups by \texttt{sft\_processed\_large}. For each group we draw 6--18 indicator titles from O*NET; the bar shows the fraction of those titles that appear in the corpus's matched-occupation set. All 23 groups have $\geq$1 match.}
    \label{fig:soc-coverage}
\end{figure}

\paragraph{SOC coverage.}
We map the matched-occupation set onto the 23 major groups of the BLS Standard Occupational Classification (2018). Each group is represented by 6--18 indicator titles drawn from O*NET; an indicator is ``covered'' if its term appears at least once in the corpus's matched occupations. \Cref{fig:soc-coverage} reports per-group coverage. All 23 groups have at least one matched indicator. Coverage is most complete on \emph{Legal}, \emph{Healthcare Practitioners}, \emph{Arts / Design / Entertainment / Sports / Media}, \emph{Education}, and \emph{Computer \& Math} roles (90--100\%); it drops on manual-trade and personal-care occupations (\emph{Personal Care and Service} 18\%, \emph{Healthcare Support} 29\%, \emph{Production} 38\%, \emph{Architecture \& Engineering} 38\%) where \texttt{sft\_processed\_large} skews away by construction --- the corpus is built around conversational personas, not skilled-trade workers.

\begin{figure}[t]
    \centering
    \includegraphics[width=\textwidth]{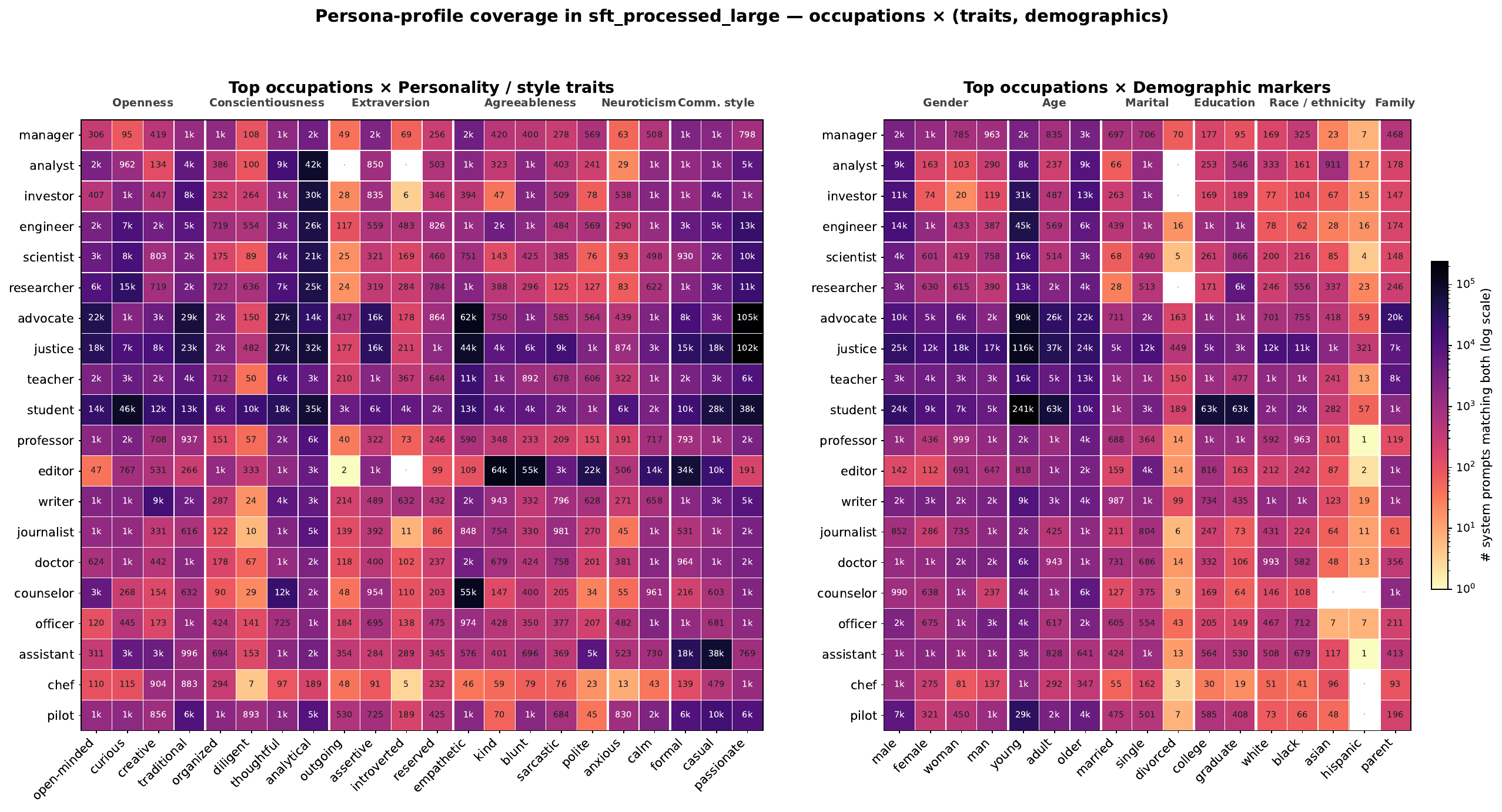}
    \caption{Persona-profile coverage along three axes simultaneously across the 19.7M system prompts of \texttt{sft\_processed\_large}.
    Rows: 20 representative occupations chosen one to two per BLS SOC 2018 major group (management $\to$ business / sciences $\to$ legal $\to$ education $\to$ arts/media $\to$ healthcare $\to$ services $\to$ transportation).
    \emph{Left panel:} 22 traits organized by Big-Five factor (Openness, Conscientiousness, Extraversion, Agreeableness, Neuroticism) plus communication style, with both poles represented (e.g., \emph{introverted} vs.\ \emph{outgoing}; \emph{anxious} vs.\ \emph{calm}; \emph{traditional} vs.\ \emph{open-minded}).
    \emph{Right panel:} 18 demographic markers grouped by gender / age / marital / education / race-ethnicity / family. Cell value $=$ \# system prompts that match \emph{both} the row term and the column term; color is log-scaled (\texttt{magma\_r}). 437/440 (left) and 333/360 (right) cells carry non-zero co-occurrence.}
    \label{fig:occ-trait-heatmap}
\end{figure}

\paragraph{Co-occurrence breadth across personality and demographic axes.}
\Cref{fig:occ-trait-heatmap} reports a two-panel co-occurrence view of the corpus along three persona axes simultaneously: \emph{(left)}~20 SOC-aligned representative occupations $\times$ 22 traits organized by Big-Five dimension (Openness / Conscientiousness / Extraversion / Agreeableness / Neuroticism) plus communication style; \emph{(right)}~the same 20 occupations $\times$ 18 demographic markers grouped by gender / age / marital / education / race-ethnicity / family role. Rows are SOC-aligned (one to two representative occupations per major BLS group) rather than raw most-frequent terms, so the row ordering is interpretable as a sweep across the labor market. Columns on the left panel cover both poles of each Big-Five factor (e.g., \emph{introverted} vs \emph{outgoing}, \emph{anxious} vs \emph{calm}, \emph{traditional} vs \emph{open-minded}). Cell counts are the number of system prompts whose residual character description matches \emph{both} the row and the column term; color is log-scaled (\texttt{magma\_r}). The matrix is densely populated --- 437/440 trait cells and 333/360 demographic cells carry non-zero co-occurrence. Dominant cells include \texttt{advocate} $\times$ \texttt{passionate} (105K), \texttt{justice} $\times$ \texttt{passionate} (102K), \texttt{student} $\times$ \texttt{young} (241K), \texttt{justice} $\times$ \texttt{young} (116K), \texttt{advocate} $\times$ \texttt{young} (90K), and \texttt{student} $\times$ \{\texttt{college}, \texttt{graduate}\} (63K each), reflecting the dominant socsci210 / advocacy / student populations.

\begin{figure}[t]
    \centering
    \includegraphics[width=\textwidth]{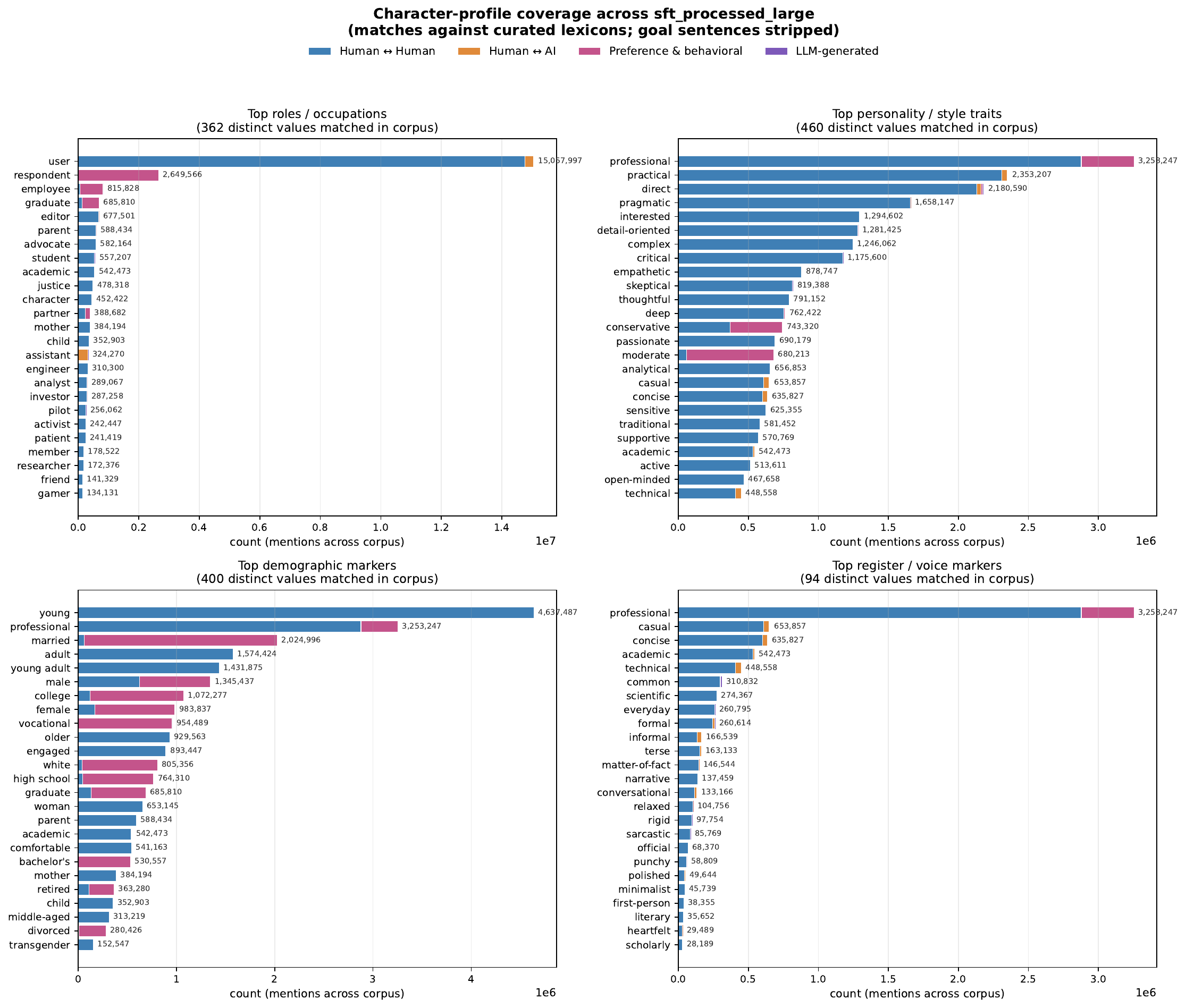}
    \caption{Top-25 most frequent terms in each profile dimension (roles / traits / demographics / register) across \texttt{sft\_processed\_large}. Bars are stacked by Tier-1 origin (blue: human-human; orange: human-AI; pink: preference / behavioral; violet: LLM-generated); panel titles report the total \# distinct values matched corpus-wide.}
    \label{fig:profile-coverage}
\end{figure}

\paragraph{Per-dimension coverage.}
\Cref{fig:profile-coverage} ranks the top-25 most frequent terms within each profile dimension (occupations, traits, demographics, register), color-stacked by Tier-1 origin. The figure highlights both the \emph{volume} contributed by each Tier-1 (almost everything is dominated by human$\leftrightarrow$human content, with socsci210 the second-largest contributor in pink) and the \emph{breadth} of distinct values matched in each dimension (panel titles).

\subsection{Data Processing}
\label{app:data-processing}

All sources are converted into a unified schema with three fields:
(1)~\texttt{user\_id}: anonymized user identifier;
(2)~\texttt{conversations}: a list of multi-turn dialogues, each containing timestamped messages with role and content;
(3)~\texttt{user\_meta}: demographic and contextual information where available.
Each dataset is processed through a single pipeline that encodes per-corpus role assignment, turn linearization, and state labels.

\paragraph{Synthesizing missing social context.}
Many of the source datasets arrive without an explicit system prompt --- open-domain AI chat logs (WildChat, LMSYS-Chat-1M, OASST), preference data (HH-RLHF, Nectar), and most ConvoKit corpora carry only the raw turns.
For these we synthesize a per-record system-prompt header using \texttt{gpt-5.4-mini-2026-03-17}.
Per conversation we generate \emph{two} prose system prompts (one per side, so the same record yields two training rows --- original and role-swapped, see below), each 2--5 sentences starting with ``You are\ldots'' and describing the simulated party's role, goal, background, and conversational style.
A deterministic per-side mode selector (sha256 of \texttt{record\_id::side}, threshold $0.20$) sends $\sim$20\% of sides to a \emph{detailed} mode (fuller backstory) and $\sim$80\% to a \emph{short} mode (style-only); same record always yields the same modes across re-runs so the corpus is exactly reproducible.
To prevent leakage, the generator sees only the first 60\% of turns ($\geq 3$ turns minimum) and never sees the conversation's outcome metadata (\texttt{state} field, escalation flags, downstream labels), so the synthesized persona cannot foreshadow the trajectory.
Datasets that natively carry persona or scenario information --- CoSER (literary characters), SOTOPIA / HAICosystem / Persona-Conflicts (goal-conditioned scenarios), Cognitive Genome and AlignX (persona-grounded simulation), and the six HUMANUAL domains (per-user backgrounds) --- retain their original system prompts without modification.

\paragraph{Role swap and filtering.}
For human and AI conversation sources, we apply a \emph{role-swap} protocol: the user side of each conversation becomes the training target, with the assistant responses serving as context.
This trains the model to produce human-like user behavior rather than assistant-like responses.
Filtering removes conversations with fewer than 2~turns, deduplicates at the conversation level using MinHash, and filters toxic or personally identifiable content.

\subsection{Train / Val / Test Split}
\label{app:splits}

The 21.4M-row corpus is partitioned into three splits with deliberately different distributional properties (\cref{tab:split-summary}): \emph{train} (21,195,418 rows), \emph{val} (28,408 rows, in-distribution), and \emph{test} (128,045 rows, profile-disjoint where feasible). All three splits are released as the HuggingFace dataset \midtrainingdata, with the per-dataset manifest \texttt{\_split\_manifest.json} stored alongside the parquet shards. The split is fully deterministic from the rules below: anyone can reproduce it bit-exact by running our published splitting script on the released corpus.

\paragraph{Profile fingerprint, the unit we hold out.}
For each system prompt we strip goal sentences via regex (e.g., ``Your goal is to \ldots'', ``You're trying to \ldots'') and match the residual against five hand-curated lexicons (occupations, personality / communication-style traits, demographics, register, settings; \cref{app:profile-coverage}). The sorted union of matched (occupation, trait, demographic, register) terms forms a \emph{profile fingerprint}; settings are recorded separately as context. Two prompts with identical fingerprints describe the same intrinsic character regardless of differing topical framing.

\paragraph{Test split (out-of-distribution, profile-disjoint).}
For each dataset $d$, with $n_d$ rows and $u_d$ unique profile fingerprints:
{\small\begin{verbatim}
target_test = max(100, floor(0.005 * n_rows))
if d in {humanual_book: 15, socsci210: 30, dailydialog: 30}:
    # deterministic-N-profile override (pre-baked in split_index.json)
    pick N profiles with lowest sha256(fingerprint); all their rows -> test
elif u_d >= 50:
    # profile_hash mode: 47 datasets
    bucket_pct = target_test / n_rows
    for each row r:
        key   = profile_fingerprint(r)
        score = int(sha256(key)[:8], 16) / 2^32
        r -> test iff score < bucket_pct
else:
    # record_hash mode: 16 datasets (boilerplate / few-profile / no-system-prompt)
    bucket_pct = target_test / n_rows
    for each row r:
        key   = f"{ds_name}::{shard_filename}::{row_idx}"
        score = int(sha256(key)[:8], 16) / 2^32
        r -> test iff score < bucket_pct
\end{verbatim}}
For the 47 profile-hash datasets (117{,}568 of 128{,}045 test rows, 91.8\%), every profile fingerprint in \texttt{test\_shard\_*.parquet} is \emph{disjoint} from every fingerprint in \texttt{train\_shard\_*.parquet}, by construction of the per-row hash. Train holds \emph{1{,}035{,}409} unique profiles, test holds \emph{6{,}281}, with only 115 overlapping (all attributable to the 16 record-hash datasets); 99.98\% of distinct test profiles are unseen in train.

Three datasets received deterministic-N-profile overrides because the default 0.5\%-bucket rule gave too few or zero test rows under hash variance (\texttt{humanual\_book}: 0 rows; \texttt{dailydialog}: 14 rows; \texttt{socsci210}: 4{,}726 rows of a 13K target). The remaining 16 record-hash datasets share profiles between train and test --- their profile spaces are too small to support disjoint splitting (single-template ToM datasets, \texttt{rm\_r1\_sft}, \texttt{psych101}, \texttt{humanual\_chat}; plus 6 datasets where the system message is empty or absent so no fingerprint can be extracted: \texttt{hh\_rlhf}, \texttt{human\_llm}, \texttt{nectar}, \texttt{tom\_fantom}, \texttt{tom\_hitom}, \texttt{tom\_paratomi}). For those, the test holdout is row-level, not profile-level.

\paragraph{Val split (in-distribution, sampled from train).}
After test rows are removed, $n_\text{val,d} = \max(30, \min(5000, \lfloor 0.005 \cdot n_\text{train,d} \rfloor))$ rows per dataset are drawn uniformly at random from the remaining train shards using a deterministic seed $\texttt{sha256(ds\_name)[:8]}$. The 5{,}000-row cap prevents the largest datasets (\texttt{alignx\_v2} at 14.7M rows; \texttt{socsci210} at 2.6M) from dominating the val set. Every val row's profile fingerprint also appears in train; this is intentional --- val measures in-distribution loss, not generalization.

\paragraph{What each split measures.}
\begin{itemize*}
    \item \textbf{Val} (in-distribution, 28K rows): checkpoint selection, learning-rate scheduling, detecting train-distribution overfit. Loss on val tracks closely with train loss.
    \item \textbf{Test} (out-of-distribution, 128K rows): detecting whether the model has acquired a behavioral capability or merely memorized (profile, behavior) mappings seen in training. A growing gap between val and test loss would indicate persona memorization rather than capability acquisition.
\end{itemize*}
We report internal loss curves on both at midtraining time but final benchmark claims in the paper come from the external evaluation suite (\cref{sec:soul}), which is held out at the source-dataset level.

\paragraph{Axis-aligned training data for every \soulindex task.}
For every \soulindex task we curate training data targeting the same Axis. The mapping splits into two regimes:
(a) \emph{Direct} --- the benchmark itself ships with a training split, and we use it as-is, enforcing example-disjointness against the test rows by the hashing rules above; and
(b) \emph{Proxy} --- the benchmark provides only an evaluation split, and we substitute closely related data from the same benchmark family or behavioral domain as a proxy training source.
\Cref{tab:soul-train-map} gives the per-task mapping with row counts; \cref{tab:split-summary} gives the per-dataset split breakdown; the published \texttt{split\_index.json} fixes every bucket boundary so the assignments are bit-exact reproducible.

\begin{table*}[t]
\centering
\small
\caption{Per-\soulindex-task training-data mapping. \emph{Direct} = the benchmark's own training split (with our hash-based test holdout). \emph{Proxy} = curated from a closely related source when the benchmark has no native training set. Row counts are train rows in \midtrainingdata.}
\label{tab:soul-train-map}
\setlength{\tabcolsep}{4pt}
\renewcommand{\arraystretch}{1.05}
\begin{tabular}{llllr}
\toprule
\textbf{Axis} & \textbf{\soulindex task} & \textbf{Training source} & \textbf{Type} & \textbf{Rows} \\
\midrule
\multirow{4}{*}{\texttt{CONV}}
  & UserLLM            & WildChat / PRISM user-side splits        & Direct & --- \\
  & MirrorBench        & Multi-turn AI-dialogue subset            & Proxy  & --- \\
  & Humanual-Chat      & \texttt{humanual\_chat}                  & Direct & 23{,}385 \\
  & SimArena-Doc       & SimulatorArena (Doc) train split         & Direct & --- \\
\midrule
\texttt{SS}
  & Sotopia-Hard       & SOTOPIA-$\pi$ scenarios                  & Proxy  & 2{,}363 \\
\midrule
\multirow{4}{*}{\texttt{COG}}
  & Fantom             & \texttt{tom\_fantom} train split         & Direct & 894 \\
  & Hitom              & \texttt{tom\_hitom} train split          & Direct & 899 \\
  & Paratomi           & \texttt{tom\_paratomi} train split       & Direct & --- \\
  & Social-R1          & ToM pool (\texttt{tom\_*})               & Proxy  & --- \\
\midrule
\multirow{9}{*}{\texttt{ROLE}}
  & Coser              & CoSER train split                        & Direct & 114{,}831 \\
  & Lifechoices        & LifeChoices train split                  & Direct & --- \\
  & Twinvoice          & Cognitive Genome persona dialogue        & Proxy  & --- \\
  & BehaviorChain      & Online-shopping persona-action traces    & Proxy  & --- \\
  & SimArena-Math      & SimulatorArena (Math) train split        & Direct & --- \\
  & Mistakes           & EduDialogue / StudyChat / MathDial       & Proxy  & 29{,}995 \\
  & Humanual-Email     & \texttt{humanual\_email}                 & Direct & 6{,}322 \\
  & Humanual-News      & \texttt{humanual\_news}                  & Direct & 49{,}148 \\
  & Humanual-Politics  & \texttt{humanual\_politics}              & Direct & 45{,}395 \\
\midrule
\multirow{5}{*}{\texttt{EVAL}}
  & AlignX             & \texttt{alignx\_v2}                      & Direct & 14.7M \\
  & HumanLLM           & Cognitive Genome persona subset          & Direct & --- \\
  & SocSci210          & \texttt{socsci210}                       & Direct & 2{,}618{,}745 \\
  & Humanual-Book      & \texttt{humanual\_book}                  & Direct & 31{,}931 \\
  & Humanual-Opinion   & \texttt{humanual\_opinion}               & Direct & 38{,}613 \\
\midrule
\multicolumn{2}{l}{\textit{Aux.\ preference / reward data (no Index task)}}
                       & HH-RLHF, Nectar, PRISM, RM-R1-SFT        & ---    & --- \\
\bottomrule
\end{tabular}
\end{table*}

\paragraph{Aggregating per-dataset loss.}
Because per-dataset test set sizes vary by 3 orders of magnitude (alignx\_v2 = 80{,}622 rows; the smallest = 42 rows), a row-pooled average would let a handful of big datasets dominate the per-skill number. We instead use the \textbf{geometric mean of per-dataset perplexities} as our per-skill aggregate, which weights each dataset equally:
\[
\mathrm{PPL}_{\text{dim}} \;=\; \mathrm{geomean}_{d \in D_{\text{dim}}}\!\big(\mathrm{PPL}_d\big) \;=\; \exp\!\Big(\tfrac{1}{|D_{\text{dim}}|}\!\sum_{d \in D_{\text{dim}}} \mathrm{NLL}_d\Big),
\]
where $\mathrm{NLL}_d$ is the per-token mean cross-entropy on dataset $d$'s held-out test split. This is the convention used in \cref{tab:ppl-per-skill}, \cref{fig:scaling-law}, and \cref{fig:skill-heatmap}.

\paragraph{Reproducing the split.}
The full pipeline is published in our code repository. To reproduce bit-exact:
\begin{enumerate*}
    \item Download \midtrainingdata and run \texttt{scripts/precompute\_split\_index.py} to compute fingerprints + per-dataset bucket boundaries (output: \texttt{split\_index.json}, $\sim$0.65 MB).
    \item Apply via \texttt{scripts/split\_apply\_index.py}, which streams each parquet shard, looks up bucket assignments, and writes \texttt{train\_shard\_NNN.parquet}, \texttt{val\_shard\_000.parquet}, \texttt{test\_shard\_NNN.parquet} per dataset.
    \item The deterministic-N-profile overrides for the three flagged datasets are pre-baked into \texttt{split\_index.json}.
\end{enumerate*}
Or skip regeneration by loading the released split directly:
{\small\begin{verbatim}
load_dataset("cmu-lti/osim-mid-training", "<config>", split="train")
\end{verbatim}}
with \texttt{split="train"}, \texttt{"val"}, or \texttt{"test"}.

\definecolor{recordbg}{HTML}{F0EDE5}
\begin{table}[ht]
\caption{Per-dataset train / val / test split of \midtrainingdata, sorted by total rows. $P_{\text{tr}}$ and $P_{\text{te}}$ are the number of unique profile fingerprints in train and test respectively. For the 46 profile-disjoint datasets these sets are non-overlapping by construction. \colorbox{recordbg}{Shaded rows} are the 16 \texttt{record\_hash} datasets where the test set holds out \emph{rows} rather than profiles, so $P_{\text{te}} \subseteq P_{\text{tr}}$ (test profiles are a subset of train profiles, not disjoint). ``---'' indicates no profile signal (e.g., the dataset's system prompt is empty or absent: \texttt{hh\_rlhf}, \texttt{human\_llm}, \texttt{nectar}, \texttt{tom\_fantom}, \texttt{tom\_hitom}, \texttt{tom\_paratomi}).}
\label{tab:split-summary}
\centering
\scriptsize
\setlength{\tabcolsep}{3pt}
\renewcommand{\arraystretch}{1.05}
\resizebox{\textwidth}{!}{%
\begin{tabular}{r l r r r r r @{\hskip 8pt} r l r r r r r}
\toprule
\# & \textbf{Dataset} & \textbf{Train} & \textbf{Val} & \textbf{Test} & \textbf{$P_{\text{tr}}$} & \textbf{$P_{\text{te}}$} & \# & \textbf{Dataset} & \textbf{Train} & \textbf{Val} & \textbf{Test} & \textbf{$P_{\text{tr}}$} & \textbf{$P_{\text{te}}$} \\
\midrule
1 & alignx\_v2 & 14{,}649{,}170 & 5{,}000 & 80{,}622 & 501{,}808 & 2{,}569 & 33 & oasst1 & 13{,}447 & 67 & 61 & 3{,}303 & 25 \\
2 & socsci210 & 2{,}618{,}745 & 5{,}000 & 21{,}053 & 1{,}788 & 30 & 34 & convokit\_tennis-corpus & 12{,}751 & 64 & 61 & 6{,}442 & 45 \\
\cellcolor{recordbg}3 & \cellcolor{recordbg}human\_llm & \cellcolor{recordbg}1{,}316{,}829 & \cellcolor{recordbg}5{,}000 & \cellcolor{recordbg}6{,}757 & \cellcolor{recordbg}--- & \cellcolor{recordbg}--- & 35 & convokit\_friends-corpus & 11{,}210 & 56 & 110 & 5{,}479 & 47 \\
4 & convokit\_wiki-AfD & 580{,}752 & 2{,}918 & 1{,}186 & 114{,}996 & 569 & \cellcolor{recordbg}36 & \cellcolor{recordbg}tom\_mindgames & \cellcolor{recordbg}11{,}033 & \cellcolor{recordbg}55 & \cellcolor{recordbg}86 & \cellcolor{recordbg}1 & \cellcolor{recordbg}1 \\
5 & convokit\_wikiconv-2018 & 231{,}829 & 1{,}164 & 2{,}635 & 65{,}107 & 320 & 37 & convokit\_parliament-corpus & 9{,}449 & 47 & 56 & 3{,}686 & 39 \\
\cellcolor{recordbg}6 & \cellcolor{recordbg}nectar & \cellcolor{recordbg}180{,}778 & \cellcolor{recordbg}908 & \cellcolor{recordbg}932 & \cellcolor{recordbg}--- & \cellcolor{recordbg}--- & \cellcolor{recordbg}38 & \cellcolor{recordbg}rm\_r1\_sft & \cellcolor{recordbg}8{,}591 & \cellcolor{recordbg}43 & \cellcolor{recordbg}119 & \cellcolor{recordbg}1 & \cellcolor{recordbg}1 \\
\cellcolor{recordbg}7 & \cellcolor{recordbg}hh\_rlhf & \cellcolor{recordbg}166{,}878 & \cellcolor{recordbg}838 & \cellcolor{recordbg}860 & \cellcolor{recordbg}--- & \cellcolor{recordbg}--- & 39 & convokit\_reddit-corpus-small & 8{,}375 & 42 & 67 & 4{,}736 & 58 \\
8 & wildchat & 165{,}346 & 830 & 1{,}260 & 35{,}118 & 184 & 40 & prism & 7{,}908 & 39 & 61 & 4{,}248 & 48 \\
9 & cornell\_movie & 163{,}927 & 823 & 1{,}124 & 75{,}003 & 387 & 41 & convokit\_reddit-coarse-disc. & 6{,}912 & 34 & 66 & 3{,}971 & 59 \\
10 & coser & 114{,}831 & 577 & 658 & 65{,}443 & 369 & 42 & humanual\_email & 6{,}322 & 31 & 154 & 387 & 7 \\
11 & lmsys & 79{,}929 & 401 & 170 & 1{,}590 & 10 & 43 & convokit\_CGA-cmv & 5{,}968 & 30 & 72 & 3{,}338 & 52 \\
\cellcolor{recordbg}12 & \cellcolor{recordbg}tom\_from\_coser & \cellcolor{recordbg}76{,}747 & \cellcolor{recordbg}385 & \cellcolor{recordbg}416 & \cellcolor{recordbg}1 & \cellcolor{recordbg}1 & \cellcolor{recordbg}44 & \cellcolor{recordbg}tom\_tominli & \cellcolor{recordbg}5{,}872 & \cellcolor{recordbg}30 & \cellcolor{recordbg}92 & \cellcolor{recordbg}1 & \cellcolor{recordbg}1 \\
13 & convokit\_chromium-corpus & 70{,}365 & 353 & 160 & 20{,}897 & 115 & 45 & convokit\_emotional-support & 5{,}061 & 30 & 109 & 3{,}371 & 73 \\
14 & convokit\_wiki-corpus & 59{,}912 & 301 & 205 & 24{,}599 & 121 & 46 & convokit\_CGA-wiki & 4{,}916 & 30 & 42 & 2{,}276 & 40 \\
\cellcolor{recordbg}15 & \cellcolor{recordbg}psych101 & \cellcolor{recordbg}57{,}653 & \cellcolor{recordbg}289 & \cellcolor{recordbg}289 & \cellcolor{recordbg}9 & \cellcolor{recordbg}7 & 47 & convokit\_small-pool & 4{,}620 & 30 & 82 & 2{,}881 & 55 \\
16 & humanual\_news & 49{,}148 & 246 & 195 & 8{,}081 & 34 & 48 & convokit\_switchboard-corpus & 4{,}487 & 30 & 103 & 3{,}211 & 66 \\
17 & empathetic & 45{,}871 & 230 & 42 & 3{,}391 & 17 & 49 & convokit\_casino-corpus & 3{,}992 & 30 & 98 & 1{,}907 & 49 \\
18 & humanual\_politics & 45{,}395 & 228 & 295 & 5{,}230 & 24 & 50 & convokit\_winning-args-corpus & 3{,}995 & 30 & 89 & 2{,}526 & 68 \\
19 & convokit\_mediasum-corpus & 38{,}139 & 191 & 1{,}646 & 20{,}025 & 103 & 51 & convokit\_persuasion4good & 3{,}887 & 30 & 151 & 2{,}530 & 70 \\
20 & humanual\_opinion & 38{,}613 & 194 & 161 & 4{,}485 & 20 & 52 & soc\_haico & 3{,}374 & 30 & 96 & 507 & 15 \\
21 & convokit\_npr-2p-corpus & 37{,}532 & 188 & 772 & 22{,}199 & 136 & 53 & soc\_persona\_conflicts & 3{,}264 & 30 & 102 & 3{,}293 & 102 \\
22 & humanual\_book & 31{,}931 & 160 & 2{,}571 & 166 & 15 & 54 & soc\_cornell & 2{,}876 & 30 & 94 & 2{,}756 & 84 \\
\cellcolor{recordbg}23 & \cellcolor{recordbg}tom\_socialiqa & \cellcolor{recordbg}33{,}067 & \cellcolor{recordbg}166 & \cellcolor{recordbg}177 & \cellcolor{recordbg}1 & \cellcolor{recordbg}1 & 55 & mathdial & 2{,}771 & 30 & 60 & 585 & 18 \\
24 & education\_dialogue & 28{,}026 & 140 & 133 & 15{,}803 & 93 & 56 & studychat & 1{,}969 & 30 & 215 & 704 & 39 \\
25 & dailydialog & 24{,}448 & 122 & 126 & 1{,}415 & 30 & 57 & soc\_sotopia\_tom\_silver & 1{,}296 & 30 & 115 & 895 & 83 \\
\cellcolor{recordbg}26 & \cellcolor{recordbg}tom\_moralstories & \cellcolor{recordbg}23{,}754 & \cellcolor{recordbg}119 & \cellcolor{recordbg}127 & \cellcolor{recordbg}1 & \cellcolor{recordbg}1 & 58 & soc\_sotopia\_pi\_bc & 1{,}067 & 30 & 97 & 895 & 81 \\
\cellcolor{recordbg}27 & \cellcolor{recordbg}humanual\_chat & \cellcolor{recordbg}23{,}385 & \cellcolor{recordbg}117 & \cellcolor{recordbg}120 & \cellcolor{recordbg}1 & \cellcolor{recordbg}1 & \cellcolor{recordbg}59 & \cellcolor{recordbg}tom\_fantom & \cellcolor{recordbg}894 & \cellcolor{recordbg}30 & \cellcolor{recordbg}100 & \cellcolor{recordbg}--- & \cellcolor{recordbg}--- \\
28 & oasst2 & 20{,}224 & 101 & 56 & 4{,}922 & 29 & \cellcolor{recordbg}60 & \cellcolor{recordbg}tom\_hitom & \cellcolor{recordbg}899 & \cellcolor{recordbg}30 & \cellcolor{recordbg}95 & \cellcolor{recordbg}--- & \cellcolor{recordbg}--- \\
29 & convokit\_CGA-cmv-large & 17{,}304 & 86 & 106 & 8{,}878 & 64 & \cellcolor{recordbg}61 & \cellcolor{recordbg}tom\_paratomi & \cellcolor{recordbg}903 & \cellcolor{recordbg}30 & \cellcolor{recordbg}91 & \cellcolor{recordbg}--- & \cellcolor{recordbg}--- \\
30 & convokit\_IDEA-NTHU-tweets & 15{,}910 & 79 & 75 & 8{,}321 & 42 & \cellcolor{recordbg}62 & \cellcolor{recordbg}tom\_grimulkan & \cellcolor{recordbg}404 & \cellcolor{recordbg}30 & \cellcolor{recordbg}105 & \cellcolor{recordbg}1 & \cellcolor{recordbg}1 \\
31 & convokit\_supreme-corpus & 15{,}340 & 77 & 55 & 6{,}198 & 45 &  &  &  &  &  &  &  \\
\cellcolor{recordbg}32 & \cellcolor{recordbg}tom\_characterllm & \cellcolor{recordbg}13{,}838 & \cellcolor{recordbg}69 & \cellcolor{recordbg}111 & \cellcolor{recordbg}1 & \cellcolor{recordbg}1 &  &  &  &  &  &  &   \\
\midrule
\multicolumn{7}{r}{\textbf{Total rows}: train=21{,}194{,}129, val=28{,}378, test=127{,}944} & & \multicolumn{6}{l}{\textbf{Total $P_{\text{tr}}$}=1{,}035{,}067, \textbf{$P_{\text{te}}$}=6{,}253 (overlap=115; corpus-unique=1{,}041{,}205)} \\
\bottomrule
\end{tabular}}
\end{table}

\section{\soulindex Task Details}
\label{app:soul-index}

\Cref{tab:soul-index} lists every \soulindex task with its parent \soul Axis, format, and metric. Per-task descriptions follow.

\begin{table}[!t]
\caption{The \soulindex evaluation suite: 23~tasks across the 5~\soul Axes. Each task targets one Axis. \textbf{Format}: D = discriminative, G = generative.}
\label{tab:soul-index}
\centering
\small
\begin{tabular}{llll}
\toprule
\textbf{\soul Axis} & \textbf{Task} & \textbf{Format} & \textbf{Metric} \\
\midrule
\multirow{4}{*}{\texttt{CONV}}
  & UserLLM~\citep{naous2025userllm}            & G (Single-turn)  & Accuracy \\
  & MirrorBench~\citep{hathidara2026mirrorbench}& G (Multi-turn)   & Diversity + LLM \\
  & Humanual-Chat~\citep{wu2026humanlm}         & G                & LLM Judge \\
  & SimArena-Doc~\citep{yao2025simulatorarena}  & G (Multi-turn)   & Human align. \\
\midrule
\texttt{SS}
  & Sotopia-Hard~\citep{zhou2024sotopia}        & G (Multi-turn)   & LLM Judge \\
\midrule
\multirow{4}{*}{\texttt{COG}}
  & Fantom~\citep{kim2023fantom}                & D (MCQ + Open)   & Accuracy \\
  & Hitom~\citep{wu2023hitom}                   & D (MCQ)          & Accuracy \\
  & Paratomi~\citep{sclar2023paratomi}          & D (QA)           & Accuracy \\
  & Social-R1~\citep{wu2026socialr1}            & D (MCQ)          & Accuracy \\
\midrule
\multirow{9}{*}{\texttt{ROLE}}
  & Coser~\citep{wang2025coser}                 & G (Multi-turn)   & LLM Judge \\
  & Lifechoices~\citep{lu2024lifechoices}       & D (MCQ)          & Accuracy \\
  & Twinvoice~\citep{du2025twinvoice}           & D (Binary)       & Accuracy \\
  & BehaviorChain                               & D (MCQ)          & Accuracy \\
  & SimArena-Math~\citep{yao2025simulatorarena} & G (Multi-turn)   & Human align. \\
  & Mistakes~\citep{ross2025mistakes}           & D (MCQ)          & Accuracy \\
  & Humanual-Email~\citep{wu2026humanlm}        & G                & LLM Judge \\
  & Humanual-News~\citep{wu2026humanlm}         & G                & LLM Judge \\
  & Humanual-Politics~\citep{wu2026humanlm}     & G                & LLM Judge \\
\midrule
\multirow{5}{*}{\texttt{EVAL}}
  & AlignX~\citep{li2025alignx}                 & D (Pref.)        & Accuracy \\
  & HumanLLM~\citep{lei2026cogenome}            & D (Pref.)        & Accuracy \\
  & SocSci210~\citep{kolluri2025socsci}         & D (Rating)       & Correlation \\
  & Humanual-Book~\citep{wu2026humanlm}         & G                & LLM Judge \\
  & Humanual-Opinion~\citep{wu2026humanlm}      & G                & LLM Judge \\
\bottomrule
\end{tabular}
\end{table}

\paragraph{\texttt{CONV} (4 tasks).}
Tasks in this Axis test whether models reproduce fine-grained dimensions of everyday human discourse --- register, turn-taking, conversational style, and online help-seeking.
UserLLM~\citep{naous2025userllm} evaluates single-turn user message generation against \textsc{WildChat} and PRISM references.
MirrorBench~\citep{hathidara2026mirrorbench} measures whether simulated user utterances match the lexical diversity and style of real users in multi-turn interaction.
Humanual-Chat is the \emph{chat} domain of HUMANUAL~\citep{wu2026humanlm}, evaluating fidelity to real chat-conversation traces.
SimArena-Doc~\citep{yao2025simulatorarena} provides annotated human--LLM conversations in document creation, testing whether simulated users match real user behavior in multi-turn task assistance.

\paragraph{\texttt{SS} (1 task).}
Sotopia-Hard~\citep{zhou2024sotopia} places two agents in scenarios requiring negotiation, collaboration, or conflict (e.g., a landlord and tenant negotiating rent), scored across seven social dimensions (goal completion, relationship, knowledge, secret leakage, social rules, financial benefits, believability).

\paragraph{\texttt{COG} (4 tasks).}
This Axis tests reasoning \emph{about} mental states.
Fantom~\citep{kim2023fantom} requires tracking who said what in multi-party conversations.
Hitom~\citep{wu2023hitom} requires nested-belief reasoning (e.g., ``Alice thinks Bob thinks\ldots'').
Paratomi~\citep{sclar2023paratomi} is a paraphrase-robust reformulation of false-belief reasoning that resists surface-form shortcuts.
Social-R1~\citep{wu2026socialr1} is an adversarial benchmark that exposes shortcut reasoning in social cognition, requiring multi-step inference.

\paragraph{\texttt{ROLE} (9 tasks).}
This Axis tests sustaining a stable role across an interaction --- whether the role is a literary character, a persona with specific traits, a student, or a real human user --- and translating that role into faithful next-turn behavior.
Coser~\citep{wang2025coser} requires sustaining a literary character's personality across 20~turns of dialogue.
Lifechoices~\citep{lu2024lifechoices} tests whether models make decisions \emph{as} specific characters would.
Twinvoice~\citep{du2025twinvoice} tests whether a model can identify which response matches a specific individual's communication style.
BehaviorChain tests whether models can predict a specific persona's next action in a sequenced behavioral chain.
SimArena-Math~\citep{yao2025simulatorarena} is the math-tutoring counterpart of SimArena-Doc, with annotated student--tutor traces.
Mistakes~\citep{ross2025mistakes} tests whether models can faithfully reproduce common student errors in K-12 math rather than defaulting to correct answers.
The remaining three tasks are HUMANUAL~\citep{wu2026humanlm} long-form domains kept on the role-play side --- Humanual-Email, Humanual-News, Humanual-Politics --- each scoring whether the model's continuation matches real human-authored text in that genre.

\paragraph{\texttt{EVAL} (5 tasks).}
This Axis tests whether models can \emph{evaluate} like humans --- a capability central to reward modeling and LLM-as-judge pipelines, and broadened here to include long-form persona-grounded judgment.
AlignX~\citep{li2025alignx} measures alignment of model preferences with crowd-sourced human preferences across multiple sub-domains.
HumanLLM~\citep{lei2026cogenome} tests whether models, given a persona profile, predict the same preferences and judgments as the matching real participant.
SocSci210~\citep{kolluri2025socsci} provides participant ratings and social-science judgments: given a character profile (e.g., ``you are a black woman'') and a survey question (e.g., ``On a scale from 1 to 7, how willing would you be to have a partner of the opposite political party?''), the model must produce a rating that correlates with the matching human respondent's.
Humanual-Book and Humanual-Opinion~\citep{wu2026humanlm} are the long-form HUMANUAL domains where the task is judgment-like rather than character-driven (book reviews and opinion writing call for evaluative reasoning), so we score them under the evaluation-as-humans Axis.

\section{Training Details}
\label{app:training}

This section gives the full hyperparameter, mixture, and hardware detail elided from \cref{sec:midtraining}.

\subsection{Midtraining hyperparameters}

\paragraph{Optimizer and schedule.}
All midtraining runs (\odyssimmodel-4B-Mid, \odyssimmodel-4B-I-Mid, the \texttt{+\,Step} controls, and the mix-ratio screens) use AdamW with $\beta_1 = 0.9$, $\beta_2 = 0.999$, weight decay $0.01$, peak learning rate $\eta_\text{peak} = 1\!\times\!10^{-5}$ with linear warm-up over the first 20 optimizer steps and a constant schedule thereafter, gradient-norm clipping at $1.0$, and mixed-precision \texttt{bfloat16}. The context is $16{,}384$ input tokens / $8{,}192$ response tokens.

\paragraph{Batch and parallelism.}
The mini-batch is $1{,}024$ conversations per optimizer step. We shard with FSDP-2, applying full parameter / gradient / optimizer-state sharding across $8$ H100-80GB GPUs (one node), with dynamic token-batching at $\le 49{,}152$ tokens per GPU.

\paragraph{Token budget.}
The default token budget is $10$B training tokens, corresponding to $4{,}500$ optimizer steps at the above batch size. In the current midtraining analysis, compute efficiency is summarized from intermediate checkpoints of the existing runs rather than from a separate token-budget grid.

\paragraph{Dataset mixture and upsampling.}
The default mixture is $100\%$ behavioral data drawn from the \odyssim corpus, with per-dataset upsampling factors that range from $0.03\times$ for the largest source (\texttt{alignx\_v2}, $14.7$M rows) to $5.37\times$ for the smallest single-source shard (\texttt{tom\_grimulkan}, $539$ rows). The mix-ratio screen varies \odyssim:\textsc{StepFun} token ratios of $0{:}100$, $50{:}50$, $70{:}30$, $90{:}10$, and $100{:}0$ under the same optimizer, token budget, and step count.

\paragraph{Generic-mix baseline (\texttt{+\,Step}).}
The \texttt{+\,Step} baseline reuses the recipe above verbatim, swapping the behavioral mixture for the publicly released Step-3.5-Flash SFT corpus~\citep{stepfun2025step35sft}---a general-purpose chat/instruct mix (Q\&A, multi-turn dialogue, tool-use) that is generic but not behavior-irrelevant.
This isolates the marginal contribution of \odyssim's persona-, roleplay-, and ToM-conditioned content beyond what broad chat/instruction midtraining already provides.

\paragraph{Excluded baselines.}
Two classes of model are deliberately excluded from the per-skill midtraining-stage diagnostic in \cref{tab:ppl-per-skill}: \emph{(a)} post-trained alignment baselines (\textsc{HER}-32B, \textsc{Sotopia-RL}-7B, GPT-5.5, GPT-5-nano), which conflate midtraining with subsequent RLHF / preference alignment and so are unsuitable for diagnosing the midtraining stage in isolation, and \emph{(b)} thinking models (HumanLM-Opinion-8B, HER-32B-thinking), whose generation streams contain interleaved thinking-token preambles that contaminate token-level loss and surface-form overlap metrics. Both classes appear instead in the evaluation suite in \cref{sec:main_results}, where the metric of interest is prompt-conditioned generation under the model's natural inference setting and these confounds are immaterial.

\subsection{Post-training hyperparameters}

RL experts (GRPO, \vfrl) are initialized from Qwen3-8B-VL-Instruct (\cref{sec:posttraining}), while the distillation SFT is applied on top of the midtrained checkpoint; all stages reuse the same FSDP-2 / \texttt{bfloat16} setup as midtraining, and the canonical method definitions are in \cref{sec:posttraining}.
DPO uses $\beta = 0.1$ on the held-out preference subsplit of the corpus (\cref{tab:split-summary}, \texttt{record\_hash} datasets).
GRPO and \vfrl sample $8$ rollouts per prompt at temperature $1.0$, with no KL loss, asymmetric (dual-clip) PPO clip ratios of $0.2$ (low) and $0.28$ (high), peak learning rate $5\!\times\!10^{-6}$ with LoRA rank $32$, and a batch of $64$ prompts $\times\,8$ rollouts per step processed in PPO mini-batches of $16$ prompts (matching the released training script). \vfrl additionally conditions a second-stage GRPO pass on verbal feedback by prepending the feedback as a leading turn (\cref{sec:posttraining}).

\subsection{Inference for evaluation}

Per-skill PPL is computed teacher-forced on the held-out evaluation split of role-swapped human turns (\cref{tab:split-summary}). BLEU is computed on free generation with greedy decoding under the same prompts. Generative evaluation tasks use temperature $0.7$ via vLLM. Cross-tokenizer rows in \cref{tab:ppl-per-skill} report BLEU only; PPL columns are populated only for models in the Qwen3 tokenizer family.

\section{Additional Results}
\label{app:results}

\subsection{Midtraining Recipe: Token Scaling and \odyssim:Step Mix}
\label{app:midtraining-recipe}

This section gives the full recipe investigation summarized in one sentence in \cref{sec:exp-midtraining}: token-scaling curves at two model scales, and the \odyssim:Step mix-ratio sweep with its (behavioral, generic-instruction) Pareto frontier.

\begin{figure}[t]
\centering
\includegraphics[width=\textwidth]{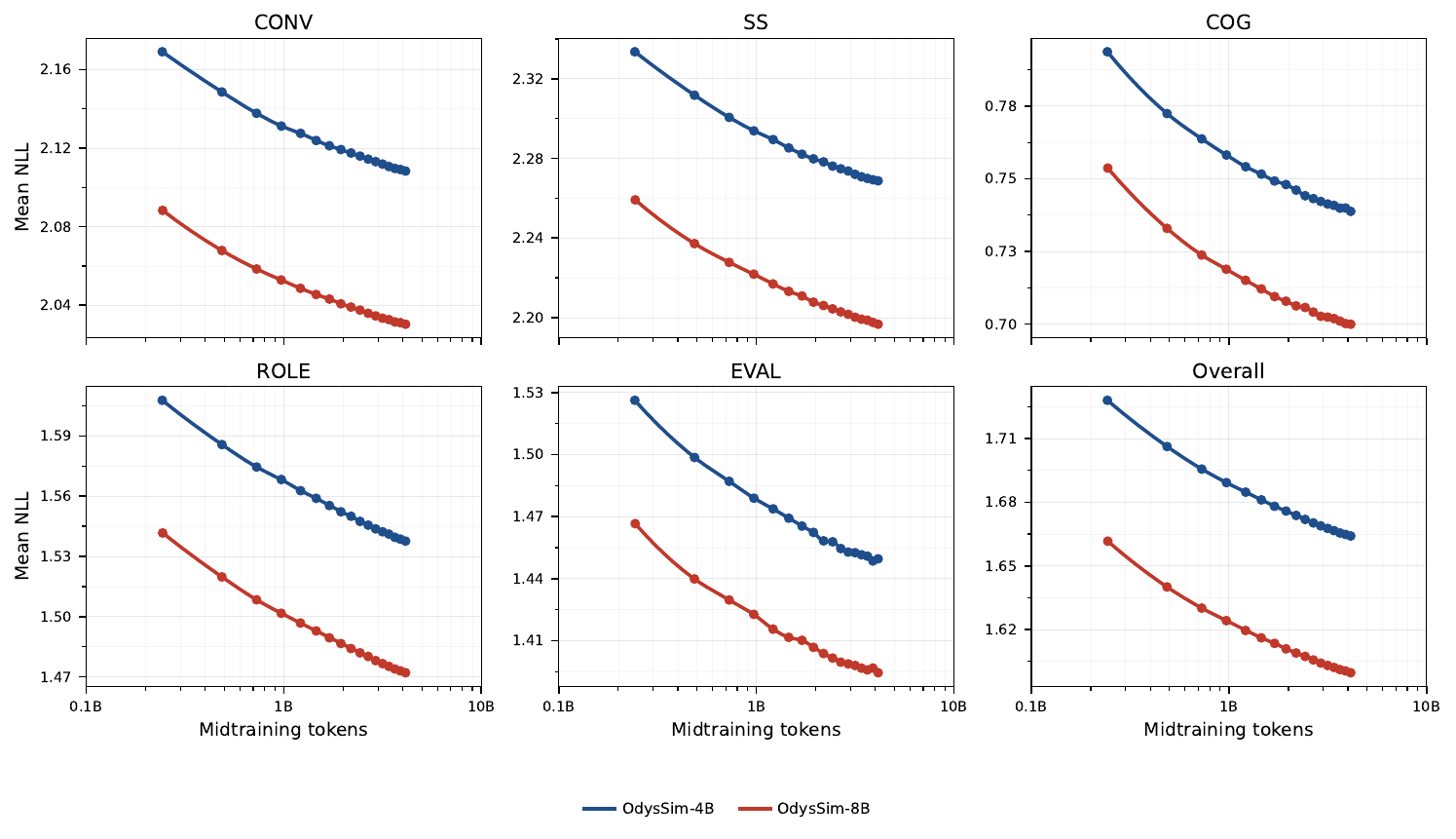}
\caption{Midtraining-token vs.\ geomean validation NLL per Axis. \emph{Solid lines}: \odyssimmodel-4B-Mid (100:0), \odyssimmodel-8B-Mid (100:0), and \odyssimmodel-8B-VL-Mid (100:0). \emph{Faint blue lines}: 4B mix-ratio sweep at 90:10, 70:30, 50:50 (\odyssim:Step). Markers are logged validation checkpoints; lines are PCHIP-smoothed in log-x space between them. Aggregation is the geometric mean over all 63 held-out evaluation datasets.}
\label{fig:scaling-law}
\end{figure}

\begin{figure}[t]
\centering
\includegraphics[width=0.7\textwidth]{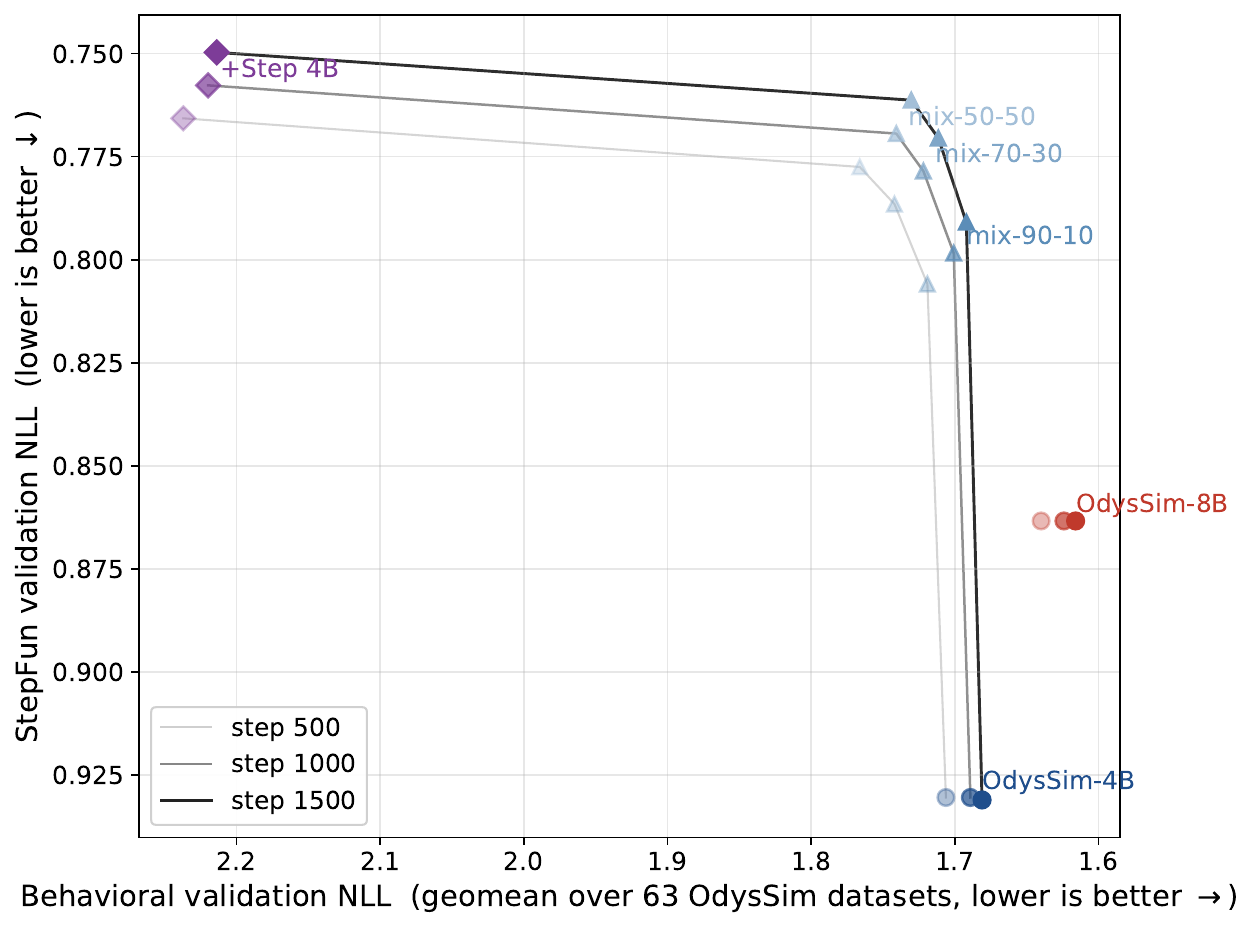}
\caption{Behavior vs general-instruction tradeoff. Both axes are mean validation NLL (nats; lower is better, axes inverted so up-and-right is better). \emph{x}: behavioral NLL averaged over the 63-dataset \odyssim val split. \emph{y}: NLL on the first 512 examples of Step \texttt{Step-3.5-Flash-SFT}; a random-sample spot-check moves the absolute NLL by $\le 0.04$ nats and preserves the cross-model ranking. Grey line traces the 4B mix-ratio sweep. Segment slopes give the marginal trade rate (nats Step gained per nat behavioral loss). \odyssimmodel-8B-Mid sits below-and-left of every 4B point, indicating that model size strictly dominates Step mixing if compute allows.}
\label{fig:pareto}
\end{figure}

\paragraph{Token-scaling curves.}
\Cref{fig:scaling-law} plots geomean validation PPL against midtraining tokens for the \odyssim recipe at two scales (4B, 8B) and two backbones (text-only Qwen3-Base, vision-language Qwen3-VL-Instruct), together with three \odyssim:Step mix ratios at 4B (90:10, 70:30, 50:50). Aggregation is the geometric mean over all 63 held-out evaluation datasets.\footnote{\textbf{Caveat: 63 vs.\ 62 datasets.} All midtraining numbers in this paper (\cref{tab:ppl-per-skill}, \cref{fig:scaling-law}, \cref{fig:pareto}) are computed on a \texttt{v2} eval split of 63 datasets, whereas the released corpus \midtrainingdata contains 62 sources. The single dataset that differs is \texttt{tom\_sotopia} (1{,}289 train / 101 test rows): it is a relabel of self-rejection-sampled rollouts from an earlier project checkpoint and was intended for post-training only, but was accidentally folded into the \texttt{v2} midtraining/eval split that produced all numbers in this paper. We removed it from the released corpus for transparency but did not re-run midtraining and evaluation, because the dataset is one of 63 in a geometric-mean aggregation (its inclusion does not change recipe rankings or qualitative conclusions) and the compute budget did not justify it. Reproducing eval against the released split should produce numerical drift well below the cross-recipe gaps reported here.} We observe three patterns.
\emph{(i) Token scaling is monotone but slow.} Each curve decreases smoothly with tokens; from $\sim$250M to $\sim$4B tokens every Axis improves by another 4--6\% PPL, roughly uniform across \texttt{CONV}, \texttt{SS}, \texttt{COG}, \texttt{ROLE}, and \texttt{EVAL}. The diminishing-returns shape is consistent with prior reports that response-imitation SFT has a ceiling in human-behavior fit~\citep{wu2026humanlm}.
\emph{(ii) Size scales the recipe.} \odyssimmodel-8B-Mid sits below \odyssimmodel-4B-Mid at every shared token count by roughly the same gap on every Axis; the VL backbone (\odyssimmodel-8B-VL-Mid) tracks \odyssimmodel-8B-Mid closely, indicating the multimodal initialisation does not cost behavioral fit.
\emph{(iii) Adding generic chat data hurts behavior monotonically.} The mix curves are ordered $50{:}50 > 70{:}30 > 90{:}10 > 100{:}0$ on every panel: any non-zero Step fraction increases behavioral PPL, and the cost grows with the fraction.

\paragraph{Behavior vs general-instruction tradeoff.}
The natural follow-up is what mixing in Step data \emph{buys} on the generic-instruction side. At matched checkpoint step 1000, we evaluate every midtraining recipe on Step's own held-out test split (\texttt{Step-3.5-Flash-SFT}, a code/math/instruction corpus and the source of our \texttt{+\,Step} baseline) and plot the resulting (behavioral, Step) NLL pairs in \Cref{fig:pareto}. Both axes are in nats, so a line segment between two recipes can be read as the \emph{exchange rate}: how many nats of behavioral loss are paid per nat of Step loss reduced.
\emph{The first 10\% of Step is essentially free; everything after is overpriced.}
Going from pure \odyssim to a 90:10 mix saves $0.13$ nats of Step loss for only $0.013$ nats of behavioral cost --- a roughly ten-to-one bargain. The next step ($90{:}10 \to 70{:}30$) is about even; after that the trade reverses ($70{:}30 \to 50{:}50$ costs twice the behavior it saves; $50{:}50 \to$ pure Step gives up $27\times$ more behavior than it recovers). Pure-Step training is dominated by 50:50 outright (same Step loss, much lower behavioral loss), so it is never the right choice if any behavioral fidelity matters.
\emph{Scale beats mixing.} \odyssimmodel-8B-Mid sits below-and-left of every 4B point in \Cref{fig:pareto}: it is simultaneously better on both axes than any 4B mix. We therefore adopt \odyssimmodel-8B-Mid at $100{:}0$ as the headline midtraining recipe and recommend \odyssimmodel-4B-Mid at $90{:}10$ as the Pareto-frontier option when the smaller scale is required.

\subsection{Full Per-Skill PPL/BLEU Table}
\label{app:ppl-per-skill-full}

\Cref{tab:ppl-per-skill-full} reports the full version of \cref{tab:ppl-per-skill}, including every Instruct (\texttt{-I}) variant, the Qwen3-32B-I reference, and the \texttt{+\,Step} generic-midtraining baseline (Qwen3-4B/4B-I trained on Step-3.5-Flash) that the main-text table compresses out for space.

\begin{table}[ht]
\caption{\textbf{Full per-skill PPL/BLEU table} (companion to \cref{tab:ppl-per-skill}). Geometric-mean PPL (PL $\downarrow$) and arithmetic-mean BLEU (BL $\uparrow$) on the held-out evaluation split of role-swapped human turns. Rows: \textbf{(A)} no midtraining, \textbf{(B)} general midtraining (\texttt{+\,Step}), \textbf{(C)} other behavior-oriented LMs, \textbf{(D)} ours (\cref{sec:midtraining}). Notation: bare = base, ``-I'' = Instruct, ``+\,Step'' = trained on Step-3.5-Flash. PPL within the Qwen3 tokenizer family only; BLEU across all rows. Rows at the last logged val checkpoint (step 4250 for 4B / 8B; step 950 for 0.6B). Best per (metric, capability) in \textbf{bold}. $^\dagger$BLEU values still computed on the V1 val split (pending re-eval on V2); affects only the I/VL variants here. The 4B-Mid and 8B-Mid base rows are computed on V2 val (matching the training data version).}
\label{tab:ppl-per-skill-full}
\centering
\small
\setlength{\tabcolsep}{5pt}
\renewcommand{\arraystretch}{1.05}
\resizebox{\textwidth}{!}{%
\begin{tabular}{l cc cc cc cc cc | cc}
\toprule
\textbf{Model} & \multicolumn{2}{c}{\texttt{CONV}} & \multicolumn{2}{c}{\texttt{SS}} & \multicolumn{2}{c}{\texttt{COG}} & \multicolumn{2}{c}{\texttt{ROLE}} & \multicolumn{2}{c}{\texttt{EVAL}} & \multicolumn{2}{c}{\textbf{Overall}} \\
 \cmidrule(lr){2-3} \cmidrule(lr){4-5} \cmidrule(lr){6-7} \cmidrule(lr){8-9} \cmidrule(lr){10-11} \cmidrule(lr){12-13}
 & PL$\downarrow$ & BL$\uparrow$ & PL$\downarrow$ & BL$\uparrow$ & PL$\downarrow$ & BL$\uparrow$ & PL$\downarrow$ & BL$\uparrow$ & PL$\downarrow$ & BL$\uparrow$ & PL$\downarrow$ & BL$\uparrow$ \\
\midrule
\multicolumn{13}{l}{\textit{(A) No-midtraining baselines}} \\
Qwen3-0.6B        & 20.75 & 0.70 & 25.25 & 0.61 & 11.17 & 2.37 & 11.39 & 6.37 & 54.98 & 3.96 & 17.43 & 2.80 \\
Qwen3-4B          & 14.08 & 1.47 & 17.52 & 1.91 & 8.07 & 5.60 & 8.56 & 10.78 & 24.14 & 6.27 & 12.00 & 5.18 \\
Qwen3-4B-I        & 21.39 & 1.75 & 29.37 & 2.50 & 19.10 & 10.62 & 14.26 & 12.25 & 24.81 & 6.74 & 19.88 & 6.78 \\
Qwen3-8B          & 14.23 & 2.87 & 17.34 & 1.45 & 9.72 & 3.63 & 8.11 & 14.94 & 34.17 & 4.17 & 12.54 & 6.17 \\
Qwen3-8B-I        & 18.44 & 2.07 & 28.36 & 0.85 & 7.00 & 4.80 & 11.68 & 11.39 & 14.38 & 6.83 & 14.13 & 5.31 \\
Qwen3-32B-I       & 15.94 & 2.38 & 25.45 & 0.89 & 6.00 & 4.67 & 10.49 & 10.59 & 12.62 & 8.15 & 12.41 & 5.28 \\
Llama-3.1-8B-I    & 14.34 & 4.11 & 20.09 & 1.93 & 4.16 & 3.58 & 7.82 & 12.41 & 13.47 & 7.81 & 10.04 & 6.23 \\
\midrule
\multicolumn{13}{l}{\textit{(B) General midtraining baseline}} \\
Qwen3-4B + Step      & 12.21 & 3.97 & 14.35 & 1.97 & 5.78 & 3.33 & 7.73 & 13.61 & 7.36 & 8.06 & 9.20 & 6.49 \\
Qwen3-4B-I + Step    & 12.22 & 4.18 & 13.63 & 2.18 & 5.36 & 3.83 & 7.49 & 14.62 & 6.88 & 8.55 & 8.88 & 6.99 \\
\midrule
\multicolumn{13}{l}{\textit{(C) Other behavior-oriented language models}} \\
UserLM-8B         & 11.62 & 5.33 & 13.29 & 5.87 & 4.02 & 6.97 & 6.61 & 4.37 & 12.90 & 1.11 & 8.38 & 5.12 \\
CoSER-8B          & 15.70 & 2.86 & 14.17 & 1.97 & 3.20 & 7.39 & 6.96 & 14.90 & 8.85 & 18.12 & 8.77 & 8.05 \\
\midrule
\multicolumn{13}{l}{\textit{(D) Ours}} \\
\odyssimmodel-0.6B-Mid & 11.99 & 5.51 & 14.18 & 2.01 & 2.65 & 11.75 & 6.46 & 15.26 & 5.68 & 43.02 & 7.35 & 11.81 \\
\odyssimmodel-4B-Mid       & 8.23 & 8.01 & 9.67 & 10.06 & 2.09 & 44.62 & 4.65 & 21.53 & 4.26 & 46.17 & 5.28 & 26.08 \\
\odyssimmodel-4B-I-Mid$^\dagger$     & 9.19 & 8.49 & 9.97 & 9.71 & 2.42 & 37.48 & 5.57 & 19.42 & 4.41 & 43.73 & 5.94 & 19.93 \\
\odyssimmodel-8B-Mid    & \textbf{7.62} & 8.48 & \textbf{9.00} & 12.44 & 2.01 & \textbf{44.73} & \textbf{4.36} & 22.49 & \textbf{4.03} & 45.47 & \textbf{4.95} & \textbf{26.72} \\
\odyssim-8B-I$^\dagger$  & 7.73 & \textbf{9.11} & 9.05 & 2.45 & \textbf{2.00} & 18.52 & 4.41 & \textbf{20.22} & 4.06 & \textbf{46.70} & 4.99 & 15.93 \\
\bottomrule
\end{tabular}}
\end{table}

\subsection{What Does Midtraining Change About the Model?}
\label{app:midtraining-probe}

\Cref{sec:exp-midtraining} showed \emph{that} midtraining helps; this section asks \emph{what} concretely changes in the model's outputs.
We triangulate with two probes (\Cref{fig:midtraining-panel}): an open-coded inventory of lexical/structural features on the BLEU-eval generations, and the HumT human-likeness scalar of Cheng et al.~\cite{cheng2025humt}.

\paragraph{Surface features.}
Reading paired (instruct-baseline, \odyssim) generations, we open-coded an inventory of \textsc{Style} features (response length, Markdown markup, em-dash usage) and \textsc{Assistant-trait} features (chatbot boilerplate, identity confusion), then scored $N{=}1{,}100$ BLEU-eval prompts (dropping COG and EVAL where conversational style is not a meaningful reference) for two off-the-shelf instruct baselines---Qwen3-4B-Instruct-2507~\citep{qwen3} and GPT-5.5~\citep{openai2025gpt55}---against \odyssim and the human gold reference.
Both instruct baselines emit verbose, Markdown-heavy responses (median $116$/$83$ words; $22.8\%$/$23.9\%$ Markdown markup; $78.2\%$/$40.7\%$ em-dash usage; $18.2\%$/$19.5\%$ bullet markup); \odyssim collapses to the human register ($18$ words, $1.6\%$ Markdown, $0.5\%$ em-dash, $0.8\%$ bullets, vs.\ human gold $23$ / $2.9\%$ / $2.1\%$ / $2.0\%$).
Assistant boilerplate (\emph{``I'd be happy to,'' ``Of course!'' ``As an AI''}) drops from $16.3\%$ (Qwen3-Inst) and $6.5\%$ (GPT-5.5) to $3.5\%$ for \odyssim, against a human-gold rate of $6.4\%$.

\paragraph{Anthropomorphism scalar.}
HumT~\citep{cheng2025humt} computes per-text human-likeness as the log-prob ratio of animate vs.\ inanimate prefixes under a fixed GPT-2 backbone (higher = more human).
On HumT's released $N{=}200$ test prompts, \odyssim's mean is $+0.13$---roughly $4\times$ Qwen3-Inst-2507's $+0.03$ and $3\times$ GPT-5.5's $+0.05$, and well above HumT's own \textsc{rejected}-anchor distribution at $+0.06$.
The SocioT companion confirms the shift is along anthropomorphism-correlated directions: \odyssim is humbler (lower status), more socially close, and slightly warmer than either instruct baseline.

Both probes converge: midtraining moves the model away from the verbose, Markdown-heavy, helpful-agent register of off-the-shelf instruct LMs toward the shorter, plainer, conversational register of human references.
The shift comes entirely from the SFT data mix (no preference tuning) and is large: an order-of-magnitude reduction in response length, $\sim$$10\times$ less structural markup, and $3$--$4\times$ higher HumT.

\end{document}